\documentclass{article}
\usepackage{fullpage}
\usepackage{graphicx}
\usepackage{xspace}
\newcommand{\concept}[1]{\textsf{\textsc{#1}}}
\newcommand{\property}[1]{\textsf{#1}}
\newcommand{\instance}[1]{\textit{#1}}
\newcommand{\datatype}[1]{\textit{#1}}
\newcommand{\datainstance}[1]{\mbox{#1}}
\newcounter{idaxiom}
\newcommand{\ontoBPMN}{\texttt{OntoBPMN.owl}\xspace}

\DeclareGraphicsExtensions{.pdf,.png,.jpg}

\begin{document}
\title{A formalisation of BPMN in Description Logics}
\author{Chiara Ghidini \and
Marco Rospocher \and 
Luciano Serafini}
\date{FBK-irst, Via Sommarive 18 Povo, 38050,Trento, Italy \\
\texttt{\{ghidini,rospocher,serafini\}@fbk.eu}}

\maketitle

\begin{abstract}
In this paper we present a textual description, in terms of Description Logics, of the BPMN Ontology (available for download at \texttt{dkm.fbk.eu/index.php/Resources}), which provides a clear semantic formalisation of the structural components of the Business Process Modelling Notation (BPMN), based on the latest stable BPMN specifications from OMG [BPMN Version 1.1 - January 2008]. The development of the ontology was guided by the description of the complete set of BPMN Element Attributes and Types contained in Annex B of the BPMN specifications.  
\end{abstract}

\section{Introduction}
\label{sec:introduction}
The ontology \ontoBPMN\footnote{Available for download at \texttt{dkm.fbk.eu/index.php/Resources}.} provides a clear semantic formalisation of the structural components of BPMN, based on the latest stable BPMN specifications from OMG \cite{BPMNv1.1}. The development of the ontology was guided by the description of the complete set of BPMN Element Attributes and Types contained in Annex B of the document cited above. The ontology currently consists of 95 Classes and 439 class axioms, 108 Object Properties and 18 Object Property Axioms, and 70 Data Properties; it has the expressiveness of \ensuremath{\mathcal{ALCHOIN(D)}}. In this paper we provide a textual description of its Description Logic version.

The core component of \ontoBPMN is the set of BPMN Elements, divided in two disjoint classes \textit{Graphical Element} and \textit{Supporting Element}. \textit{Graphical Element} contains the main elements used to describe Business Processes, namely \emph{Flow Object}, \emph{Connecting Object}, \emph{Swimline}, and \emph{Artifact}, then further specified in terms of sub-classes. For instance \emph{Connecting Object} is then composed of the disjoint (sub-)classes \emph{Sequence Flow}, \emph{Message Flow}, and \emph{Association}, and do on. \textit{Supporting Element} instead contains 16 additional types of elements, and few additional subclasses, mainly used to specify the attributes of Graphical Objects. To provide an example, the supporting element \emph{input\_set} is used to define an attribute of the graphical object \emph{Activity} which describes the data requirements for input of the activity. 

Note that, while the taxonomy of concepts defines an important part of \ontoBPMN, it constitutes only part of the OWL version of BPMN: in fact, it also specifies the rich set of elements' attributes, and the properties which describe how to use these elements to compose the business process diagrams. 
As an example, BPMN specifies that \textit{Connecting Object} has two attributes (\textit{SourceRef}, \textit{TargetRef}) which point to the two corresponding \textit{Graphical Elements} connected by it. 
As another example, BPMN not only introduces the notion of \textit{Start Event} as a particular, optional, Event, but also specifies that \emph{``The Condition attribute for all outgoing Sequence Flow [from a Start Event] MUST be set to None''}. Thus the BPMN specification tells that the graphical element \textit{Start Event} is a sub-class of \emph{Event}. Moreover it tells us that if an object of kind \emph{Start Event} is connected to an object of kind \emph{Sequence Flow}, then this Sequence Flow object must have a Condition attribute whose value is ``None''.  As a consequence of our effort towards the modelling of properties, \ontoBPMN contains, at the current state more than 400 class axioms, which describe a wide set of properties of the BPMN elements.  

While our aim is to formalise the widest set of BPMN specifications, the \ontoBPMN ontology does not contain a description of all the properties documented in \cite{BPMNv1.1}. First of all, because we have chosen not to formalise properties which refer to the execution behaviour of the process. Second, because of well known limitations in the expressiveness of the OWL language. In this specific case, most of the properties of BPMN that are not expressible in OWL, and therefore not included in \ontoBPMN, concern: (i) attributes' default values, and (ii) all the properties that, translated in first order logic, require more than two variables. Prototypical examples of this kind of properties are the ones which refer to the uniqueness, or equality, of objects: for instance the properties which specify that \emph{``two objects cannot have the same object identifier''} or that \emph{``all outgoing sequence flows connected to an inclusive gateway must have the same conditional expression attached''}. 

\section{The BPMN Ontology}
\label{sec:ontology}
\vspace{0.4cm} \hrule \vspace{0.2cm} \noindent \textbf{Class}: $\concept{business\_process\_diagram}$\vspace{0.2cm} \hrule \vspace{0.2cm} 
\noindent \textbf{Label}: Business Process Diagram\vspace{0.1cm} \\
\noindent \textbf{Description}: Gather the set of attributes of a 
	Business Process Diagram\vspace{0.1cm} \\
\noindent \stepcounter{idaxiom} $AX\_\arabic{idaxiom}$ $\concept{business\_process\_diagram} \sqsubseteq  (=1) \property{has\_business\_process\_diagram\_id}$\vspace{0.1cm} \\
\par \vspace{-0.1cm} \noindent \textbf{Property}: $\property{has\_business\_process\_diagram\_id}$\vspace{0.1cm} \\
\noindent \textbf{Label}: Id\vspace{0.1cm} \\
\noindent \textbf{Description}: This is a unique Id that 
	identifies the object from other objects within the 
	business\_process\_diagram.\vspace{0.1cm} \\
\noindent \stepcounter{idaxiom} $AX\_\arabic{idaxiom}$ $\property{has\_business\_process\_diagram\_id} \mbox{ has range }\concept{object}$\vspace{0.1cm} \\
\noindent \stepcounter{idaxiom} $AX\_\arabic{idaxiom}$ $\property{has\_business\_process\_diagram\_id} \mbox{ has domain }\concept{business\_process\_diagram}$\vspace{0.1cm} \\
\noindent \stepcounter{idaxiom} $AX\_\arabic{idaxiom}$ $\concept{business\_process\_diagram} \sqsubseteq  (=1) \property{has\_business\_process\_diagram\_name}$\vspace{0.1cm} \\
\par \vspace{-0.1cm} \noindent \textbf{Property}: $\property{has\_business\_process\_diagram\_name}$\vspace{0.1cm} \\
\noindent \textbf{Label}: Name\vspace{0.1cm} \\
\noindent \textbf{Description}: Name is an attribute that is
	 text description of the Diagram.\vspace{0.1cm} \\
\noindent \stepcounter{idaxiom} $AX\_\arabic{idaxiom}$ $\property{has\_business\_process\_diagram\_name} \mbox{ has range }\datatype{xsd:string}$\vspace{0.1cm} \\
\noindent \stepcounter{idaxiom} $AX\_\arabic{idaxiom}$ $\property{has\_business\_process\_diagram\_name} \mbox{ has domain }\concept{business\_process\_diagram}$\vspace{0.1cm} \\
\noindent \stepcounter{idaxiom} $AX\_\arabic{idaxiom}$ $\concept{business\_process\_diagram} \sqsubseteq (\geq1) \property{has\_business\_process\_diagram\_version}$\vspace{0.1cm} \\
\par \vspace{-0.1cm} \noindent \textbf{Property}: $\property{has\_business\_process\_diagram\_version}$\vspace{0.1cm} \\
\noindent \textbf{Label}: Version\vspace{0.1cm} \\
\noindent \textbf{Description}: This defines the Version 
	number of the Diagram.\vspace{0.1cm} \\
\noindent \stepcounter{idaxiom} $AX\_\arabic{idaxiom}$ $\property{has\_business\_process\_diagram\_version} \mbox{ has range }\datatype{xsd:string}$\vspace{0.1cm} \\
\noindent \stepcounter{idaxiom} $AX\_\arabic{idaxiom}$ $\property{has\_business\_process\_diagram\_version} \mbox{ has domain }\concept{business\_process\_diagram}$\vspace{0.1cm} \\
\noindent \stepcounter{idaxiom} $AX\_\arabic{idaxiom}$ $\concept{business\_process\_diagram} \sqsubseteq (\geq1) \property{has\_business\_process\_diagram\_author}$\vspace{0.1cm} \\
\par \vspace{-0.1cm} \noindent \textbf{Property}: $\property{has\_business\_process\_diagram\_author}$\vspace{0.1cm} \\
\noindent \textbf{Label}: Author\vspace{0.1cm} \\
\noindent \textbf{Description}: This holds the name of the
	author of the Diagram.\vspace{0.1cm} \\
\noindent \stepcounter{idaxiom} $AX\_\arabic{idaxiom}$ $\property{has\_business\_process\_diagram\_author} \mbox{ has range }\datatype{xsd:string}$\vspace{0.1cm} \\
\noindent \stepcounter{idaxiom} $AX\_\arabic{idaxiom}$ $\property{has\_business\_process\_diagram\_author} \mbox{ has domain }\concept{business\_process\_diagram}$\vspace{0.1cm} \\
\noindent \stepcounter{idaxiom} $AX\_\arabic{idaxiom}$ $\concept{business\_process\_diagram} \sqsubseteq (\geq1) \property{has\_business\_process\_diagram\_language}$\vspace{0.1cm} \\
\par \vspace{-0.1cm} \noindent \textbf{Property}: $\property{has\_business\_process\_diagram\_language}$\vspace{0.1cm} \\
\noindent \textbf{Label}: Language\vspace{0.1cm} \\
\noindent \textbf{Description}: This holds the name of
	the language in which text is written. The default is English.\vspace{0.1cm} \\
\noindent \stepcounter{idaxiom} $AX\_\arabic{idaxiom}$ $\property{has\_business\_process\_diagram\_language} \mbox{ has range }\datatype{xsd:string}$\vspace{0.1cm} \\
\noindent \stepcounter{idaxiom} $AX\_\arabic{idaxiom}$ $\property{has\_business\_process\_diagram\_language} \mbox{ has domain }\concept{business\_process\_diagram}$\vspace{0.1cm} \\
\noindent \stepcounter{idaxiom} $AX\_\arabic{idaxiom}$ $\concept{business\_process\_diagram} \sqsubseteq (\geq1) \property{has\_business\_process\_diagram\_query\_language}$\vspace{0.1cm} \\
\par \vspace{-0.1cm} \noindent \textbf{Property}: $\property{has\_business\_process\_diagram\_query\_language}$\vspace{0.1cm} \\
\noindent \textbf{Label}: Query Language\vspace{0.1cm} \\
\noindent \textbf{Description}: A Language MAY be
	provided so that the syntax of queries used in the Diagram can
	be understood.\vspace{0.1cm} \\
\noindent \stepcounter{idaxiom} $AX\_\arabic{idaxiom}$ $\property{has\_business\_process\_diagram\_query\_language} \mbox{ has range }\datatype{xsd:string}$\vspace{0.1cm} \\
\noindent \stepcounter{idaxiom} $AX\_\arabic{idaxiom}$ $\property{has\_business\_process\_diagram\_query\_language} \mbox{ has domain }\concept{business\_process\_diagram}$\vspace{0.1cm} \\
\noindent \stepcounter{idaxiom} $AX\_\arabic{idaxiom}$ $\concept{business\_process\_diagram} \sqsubseteq (\geq1) \property{has\_business\_process\_diagram\_creation\_date}$\vspace{0.1cm} \\
\par \vspace{-0.1cm} \noindent \textbf{Property}: $\property{has\_business\_process\_diagram\_creation\_date}$\vspace{0.1cm} \\
\noindent \textbf{Label}: Creation Date\vspace{0.1cm} \\
\noindent \textbf{Description}: This defines the
	date on which the Diagram was create (for this Version).\vspace{0.1cm} \\
\noindent \stepcounter{idaxiom} $AX\_\arabic{idaxiom}$ $\property{has\_business\_process\_diagram\_creation\_date} \mbox{ has range }\datatype{xsd:date}$\vspace{0.1cm} \\
\noindent \stepcounter{idaxiom} $AX\_\arabic{idaxiom}$ $\property{has\_business\_process\_diagram\_creation\_date} \mbox{ has domain }\concept{business\_process\_diagram}$\vspace{0.1cm} \\
\noindent \stepcounter{idaxiom} $AX\_\arabic{idaxiom}$ $\concept{business\_process\_diagram} \sqsubseteq (\geq1) \property{has\_business\_process\_diagram\_modification\_date}$\vspace{0.1cm} \\
\par \vspace{-0.1cm} \noindent \textbf{Property}: $\property{has\_business\_process\_diagram\_modification\_date}$\vspace{0.1cm} \\
\noindent \textbf{Label}: Modification Date\vspace{0.1cm} \\
\noindent \textbf{Description}: This defines
	the date on which the Diagram was last modified (for this Version).\vspace{0.1cm} \\
\noindent \stepcounter{idaxiom} $AX\_\arabic{idaxiom}$ $\property{has\_business\_process\_diagram\_modification\_date} \mbox{ has range }\datatype{xsd:date}$\vspace{0.1cm} \\
\noindent \stepcounter{idaxiom} $AX\_\arabic{idaxiom}$ $\property{has\_business\_process\_diagram\_modification\_date} \mbox{ has domain }\concept{business\_process\_diagram}$\vspace{0.1cm} \\
\noindent \stepcounter{idaxiom} $AX\_\arabic{idaxiom}$ $\concept{business\_process\_diagram} \sqsubseteq (\leq1) \property{has\_business\_process\_diagram\_pools}$\vspace{0.1cm} \\
\par \vspace{-0.1cm} \noindent \textbf{Property}: $\property{has\_business\_process\_diagram\_pools}$\vspace{0.1cm} \\
\noindent \textbf{Label}: Pools\vspace{0.1cm} \\
\noindent \textbf{Description}: A BPD SHALL contain one
	or more Pools. The boundary of one of the Pools MAY be invisible 
	(especially if there is only one Pool in the Diagram). Refer to "Pool"
	on page 75 for more information about Pools.\vspace{0.1cm} \\
\noindent \stepcounter{idaxiom} $AX\_\arabic{idaxiom}$ $\property{has\_business\_process\_diagram\_pools} \mbox{ has range }\concept{pool}$\vspace{0.1cm} \\
\noindent \stepcounter{idaxiom} $AX\_\arabic{idaxiom}$ $\property{has\_business\_process\_diagram\_pools} \mbox{ has domain }\concept{business\_process\_diagram}$\vspace{0.1cm} \\
\noindent \stepcounter{idaxiom} $AX\_\arabic{idaxiom}$ $\concept{business\_process\_diagram} \sqsubseteq (\geq1) \property{has\_business\_process\_diagram\_documentation}$\vspace{0.1cm} \\
\par \vspace{-0.1cm} \noindent \textbf{Property}: $\property{has\_business\_process\_diagram\_documentation}$\vspace{0.1cm} \\
\noindent \textbf{Label}: Documentation\vspace{0.1cm} \\
\noindent \textbf{Description}: The modeler MAY 
	add optional text documentation about the Diagram.\vspace{0.1cm} \\
\noindent \stepcounter{idaxiom} $AX\_\arabic{idaxiom}$ $\property{has\_business\_process\_diagram\_documentation} \mbox{ has range }\datatype{xsd:string}$\vspace{0.1cm} \\
\noindent \stepcounter{idaxiom} $AX\_\arabic{idaxiom}$ $\property{has\_business\_process\_diagram\_documentation} \mbox{ has domain }\concept{business\_process\_diagram}$\vspace{0.1cm} \\
\\
\vspace{0.4cm} \hrule \vspace{0.2cm} \noindent \textbf{Class}: $\concept{BPMN\_element}$\vspace{0.2cm} \hrule \vspace{0.2cm} 
\noindent \textbf{Label}: BPMN element\vspace{0.1cm} \\
\noindent \textbf{Description}: Base element\vspace{0.1cm} \\
\noindent \stepcounter{idaxiom} $AX\_\arabic{idaxiom}$ $\concept{BPMN\_element} \equiv \concept{graphical\_element} \sqcup \concept{supporting\_element}$\vspace{0.1cm} \\
\noindent \stepcounter{idaxiom} $AX\_\arabic{idaxiom}$ $\concept{graphical\_element} \sqsubseteq  \neg \concept{supporting\_element}$\vspace{0.1cm} \\
\noindent \stepcounter{idaxiom} $AX\_\arabic{idaxiom}$ $\concept{BPMN\_element} \sqsubseteq  (=1) \property{has\_BPMN\_element\_id}$\vspace{0.1cm} \\
\par \vspace{-0.1cm} \noindent \textbf{Property}: $\property{has\_BPMN\_element\_id}$\vspace{0.1cm} \\
\noindent \textbf{Label}: Id\vspace{0.1cm} \\
\noindent \textbf{Description}: This is a unique Id that identifies the 
	object from other objects within the Diagram.\vspace{0.1cm} \\
\noindent \stepcounter{idaxiom} $AX\_\arabic{idaxiom}$ $\property{has\_BPMN\_element\_id} \mbox{ has range }\concept{object}$\vspace{0.1cm} \\
\noindent \stepcounter{idaxiom} $AX\_\arabic{idaxiom}$ $\property{has\_BPMN\_element\_id} \mbox{ has domain }\concept{BPMN\_element}$\vspace{0.1cm} \\
\par \vspace{-0.1cm} \noindent \textbf{Property}: $\property{has\_BPMN\_element\_category}$\vspace{0.1cm} \\
\noindent \textbf{Label}: Category\vspace{0.1cm} \\
\noindent \textbf{Description}: The modeler MAY add one or more 
	defined Categories, which have user-defined semantics, and that can be 
	used for purposes such as reporting and analysis. The details of 
	Catogories is defined in Category on page 269.\vspace{0.1cm} \\
\noindent \stepcounter{idaxiom} $AX\_\arabic{idaxiom}$ $\property{has\_BPMN\_element\_category} \mbox{ has range }\concept{category}$\vspace{0.1cm} \\
\noindent \stepcounter{idaxiom} $AX\_\arabic{idaxiom}$ $\property{has\_BPMN\_element\_category} \mbox{ has domain }\concept{BPMN\_element}$\vspace{0.1cm} \\
\noindent \stepcounter{idaxiom} $AX\_\arabic{idaxiom}$ $\concept{BPMN\_element} \sqsubseteq (\geq1) \property{has\_BPMN\_element\_documentation}$\vspace{0.1cm} \\
\par \vspace{-0.1cm} \noindent \textbf{Property}: $\property{has\_BPMN\_element\_documentation}$\vspace{0.1cm} \\
\noindent \textbf{Label}: Documentation\vspace{0.1cm} \\
\noindent \textbf{Description}: The modeler MAY add text 
	documentation about the object.\vspace{0.1cm} \\
\noindent \stepcounter{idaxiom} $AX\_\arabic{idaxiom}$ $\property{has\_BPMN\_element\_documentation} \mbox{ has range }\datatype{xsd:string}$\vspace{0.1cm} \\
\noindent \stepcounter{idaxiom} $AX\_\arabic{idaxiom}$ $\property{has\_BPMN\_element\_documentation} \mbox{ has domain }\concept{BPMN\_element}$\vspace{0.1cm} \\
\vspace{0.4cm} \hrule \vspace{0.2cm} \noindent \textbf{Class}: $\concept{graphical\_element}$\vspace{0.2cm} \hrule \vspace{0.2cm} 
\noindent \textbf{Label}: Graphical element\vspace{0.1cm} \\
\noindent \textbf{Description}: These are the elements that define the 
	basic look-and-feel of BPMN. Most business processes will be modeled 
	adequately with these elements\vspace{0.1cm} \\
\noindent \stepcounter{idaxiom} $AX\_\arabic{idaxiom}$ $\concept{graphical\_element} \equiv \concept{flow\_object} \sqcup (\concept{connecting\_object} \sqcup (\concept{swimlane} \sqcup \concept{artifact}))$\vspace{0.1cm} \\
\noindent \stepcounter{idaxiom} $AX\_\arabic{idaxiom}$ $\concept{flow\_object} \sqsubseteq  \neg \concept{connecting\_object}$\vspace{0.1cm} \\
\noindent \stepcounter{idaxiom} $AX\_\arabic{idaxiom}$ $\concept{flow\_object} \sqsubseteq  \neg \concept{swimlane}$\vspace{0.1cm} \\
\noindent \stepcounter{idaxiom} $AX\_\arabic{idaxiom}$ $\concept{flow\_object} \sqsubseteq  \neg \concept{artifact}$\vspace{0.1cm} \\
\noindent \stepcounter{idaxiom} $AX\_\arabic{idaxiom}$ $\concept{connecting\_object} \sqsubseteq  \neg \concept{swimlane}$\vspace{0.1cm} \\
\noindent \stepcounter{idaxiom} $AX\_\arabic{idaxiom}$ $\concept{connecting\_object} \sqsubseteq  \neg \concept{artifact}$\vspace{0.1cm} \\
\noindent \stepcounter{idaxiom} $AX\_\arabic{idaxiom}$ $\concept{swimlane} \sqsubseteq  \neg \concept{artifact}$\vspace{0.1cm} \\
\vspace{0.4cm} \hrule \vspace{0.2cm} \noindent \textbf{Class}: $\concept{flow\_object}$\vspace{0.2cm} \hrule \vspace{0.2cm} 
\noindent \textbf{Label}: Flow Object\vspace{0.1cm} \\
\noindent \textbf{Description}: Flow objects are the main graphical elements to 
	define the behavior of a Business Process. There are three Flow Objects:
	Events, Activities and Gateways\vspace{0.1cm} \\
\noindent \stepcounter{idaxiom} $AX\_\arabic{idaxiom}$ $\concept{flow\_object} \equiv \concept{event} \sqcup (\concept{activity} \sqcup \concept{gateway})$\vspace{0.1cm} \\
\noindent \stepcounter{idaxiom} $AX\_\arabic{idaxiom}$ $\concept{event} \sqsubseteq  \neg \concept{activity}$\vspace{0.1cm} \\
\noindent \stepcounter{idaxiom} $AX\_\arabic{idaxiom}$ $\concept{event} \sqsubseteq  \neg \concept{gateway}$\vspace{0.1cm} \\
\noindent \stepcounter{idaxiom} $AX\_\arabic{idaxiom}$ $\concept{activity} \sqsubseteq  \neg \concept{gateway}$\vspace{0.1cm} \\
\noindent \stepcounter{idaxiom} $AX\_\arabic{idaxiom}$ $\concept{flow\_object} \sqsubseteq  (=1) \property{has\_flow\_object\_name}$\vspace{0.1cm} \\
\par \vspace{-0.1cm} \noindent \textbf{Property}: $\property{has\_flow\_object\_name}$\vspace{0.1cm} \\
\noindent \textbf{Label}: Name\vspace{0.1cm} \\
\noindent \textbf{Description}: Name is an attribute that is a text 
	description of the object.\vspace{0.1cm} \\
\noindent \stepcounter{idaxiom} $AX\_\arabic{idaxiom}$ $\property{has\_flow\_object\_name} \mbox{ has domain }\concept{flow\_object}$\vspace{0.1cm} \\
\noindent \stepcounter{idaxiom} $AX\_\arabic{idaxiom}$ $\property{has\_flow\_object\_name} \mbox{ has range }\datatype{xsd:string}$\vspace{0.1cm} \\
\par \vspace{-0.1cm} \noindent \textbf{Property}: $\property{has\_flow\_object\_assignment}$\vspace{0.1cm} \\
\noindent \textbf{Label}: Assignment\vspace{0.1cm} \\
\noindent \textbf{Description}: One or more assignment expressions
	MAY be made for the object. For activities, the Assignment SHALL be 
	performed as defined by the AssignTime attribute. The Details of the 
	Assignment is defined in Assignment on page 269.\vspace{0.1cm} \\
\noindent \stepcounter{idaxiom} $AX\_\arabic{idaxiom}$ $\property{has\_flow\_object\_assignment} \mbox{ has domain }\concept{flow\_object}$\vspace{0.1cm} \\
\noindent \stepcounter{idaxiom} $AX\_\arabic{idaxiom}$ $\property{has\_flow\_object\_assignment} \mbox{ has range }\concept{assignment}$\vspace{0.1cm} \\
\vspace{0.4cm} \hrule \vspace{0.2cm} \noindent \textbf{Class}: $\concept{event}$\vspace{0.2cm} \hrule \vspace{0.2cm} 
\noindent \textbf{Label}: Event\vspace{0.1cm} \\
\noindent \textbf{Description}: An event is something that "happens" during the course 
	of a business process. These events affect the flow of the process and 
	usually have a cause (trigger) or an impact  (result). Events are 
	circles with open centers to allow internal markers to differentiate 
	different triggers or results. There are three types of Events, based 
	on when they affect the flow: Start, Intermediate, and End.\vspace{0.1cm} \\
\noindent \stepcounter{idaxiom} $AX\_\arabic{idaxiom}$ $\concept{event} \sqsubseteq  (=1) \property{has\_event\_type}$\vspace{0.1cm} \\
\par \vspace{-0.1cm} \noindent \textbf{Property}: $\property{has\_event\_type}$\vspace{0.1cm} \\
\noindent \textbf{Label}: EventType\vspace{0.1cm} \\
\noindent \textbf{Description}: An event is associated with a flow Dimension 
	(e.g.,Start, Intermediate, End)\vspace{0.1cm} \\
\noindent \stepcounter{idaxiom} $AX\_\arabic{idaxiom}$ $\property{has\_event\_type} \mbox{ has domain }\concept{event}$\vspace{0.1cm} \\
\noindent \stepcounter{idaxiom} $AX\_\arabic{idaxiom}$ $\property{has\_event\_type} \mbox{ has range }\concept{event\_types}$\vspace{0.1cm} \\
\noindent \stepcounter{idaxiom} $AX\_\arabic{idaxiom}$ $\concept{event\_types} \equiv \{\instance{start}, \instance{intermediate}, \instance{end}\}$\vspace{0.1cm} \\
\par \vspace{-0.1cm} \noindent \textbf{Instance}: $\instance{start}$\vspace{0.1cm} \\
\noindent \textbf{Label}: start\vspace{0.1cm} \\
\par \vspace{-0.1cm} \noindent \textbf{Instance}: $\instance{intermediate}$\vspace{0.1cm} \\
\noindent \textbf{Label}: intermediate\vspace{0.1cm} \\
\par \vspace{-0.1cm} \noindent \textbf{Instance}: $\instance{end}$\vspace{0.1cm} \\
\noindent \textbf{Label}: end\vspace{0.1cm} \\
\noindent \stepcounter{idaxiom} $AX\_\arabic{idaxiom}$ $\concept{start\_event} \equiv \concept{event} \sqcap \exists\property{has\_event\_type}.\{\instance{start}\}$\vspace{0.1cm} \\
\noindent \stepcounter{idaxiom} $AX\_\arabic{idaxiom}$ $\concept{intermediate\_event} \equiv \concept{event} \sqcap \exists\property{has\_event\_type}.\{\instance{intermediate}\}$\vspace{0.1cm} \\
\noindent \stepcounter{idaxiom} $AX\_\arabic{idaxiom}$ $\concept{end\_event} \equiv \concept{event} \sqcap \exists\property{has\_event\_type}.\{\instance{end}\}$\vspace{0.1cm} \\
\noindent \stepcounter{idaxiom} $AX\_\arabic{idaxiom}$ $\concept{start\_event} \sqsubseteq  \neg \concept{intermediate\_event}$\vspace{0.1cm} \\
\noindent \stepcounter{idaxiom} $AX\_\arabic{idaxiom}$ $\concept{start\_event} \sqsubseteq  \neg \concept{end\_event}$\vspace{0.1cm} \\
\noindent \stepcounter{idaxiom} $AX\_\arabic{idaxiom}$ $\concept{intermediate\_event} \sqsubseteq  \neg \concept{end\_event}$\vspace{0.1cm} \\
\vspace{0.4cm} \hrule \vspace{0.2cm} \noindent \textbf{Class}: $\concept{start\_event}$\vspace{0.2cm} \hrule \vspace{0.2cm} 
\noindent \textbf{Label}: Start\vspace{0.1cm} \\
\noindent \textbf{Description}: As the name implies, the Start Event
                      indicates where a particular process will
                      start.\vspace{0.1cm} \\
\par \vspace{-0.1cm} \noindent \textbf{Property}: $\property{has\_start\_event\_trigger}$\vspace{0.1cm} \\
\noindent \textbf{Label}: Trigger\vspace{0.1cm} \\
\noindent \textbf{Description}: Trigger (EventDetail)) is an attribute
	that defines the type of trigger expected for a Start Event. Of the set
	of EventDetailTypes (see Section B.11.7, "Event Details," on page 270),
	only four (4) can be applied to a Start Event: Message, Timer, 
	Conditional, and Signal (see Table 9.4). 
	If there is no EventDetail is defined, then this is considered a None 
	End Event and the Event will not have an internal marker (see Table 9.4).
	If there is more than one EventDetail is defined, this is considered a 
	Multiple End Event and the Event will have the star internal marker 
	(see Table 9.4).\vspace{0.1cm} \\
\noindent \stepcounter{idaxiom} $AX\_\arabic{idaxiom}$ $\property{has\_start\_event\_trigger} \mbox{ has domain }\concept{start\_event}$\vspace{0.1cm} \\
\noindent \stepcounter{idaxiom} $AX\_\arabic{idaxiom}$ $\property{has\_start\_event\_trigger} \mbox{ has range }\concept{message\_event\_detail} \sqcup \concept{timer\_event\_detail} \sqcup \\ \concept{conditional\_event\_detail} \sqcup \concept{signal\_event\_detail}$\vspace{0.1cm} \\
\vspace{0.4cm} \hrule \vspace{0.2cm} \noindent \textbf{Class}: $\concept{end\_event}$\vspace{0.2cm} \hrule \vspace{0.2cm} 
\noindent \textbf{Label}: End\vspace{0.1cm} \\
\noindent \textbf{Description}: As the name implies, the End Event
                     indicates where a process will end.\vspace{0.1cm} \\
\par \vspace{-0.1cm} \noindent \textbf{Property}: $\property{has\_end\_event\_result}$\vspace{0.1cm} \\
\noindent \textbf{Label}: Result\vspace{0.1cm} \\
\noindent \textbf{Description}: Result (EventDetail) is an attribute that
	defines the type of result expected for an End Event. Of the set of 
	EventDetailTypes (see Section B.11.7, "Event Details," on page 270), 
	only six (6) can be applied to an End Event: Message, Error, Cancel, 
	Compensation, Signal, and Terminate (see Table 9.6).
	If there is no EventDetail is defined, then this is considered a None 
	End Event and the Event will not have an internal marker
	(see Table 9.6).
	If there is more than one EventDetail is defined, this is considered a 
	Multiple End Event and the Event will have the star internal marker 
	(see Table 9.6).\vspace{0.1cm} \\
\noindent \stepcounter{idaxiom} $AX\_\arabic{idaxiom}$ $\property{has\_end\_event\_result} \mbox{ has domain }\concept{end\_event}$\vspace{0.1cm} \\
\noindent \stepcounter{idaxiom} $AX\_\arabic{idaxiom}$ $\property{has\_end\_event\_result} \mbox{ has range }\concept{message\_event\_detail} \sqcup \concept{error\_event\_detail} \sqcup \concept{cancel\_event\_detail} \sqcup \concept{compensation\_event\_detail} \sqcup \concept{signal\_event\_detail} \sqcup \concept{terminate\_event\_detail}$\vspace{0.1cm} \\
\vspace{0.4cm} \hrule \vspace{0.2cm} \noindent \textbf{Class}: $\concept{intermediate\_event}$\vspace{0.2cm} \hrule \vspace{0.2cm} 
\noindent \textbf{Label}: Intermediate\vspace{0.1cm} \\
\noindent \textbf{Description}: Intermediate Events occur between a Start
                    Event and an End Event. It will affect the
		    flow of the process, but will not start or
		    (directly) terminate the process.\vspace{0.1cm} \\
\noindent \stepcounter{idaxiom} $AX\_\arabic{idaxiom}$ $\concept{intermediate\_event} \sqsubseteq (\geq1) \property{has\_intermediate\_event\_target}$\vspace{0.1cm} \\
\par \vspace{-0.1cm} \noindent \textbf{Property}: $\property{has\_intermediate\_event\_trigger}$\vspace{0.1cm} \\
\noindent \textbf{Label}: Trigger\vspace{0.1cm} \\
\noindent \textbf{Description}: Trigger (EventDetail) is an 
	attribute that defines the type of trigger expected for an Intermediate 
	Event. Of the set of EventDetailTypes (see Section B.11.7, Event 
	Details, on page 270), only eight (8) can be applied to an Intermediate 
	Event: Message, Timer, Error, Cancel, Compensation, Conditional, Link, 
	and Signal (see Table 9.8).
	If there is no EventDetail is defined, then this is considered a None 
	Intermediate Event and the Event will not have an internal marker 
	(see Table 9.8).
	If there is more than one EventDetail is defined, this is considered a 
	Multiple Intermediate Event and the Event will have the star internal 
	marker (see Table 9.8).\vspace{0.1cm} \\
\noindent \stepcounter{idaxiom} $AX\_\arabic{idaxiom}$ $\property{has\_intermediate\_event\_trigger} \mbox{ has domain }\concept{intermediate\_event}$\vspace{0.1cm} \\
\noindent \stepcounter{idaxiom} $AX\_\arabic{idaxiom}$ $\property{has\_intermediate\_event\_trigger} \mbox{ has range }\concept{message\_event\_detail} \sqcup \concept{timer\_event\_detail} \sqcup \\ \concept{error\_event\_detail} \sqcup \concept{cancel\_event\_detail} \sqcup \concept{compensation\_event\_detail} \sqcup \concept{conditional\_event\_detail} \sqcup \concept{link\_event\_detail} \sqcup \concept{signal\_event\_detail}$\vspace{0.1cm} \\
\par \vspace{-0.1cm} \noindent \textbf{Property}: $\property{has\_intermediate\_event\_target}$\vspace{0.1cm} \\
\noindent \textbf{Label}: Target\vspace{0.1cm} \\
\noindent \textbf{Description}: A Target MAY be included for the
	Intermediate Event. The Target MUST be an activity (Sub-Process or 
	Task). This means that the Intermediate Event is attached to the 
	boundary of the activity and is used to signify an exception or 
	compensation for that activity.\vspace{0.1cm} \\
\noindent \stepcounter{idaxiom} $AX\_\arabic{idaxiom}$ $\property{has\_intermediate\_event\_target} \mbox{ has domain }\concept{intermediate\_event}$\vspace{0.1cm} \\
\noindent \stepcounter{idaxiom} $AX\_\arabic{idaxiom}$ $\property{has\_intermediate\_event\_target} \mbox{ has range }\concept{activity}$\vspace{0.1cm} \\
\vspace{0.4cm} \hrule \vspace{0.2cm} \noindent \textbf{Class}: $\concept{activity}$\vspace{0.2cm} \hrule \vspace{0.2cm} 
\noindent \textbf{Label}: Activity\vspace{0.1cm} \\
\noindent \textbf{Description}: An activity is a generic term for work that company
         performs. An activity can be atomic or non-atomic (compound). The types
	 of activities that are a part of a Process Model are: Process, 
	 Sub-Process, and Task. Tasks and Sub-Processes are rounded rectangles. 
	 Processes are either unbounded or a contained within a Pool.\vspace{0.1cm} \\
\noindent \stepcounter{idaxiom} $AX\_\arabic{idaxiom}$ $\concept{activity} \equiv \concept{sub\_process} \sqcup \concept{task}$\vspace{0.1cm} \\
\noindent \stepcounter{idaxiom} $AX\_\arabic{idaxiom}$ $\concept{sub\_process} \sqsubseteq  \neg \concept{task}$\vspace{0.1cm} \\
\noindent \stepcounter{idaxiom} $AX\_\arabic{idaxiom}$ $\concept{activity} \sqsubseteq  (=1) \property{has\_activity\_activity\_type}$\vspace{0.1cm} \\
\par \vspace{-0.1cm} \noindent \textbf{Property}: $\property{has\_activity\_activity\_type}$\vspace{0.1cm} \\
\noindent \textbf{Label}: ActivityType\vspace{0.1cm} \\
\noindent \textbf{Description}: The ActivityType MUST be of type 
	Task or Sub-Process.\vspace{0.1cm} \\
\noindent \stepcounter{idaxiom} $AX\_\arabic{idaxiom}$ $\property{has\_activity\_activity\_type} \mbox{ has domain }\concept{activity}$\vspace{0.1cm} \\
\noindent \stepcounter{idaxiom} $AX\_\arabic{idaxiom}$ $\property{has\_activity\_activity\_type} \mbox{ has range }\concept{activity\_types}$\vspace{0.1cm} \\
\vspace{0.4cm} \hrule \vspace{0.2cm} \noindent \textbf{Class}: $\concept{activity\_types}$\vspace{0.2cm} \hrule \vspace{0.2cm} 
\noindent \textbf{Label}: Activity Types\vspace{0.1cm} \\
\noindent \stepcounter{idaxiom} $AX\_\arabic{idaxiom}$ $\concept{activity\_types} \equiv \{\instance{task\_activity\_type}, \instance{sub\_process\_activity\_type}\}$\vspace{0.1cm} \\
\par \vspace{-0.1cm} \noindent \textbf{Instance}: $\instance{task\_activity\_type}$\vspace{0.1cm} \\
\noindent \textbf{Label}: task\vspace{0.1cm} \\
\par \vspace{-0.1cm} \noindent \textbf{Instance}: $\instance{sub\_process\_activity\_type}$\vspace{0.1cm} \\
\noindent \textbf{Label}: sub\_process\vspace{0.1cm} \\
\noindent \stepcounter{idaxiom} $AX\_\arabic{idaxiom}$ $( \neg \{\instance{task\_activity\_type}\})(\instance{sub\_process\_activity\_type})$\vspace{0.1cm} \\
\noindent \stepcounter{idaxiom} $AX\_\arabic{idaxiom}$ $\concept{task} \equiv \concept{activity} \sqcap \exists\property{has\_activity\_activity\_type}.\{\instance{task\_activity\_type}\}$\vspace{0.1cm} \\
\noindent \stepcounter{idaxiom} $AX\_\arabic{idaxiom}$ $\concept{sub\_process} \equiv \concept{activity} \sqcap \exists\property{has\_activity\_activity\_type}.\{\instance{sub\_process\_activity\_type}\}$\vspace{0.1cm} \\
\noindent \stepcounter{idaxiom} $AX\_\arabic{idaxiom}$ $\concept{activity} \sqsubseteq  (=1) \property{has\_activity\_status}$\vspace{0.1cm} \\
\par \vspace{-0.1cm} \noindent \textbf{Property}: $\property{has\_activity\_status}$\vspace{0.1cm} \\
\noindent \textbf{Label}: Status\vspace{0.1cm} \\
\noindent \textbf{Description}: The Status of an activity is determined 
	when the activity is being executed by a process engine. The Status of 
	an activity can be used within Assignment Expressions.\vspace{0.1cm} \\
\noindent \stepcounter{idaxiom} $AX\_\arabic{idaxiom}$ $\property{has\_activity\_status} \mbox{ has domain }\concept{activity}$\vspace{0.1cm} \\
\noindent \stepcounter{idaxiom} $AX\_\arabic{idaxiom}$ $\property{has\_activity\_status}$ $\mbox{ has range }$ $\datatype{xsd:string}\{"\datainstance{None}", "\datainstance{Ready}", "\datainstance{Active}", "\datainstance{Cancelled}", "\datainstance{Aborting}", "\datainstance{Aborted}", \\"\datainstance{Completing}", "\datainstance{Completed}"\}$\vspace{0.1cm} \\
\par \vspace{-0.1cm} \noindent \textbf{Property}: $\property{has\_activity\_performers}$\vspace{0.1cm} \\
\noindent \textbf{Label}: Performers\vspace{0.1cm} \\
\noindent \textbf{Description}: One or more Performers MAY be entered.
	The Performers attribute defines the resource that will be responsible 
	for the activity. The Performers entry could be in the form of a 
	specific individual, a group, an organization role or position, or an
	organization.\vspace{0.1cm} \\
\noindent \stepcounter{idaxiom} $AX\_\arabic{idaxiom}$ $\property{has\_activity\_performers} \mbox{ has domain }\concept{activity}$\vspace{0.1cm} \\
\noindent \stepcounter{idaxiom} $AX\_\arabic{idaxiom}$ $\property{has\_activity\_performers} \mbox{ has range }\datatype{xsd:string}$\vspace{0.1cm} \\
\par \vspace{-0.1cm} \noindent \textbf{Property}: $\property{has\_activity\_properties}$\vspace{0.1cm} \\
\noindent \textbf{Label}: Properties\vspace{0.1cm} \\
\noindent \textbf{Description}: Modeler-defined Properties MAY be 
	added to a activity. These Properties are "local" to the activity. All 
	Tasks, Sub-activity objects, and Sub-activityes that are embedded SHALL 
	have access to these Properties. The fully delineated name of these 
	properties is "activity name.property name" (e.g., "Add 
	Customer.Customer Name"). Further details about the definition of a 
	Property can be found in "Property on page 276."\vspace{0.1cm} \\
\noindent \stepcounter{idaxiom} $AX\_\arabic{idaxiom}$ $\property{has\_activity\_properties} \mbox{ has domain }\concept{activity}$\vspace{0.1cm} \\
\noindent \stepcounter{idaxiom} $AX\_\arabic{idaxiom}$ $\property{has\_activity\_properties} \mbox{ has range }\concept{property}$\vspace{0.1cm} \\
\par \vspace{-0.1cm} \noindent \textbf{Property}: $\property{has\_activity\_input\_sets}$\vspace{0.1cm} \\
\noindent \textbf{Label}: Input set\vspace{0.1cm} \\
\noindent \textbf{Description}: The InputSets attribute defines the 
	data requirements for input to the activity. Zero or more InputSets MAY
	be defined. Each Input set is sufficient to allow the activity to be 
	performed (if it has first been instantiated by the appropriate signal
	arriving from an incoming Sequence Flow). Further details about the 
	definition of an Input- Set can be found in Section B.11.10, "InputSet,"
	on page 274.\vspace{0.1cm} \\
\noindent \stepcounter{idaxiom} $AX\_\arabic{idaxiom}$ $\property{has\_activity\_input\_sets} \mbox{ has domain }\concept{activity}$\vspace{0.1cm} \\
\noindent \stepcounter{idaxiom} $AX\_\arabic{idaxiom}$ $\property{has\_activity\_input\_sets} \mbox{ has range }\concept{input\_set}$\vspace{0.1cm} \\
\par \vspace{-0.1cm} \noindent \textbf{Property}: $\property{has\_activity\_output\_sets}$\vspace{0.1cm} \\
\noindent \textbf{Label}: Output set\vspace{0.1cm} \\
\noindent \textbf{Description}: The OutputSets attribute defines the 
	data requirements for output from the activity. Zero or more OutputSets
	MAY be defined. At the completion of the activity, only one of the 
	OutputSets may be produced--It is up to the implementation of the 
	activity to determine which set will be produced. However, the IORules 
	attribute MAY indicate a relationship between an OutputSet and an 
	InputSet that started the activity. Further details about the definition
	of an OutputSet can be found in Section B.11.13, "OutputSet," on page 
	275.\vspace{0.1cm} \\
\noindent \stepcounter{idaxiom} $AX\_\arabic{idaxiom}$ $\property{has\_activity\_output\_sets} \mbox{ has domain }\concept{activity}$\vspace{0.1cm} \\
\noindent \stepcounter{idaxiom} $AX\_\arabic{idaxiom}$ $\property{has\_activity\_output\_sets} \mbox{ has range }\concept{output\_set}$\vspace{0.1cm} \\
\par \vspace{-0.1cm} \noindent \textbf{Property}: $\property{has\_activity\_IO\_rules}$\vspace{0.1cm} \\
\noindent \textbf{Label}: IO Rules\vspace{0.1cm} \\
\noindent \textbf{Description}: The IORules attribute is a collection of
	expressions, each of which specifies the required relationship between
	one input and one output. That is, if the activity is instantiated with 
	a specified input, that activity shall complete with the specified 
	output.\vspace{0.1cm} \\
\noindent \stepcounter{idaxiom} $AX\_\arabic{idaxiom}$ $\property{has\_activity\_IO\_rules} \mbox{ has domain }\concept{activity}$\vspace{0.1cm} \\
\noindent \stepcounter{idaxiom} $AX\_\arabic{idaxiom}$ $\property{has\_activity\_IO\_rules} \mbox{ has range }\concept{expression}$\vspace{0.1cm} \\
\noindent \stepcounter{idaxiom} $AX\_\arabic{idaxiom}$ $\concept{activity} \sqsubseteq  (=1) \property{has\_activity\_start\_quantity}$\vspace{0.1cm} \\
\par \vspace{-0.1cm} \noindent \textbf{Property}: $\property{has\_activity\_start\_quantity}$\vspace{0.1cm} \\
\noindent \textbf{Label}: StartQuantity\vspace{0.1cm} \\
\noindent \textbf{Description}: The default value is 1. The value 
	MUST NOT be less than 1. This attribute defines the number of Tokens 
	that must arrive before the activity can begin.\vspace{0.1cm} \\
\noindent \stepcounter{idaxiom} $AX\_\arabic{idaxiom}$ $\property{has\_activity\_start\_quantity} \mbox{ has domain }\concept{activity}$\vspace{0.1cm} \\
\noindent \stepcounter{idaxiom} $AX\_\arabic{idaxiom}$ $\property{has\_activity\_start\_quantity} \mbox{ has range }\datatype{xsd:positiveInteger}$\vspace{0.1cm} \\
\noindent \stepcounter{idaxiom} $AX\_\arabic{idaxiom}$ $\concept{activity} \sqsubseteq  (=1) \property{has\_activity\_completion\_quantity}$\vspace{0.1cm} \\
\par \vspace{-0.1cm} \noindent \textbf{Property}: $\property{has\_activity\_completion\_quantity}$\vspace{0.1cm} \\
\noindent \textbf{Label}: CompletionQuantity\vspace{0.1cm} \\
\noindent \textbf{Description}: The default value is 1. The 
	value MUST NOT be less than 1. This attribute defines the number of 
	Tokens that must be generated from the activity. This number of Tokens
	will be sent done any outgoing Sequence Flow (assuming any Sequence Flow
	Conditions are satisfied).\vspace{0.1cm} \\
\noindent \stepcounter{idaxiom} $AX\_\arabic{idaxiom}$ $\property{has\_activity\_completion\_quantity} \mbox{ has domain }\concept{activity}$\vspace{0.1cm} \\
\noindent \stepcounter{idaxiom} $AX\_\arabic{idaxiom}$ $\property{has\_activity\_completion\_quantity} \mbox{ has range }\datatype{xsd:positiveInteger}$\vspace{0.1cm} \\
\noindent \stepcounter{idaxiom} $AX\_\arabic{idaxiom}$ $\concept{activity} \sqsubseteq (\geq1) \property{has\_activity\_loop\_type}$\vspace{0.1cm} \\
\par \vspace{-0.1cm} \noindent \textbf{Property}: $\property{has\_activity\_loop\_type}$\vspace{0.1cm} \\
\noindent \textbf{Label}: LoopType\vspace{0.1cm} \\
\noindent \textbf{Description}: LoopType is an attribute and is by 
	default None, but MAY be set to Standard or MultiInstance. If so, the 
	Loop marker SHALL be placed at the bottom center of the activity shape 
	(see Figure 9.6 and Figure 9.15).
	A Task of type Receive that has its Instantiate attribute set to True 
	MUST NOT have a Standard or MultiInstance LoopType.\vspace{0.1cm} \\
\noindent \stepcounter{idaxiom} $AX\_\arabic{idaxiom}$ $\property{has\_activity\_loop\_type} \mbox{ has domain }\concept{activity}$\vspace{0.1cm} \\
\noindent \stepcounter{idaxiom} $AX\_\arabic{idaxiom}$ $\property{has\_activity\_loop\_type} \mbox{ has range }\concept{loop\_types}$\vspace{0.1cm} \\
\vspace{0.4cm} \hrule \vspace{0.2cm} \noindent \textbf{Class}: $\concept{loop\_types}$\vspace{0.2cm} \hrule \vspace{0.2cm} 
\noindent \textbf{Label}: Loop Types\vspace{0.1cm} \\
\noindent \stepcounter{idaxiom} $AX\_\arabic{idaxiom}$ $\concept{loop\_types} \equiv \{\instance{standard}, \instance{multi\_instance}\}$\vspace{0.1cm} \\
\par \vspace{-0.1cm} \noindent \textbf{Instance}: $\instance{standard}$\vspace{0.1cm} \\
\noindent \textbf{Label}: standard\vspace{0.1cm} \\
\par \vspace{-0.1cm} \noindent \textbf{Instance}: $\instance{multi\_instance}$\vspace{0.1cm} \\
\noindent \textbf{Label}: multi\_instance\vspace{0.1cm} \\
\noindent \stepcounter{idaxiom} $AX\_\arabic{idaxiom}$ $( \neg \{\instance{standard}\})(\instance{multi\_instance})$\vspace{0.1cm} \\
\noindent \stepcounter{idaxiom} $AX\_\arabic{idaxiom}$ $\concept{standard\_loop\_activity} \equiv \concept{activity} \sqcap \exists\property{has\_activity\_loop\_type}.\{\instance{standard}\}$\vspace{0.1cm} \\
\noindent \stepcounter{idaxiom} $AX\_\arabic{idaxiom}$ $\concept{multi\_instance\_loop\_activity} \equiv \concept{activity} \sqcap \exists\property{has\_activity\_loop\_type}.\{\instance{multi\_instance}\}$\vspace{0.1cm} \\
\vspace{0.4cm} \hrule \vspace{0.2cm} \noindent \textbf{Class}: $\concept{standard\_loop\_activity}$\vspace{0.2cm} \hrule \vspace{0.2cm} 
\noindent \textbf{Label}: Standard Loop Activity\vspace{0.1cm} \\
\noindent \textbf{Description}: An activity is a generic term for work 
	that company performs. An activity can be atomic or non-atomic 
	(compound). The types of activities that are a part of a Process Model 
	are: Process, Sub-Process, and Task. Tasks and Sub-Processes are rounded
	rectangles. Processes are either unbounded or a contained within a 
	Pool.\vspace{0.1cm} \\
\noindent \stepcounter{idaxiom} $AX\_\arabic{idaxiom}$ $\concept{standard\_loop\_activity} \sqsubseteq  (=1) \property{has\_standard\_loop\_activity\_loop\_condition}$\vspace{0.1cm} \\
\par \vspace{-0.1cm} \noindent \textbf{Property}: $\property{has\_standard\_loop\_activity\_loop\_condition}$\vspace{0.1cm} \\
\noindent \textbf{Label}: Loop Condition\vspace{0.1cm} \\
\noindent \textbf{Description}: Standard Loops MUST 
	have a boolean Expression to be evaluated, plus the timing when the 
	expression SHALL be evaluated. The attributes of an Expression can be
	found in "Expression on page 273."\vspace{0.1cm} \\
\noindent \stepcounter{idaxiom} $AX\_\arabic{idaxiom}$ $\property{has\_standard\_loop\_activity\_loop\_condition} \mbox{ has domain }\concept{standard\_loop\_activity}$\vspace{0.1cm} \\
\noindent \stepcounter{idaxiom} $AX\_\arabic{idaxiom}$ $\property{has\_standard\_loop\_activity\_loop\_condition} \mbox{ has range }\concept{expression}$\vspace{0.1cm} \\
\noindent \stepcounter{idaxiom} $AX\_\arabic{idaxiom}$ $\concept{standard\_loop\_activity} \sqsubseteq  (=1) \property{has\_standard\_loop\_activity\_loop\_counter}$\vspace{0.1cm} \\
\par \vspace{-0.1cm} \noindent \textbf{Property}: $\property{has\_standard\_loop\_activity\_loop\_counter}$\vspace{0.1cm} \\
\noindent \textbf{Label}: Loop Counter\vspace{0.1cm} \\
\noindent \textbf{Description}: The LoopCounter 
	attribute is used at runtime to count the number of loops and is
	automatically updated by the process engine. The LoopCounter attribute 
	MUST be incremented at the start of a loop. The modeler may use the
	attribute in the LoopCondition Expression.\vspace{0.1cm} \\
\noindent \stepcounter{idaxiom} $AX\_\arabic{idaxiom}$ $\property{has\_standard\_loop\_activity\_loop\_counter} \mbox{ has domain }\concept{standard\_loop\_activity}$\vspace{0.1cm} \\
\noindent \stepcounter{idaxiom} $AX\_\arabic{idaxiom}$ $\property{has\_standard\_loop\_activity\_loop\_counter} \mbox{ has range }\datatype{xsd:int}$\vspace{0.1cm} \\
\noindent \stepcounter{idaxiom} $AX\_\arabic{idaxiom}$ $\concept{standard\_loop\_activity} \sqsubseteq (\geq1) \property{has\_standard\_loop\_activity\_loop\_maximum}$\vspace{0.1cm} \\
\par \vspace{-0.1cm} \noindent \textbf{Property}: $\property{has\_standard\_loop\_activity\_loop\_maximum}$\vspace{0.1cm} \\
\noindent \textbf{Label}: Loop Maximum\vspace{0.1cm} \\
\noindent \textbf{Description}: The Maximum an 
	optional attribute that provides is a simple way to add a cap to the
	number of loops. This SHALL be added to the Expression defined in the
	LoopCondition.\vspace{0.1cm} \\
\noindent \stepcounter{idaxiom} $AX\_\arabic{idaxiom}$ $\property{has\_standard\_loop\_activity\_loop\_maximum} \mbox{ has domain }\concept{standard\_loop\_activity}$\vspace{0.1cm} \\
\noindent \stepcounter{idaxiom} $AX\_\arabic{idaxiom}$ $\property{has\_standard\_loop\_activity\_loop\_maximum} \mbox{ has range }\datatype{xsd:int}$\vspace{0.1cm} \\
\noindent \stepcounter{idaxiom} $AX\_\arabic{idaxiom}$ $\concept{standard\_loop\_activity} \sqsubseteq (\geq1) \property{has\_standard\_loop\_activity\_test\_time}$\vspace{0.1cm} \\
\par \vspace{-0.1cm} \noindent \textbf{Property}: $\property{has\_standard\_loop\_activity\_test\_time}$\vspace{0.1cm} \\
\noindent \textbf{Label}: Test Time\vspace{0.1cm} \\
\noindent \textbf{Description}: The expressions that are 
	evaluated Before the activity begins are equivalent to a programming 
	while function.
	The expression that are evaluated After the activity finishes are 
	equivalent to a programming until function.\vspace{0.1cm} \\
\noindent \stepcounter{idaxiom} $AX\_\arabic{idaxiom}$ $\property{has\_standard\_loop\_activity\_test\_time} \mbox{ has domain }\concept{standard\_loop\_activity}$\vspace{0.1cm} \\
\noindent \stepcounter{idaxiom} $AX\_\arabic{idaxiom}$ $\property{has\_standard\_loop\_activity\_test\_time} \mbox{ has range }\datatype{xsd:string}\{"\datainstance{Before}","\datainstance{After}"\}$\vspace{0.1cm} \\
\vspace{0.4cm} \hrule \vspace{0.2cm} \noindent \textbf{Class}: $\concept{multi\_instance\_loop\_activity}$\vspace{0.2cm} \hrule \vspace{0.2cm} 
\noindent \textbf{Label}: Multi Instance Loop Activity\vspace{0.1cm} \\
\noindent \textbf{Description}: An activity is a generic term 
	for work that company performs. An activity can be atomic or non-atomic
	(compound). The types of activities that are a part of a Process Model 
	are: Process, Sub-Process, and Task. Tasks and Sub-Processes are 
	rounded rectangles. Processes are either unbounded or a contained 
	within a Pool.\vspace{0.1cm} \\
\noindent \stepcounter{idaxiom} $AX\_\arabic{idaxiom}$ $\concept{multi\_instance\_loop\_activity} \sqsubseteq  (=1) \property{has\_multi\_instance\_loop\_activity\_MI\_condition}$\vspace{0.1cm} \\
\par \vspace{-0.1cm} \noindent \textbf{Property}: $\property{has\_multi\_instance\_loop\_activity\_MI\_condition}$\vspace{0.1cm} \\
\noindent \textbf{Label}: MI\_Condition\vspace{0.1cm} \\
\noindent \textbf{Description}: MultiInstance 
	Loops MUST have a numeric Expression to be evaluated--the Expression 
	MUST resolve to an integer. The attributes of an Expression can be
	found in "Expression on page 273."\vspace{0.1cm} \\
\noindent \stepcounter{idaxiom} $AX\_\arabic{idaxiom}$ $\property{has\_multi\_instance\_loop\_activity\_MI\_condition} \mbox{ has domain }\concept{multi\_instance\_loop\_activity}$\vspace{0.1cm} \\
\noindent \stepcounter{idaxiom} $AX\_\arabic{idaxiom}$ $\property{has\_multi\_instance\_loop\_activity\_MI\_condition} \mbox{ has range }\concept{expression}$\vspace{0.1cm} \\
\noindent \stepcounter{idaxiom} $AX\_\arabic{idaxiom}$ $\concept{multi\_instance\_loop\_activity} \sqsubseteq  (=1) \property{has\_multi\_instance\_loop\_activity\_loop\_counter}$\vspace{0.1cm} \\
\par \vspace{-0.1cm} \noindent \textbf{Property}: $\property{has\_multi\_instance\_loop\_activity\_loop\_counter}$\vspace{0.1cm} \\
\noindent \textbf{Label}: Loop Counter\vspace{0.1cm} \\
\noindent \textbf{Description}: The LoopCounter 
	attribute is only applied for Sequential MultiInstance Loops and for 
	processes that are being executed by a process engine. The attribute is 
	updated at runtime by a process engine to count the number of loops as 
	they occur. The LoopCounter attribute MUST be incremented at the start 
	of a loop. Unlike a Standard loop, the modeler does not use this 
	attribute in the MI\_Condition Expression, but it can be used for 
	tracking the status of a loop.\vspace{0.1cm} \\
\noindent \stepcounter{idaxiom} $AX\_\arabic{idaxiom}$ $\property{has\_multi\_instance\_loop\_activity\_loop\_counter} \mbox{ has domain }\concept{multi\_instance\_loop\_activity}$\vspace{0.1cm} \\
\noindent \stepcounter{idaxiom} $AX\_\arabic{idaxiom}$ $\property{has\_multi\_instance\_loop\_activity\_loop\_counter} \mbox{ has range }\datatype{xsd:int}$\vspace{0.1cm} \\
\noindent \stepcounter{idaxiom} $AX\_\arabic{idaxiom}$ $\concept{multi\_instance\_loop\_activity} \sqsubseteq  (=1) \property{has\_multi\_instance\_loop\_activity\_MI\_ordering}$\vspace{0.1cm} \\
\par \vspace{-0.1cm} \noindent \textbf{Property}: $\property{has\_multi\_instance\_loop\_activity\_MI\_ordering}$\vspace{0.1cm} \\
\noindent \textbf{Label}: MI\_ordering\vspace{0.1cm} \\
\noindent \textbf{Description}: This applies to 
	only MultiInstance Loops. The MI\_Ordering attribute defines whether the
	loop instances will be performed sequentially or in parallel.
	Sequential MI\_Ordering is a more traditional loop.
	Parallel MI\_Ordering is equivalent to multi-instance specifications 
	that other notations, such as UML Activity Diagrams use. If set to 
	Parallel, the Parallel marker SHALL replace the Loop Marker at the 
	bottom center of the activity shape (see Figure 9.9 and Figure 9.15).\vspace{0.1cm} \\
\noindent \stepcounter{idaxiom} $AX\_\arabic{idaxiom}$ $\property{has\_multi\_instance\_loop\_activity\_MI\_ordering} \mbox{ has domain }\concept{multi\_instance\_loop\_activity}$\vspace{0.1cm} \\
\noindent \stepcounter{idaxiom} $AX\_\arabic{idaxiom}$ $\property{has\_multi\_instance\_loop\_activity\_MI\_ordering} \mbox{ has range }\datatype{xsd:string}\{"\datainstance{Parallel}","\datainstance{Sequential}"\}$\vspace{0.1cm} \\
\noindent \stepcounter{idaxiom} $AX\_\arabic{idaxiom}$ $\concept{multi\_instance\_loop\_activity} \sqsubseteq ( \neg \exists\property{has\_multi\_instance\_loop\_activity\_MI\_ordering}.\{"\datainstance{Parallel}"\}) \sqcup ((\exists\property{has\_multi\_instance\_loop\_activity\_MI\_ordering}.\{"\datainstance{Parallel}"\}) \sqcap  (=1) \property{has\_multi\_instance\_loop\_activity\_MI\_flow\_condition})$\vspace{0.1cm} \\
\par \vspace{-0.1cm} \noindent \textbf{Property}: $\property{has\_multi\_instance\_loop\_activity\_MI\_flow\_condition}$\vspace{0.1cm} \\
\noindent \textbf{Label}: MI\_FlowCondition\vspace{0.1cm} \\
\noindent \textbf{Description}: This 
	attribute is equivalent to using a Gateway to control the flow past a 
	set of parallel paths.
	- An MI\_FlowCondition of "None" is the same as uncontrolled flow (no
	Gateway) and means that all activity instances SHALL generate a token 
	that will continue when that instance is completed.
	- An MI\_FlowCondition of "One" is the same as an Exclusive Gateway and
	means that the Token SHALL continue past the activity after only one of
	the activity instances has completed. The activity will continue its 
	other instances, but additional Tokens MUST NOT be passed from the 
	activity.
	- An MI\_FlowCondition of "All" is the same as a Parallel Gateway and 
	means that the Token SHALL continue past the activity after all of the 
	activity instances have completed.
	- An MI\_FlowCondition of "Complex" is similar to that of a Complex
	Gateway. The ComplexMI\_FlowCondition attribute will determine the Token 
	flow.\vspace{0.1cm} \\
\noindent \stepcounter{idaxiom} $AX\_\arabic{idaxiom}$ $\property{has\_multi\_instance\_loop\_activity\_MI\_flow\_condition} \mbox{ has domain }\concept{multi\_instance\_loop\_activity}$\vspace{0.1cm} \\
\noindent \stepcounter{idaxiom} $AX\_\arabic{idaxiom}$ $\property{has\_multi\_instance\_loop\_activity\_MI\_flow\_condition} \mbox{ has range }\datatype{xsd:string}\{"\datainstance{None}","\datainstance{One}","\datainstance{All}","\datainstance{Complex}"\}$\vspace{0.1cm} \\
\noindent \stepcounter{idaxiom} $AX\_\arabic{idaxiom}$ $\concept{multi\_instance\_loop\_activity} \sqsubseteq ( \neg \exists\property{has\_multi\_instance\_loop\_activity\_MI\_flow\_condition}.\{"\datainstance{Complex}"\}) \sqcup ((\exists\property{has\_multi\_instance\_loop\_activity\_MI\_flow\_condition}.\{"\datainstance{Complex}"\}) \sqcap \\ (=1) \property{has\_multi\_instance\_loop\_activity\_complex\_MI\_flow\_condition})$\vspace{0.1cm} \\
\par \vspace{-0.1cm} \noindent \textbf{Property}: $\property{has\_multi\_instance\_loop\_activity\_complex\_MI\_flow\_condition}$\vspace{0.1cm} \\
\noindent \textbf{Label}: ComplexMI\_FlowCcondition\vspace{0.1cm} \\
\noindent \textbf{Description}: If
	the MI\_FlowCondition attribute is set to "Complex," then an Expression 
	Must be entered. This Expression that MAY reference Process data. The 
	expression will be evaluated after each iteration of the Activity and 
	SHALL resolve to a boolean. If the result of the expression evaluation 
	is TRUE, then a Token will be sent down the activity's outgoing 
	Sequence Flow. Otherwise, no Token will be sent. The attributes of an 
	Expression can be found in "Expression on page 273."\vspace{0.1cm} \\
\noindent \stepcounter{idaxiom} $AX\_\arabic{idaxiom}$ $\property{has\_multi\_instance\_loop\_activity\_complex\_MI\_flow\_condition} \mbox{ has domain }\concept{multi\_instance\_loop\_activity}$\vspace{0.1cm} \\
\noindent \stepcounter{idaxiom} $AX\_\arabic{idaxiom}$ $\property{has\_multi\_instance\_loop\_activity\_complex\_MI\_flow\_condition} \mbox{ has range }\concept{expression}$\vspace{0.1cm} \\
\vspace{0.4cm} \hrule \vspace{0.2cm} \noindent \textbf{Class}: $\concept{sub\_process}$\vspace{0.2cm} \hrule \vspace{0.2cm} 
\noindent \textbf{Label}: Sub-process\vspace{0.1cm} \\
\noindent \textbf{Description}: A Sub-Process is a compound activity
	               that is included within a Process. It is
                       compound in that it can be broken down
                       into a finer level of detail (a Process)
                       through a set of sub-activities.\vspace{0.1cm} \\
\noindent \stepcounter{idaxiom} $AX\_\arabic{idaxiom}$ $\concept{sub\_process} \sqsubseteq  (=1) \property{has\_sub\_process\_sub\_process\_type}$\vspace{0.1cm} \\
\par \vspace{-0.1cm} \noindent \textbf{Property}: $\property{has\_sub\_process\_sub\_process\_type}$\vspace{0.1cm} \\
\noindent \textbf{Label}: SubProcessType\vspace{0.1cm} \\
\noindent \textbf{Description}: SubProcessType is an 
	attribute that defines whether the Sub-Process details are embedded with
	in the higher level Process or refers to another, re-usable Process.
	The default is Embedded.\vspace{0.1cm} \\
\noindent \stepcounter{idaxiom} $AX\_\arabic{idaxiom}$ $\property{has\_sub\_process\_sub\_process\_type} \mbox{ has domain }\concept{sub\_process}$\vspace{0.1cm} \\
\noindent \stepcounter{idaxiom} $AX\_\arabic{idaxiom}$ $\property{has\_sub\_process\_sub\_process\_type} \mbox{ has range }\concept{sub\_process\_types}$\vspace{0.1cm} \\
\noindent \stepcounter{idaxiom} $AX\_\arabic{idaxiom}$ $\concept{sub\_process\_types} \equiv \{\instance{embedded}, \instance{reusable}, \instance{reference}\}$\vspace{0.1cm} \\
\par \vspace{-0.1cm} \noindent \textbf{Instance}: $\instance{embedded}$\vspace{0.1cm} \\
\noindent \textbf{Label}: Embedded\vspace{0.1cm} \\
\par \vspace{-0.1cm} \noindent \textbf{Instance}: $\instance{reusable}$\vspace{0.1cm} \\
\noindent \textbf{Label}: Reusable\vspace{0.1cm} \\
\par \vspace{-0.1cm} \noindent \textbf{Instance}: $\instance{reference}$\vspace{0.1cm} \\
\noindent \textbf{Label}: Reference\vspace{0.1cm} \\
\noindent \stepcounter{idaxiom} $AX\_\arabic{idaxiom}$ $\concept{embedded\_sub\_process} \equiv \concept{sub\_process} \sqcap \exists\property{has\_sub\_process\_sub\_process\_type}.\{\instance{embedded}\}$\vspace{0.1cm} \\
\noindent \stepcounter{idaxiom} $AX\_\arabic{idaxiom}$ $\concept{reusable\_sub\_process} \equiv \concept{sub\_process} \sqcap \exists\property{has\_sub\_process\_sub\_process\_type}.\{\instance{reusable}\}$\vspace{0.1cm} \\
\noindent \stepcounter{idaxiom} $AX\_\arabic{idaxiom}$ $\concept{reference\_sub\_process} \equiv \concept{sub\_process} \sqcap \exists\property{has\_sub\_process\_sub\_process\_type}.\{\instance{reference}\}$\vspace{0.1cm} \\
\noindent \stepcounter{idaxiom} $AX\_\arabic{idaxiom}$ $\concept{embedded\_sub\_process} \sqsubseteq  \neg \concept{reusable\_sub\_process}$\vspace{0.1cm} \\
\noindent \stepcounter{idaxiom} $AX\_\arabic{idaxiom}$ $\concept{embedded\_sub\_process} \sqsubseteq  \neg \concept{reference\_sub\_process}$\vspace{0.1cm} \\
\noindent \stepcounter{idaxiom} $AX\_\arabic{idaxiom}$ $\concept{reusable\_sub\_process} \sqsubseteq  \neg \concept{reference\_sub\_process}$\vspace{0.1cm} \\
\noindent \stepcounter{idaxiom} $AX\_\arabic{idaxiom}$ $\concept{sub\_process} \sqsubseteq  (=1) \property{has\_sub\_process\_is\_a\_transaction}$\vspace{0.1cm} \\
\par \vspace{-0.1cm} \noindent \textbf{Property}: $\property{has\_sub\_process\_is\_a\_transaction}$\vspace{0.1cm} \\
\noindent \textbf{Label}: IsATransaction\vspace{0.1cm} \\
\noindent \textbf{Description}: TIsATransaction determines 
	whether or not the behavior of the Sub-Process will follow the behavior
	of a Transaction (see "Sub-Process Behavior as a Transaction on page 
	62.")\vspace{0.1cm} \\
\noindent \stepcounter{idaxiom} $AX\_\arabic{idaxiom}$ $\property{has\_sub\_process\_is\_a\_transaction} \mbox{ has domain }\concept{sub\_process}$\vspace{0.1cm} \\
\noindent \stepcounter{idaxiom} $AX\_\arabic{idaxiom}$ $\property{has\_sub\_process\_is\_a\_transaction} \mbox{ has range }\datatype{xsd:boolean}$\vspace{0.1cm} \\
\noindent \stepcounter{idaxiom} $AX\_\arabic{idaxiom}$ $\concept{sub\_process} \sqsubseteq ((\exists\property{has\_sub\_process\_is\_a\_transaction}.\{"\datainstance{false}"\}) \sqcap ( (=0) \property{has\_sub\_process\_sub\_transaction\_ref})) \sqcup ((\exists\property{has\_sub\_process\_is\_a\_transaction}.\{"\datainstance{true}"\}) \sqcap ( (=1) \property{has\_sub\_process\_sub\_transaction\_ref}))$\vspace{0.1cm} \\
\par \vspace{-0.1cm} \noindent \textbf{Property}: $\property{has\_sub\_process\_sub\_transaction\_ref}$\vspace{0.1cm} \\
\noindent \textbf{Label}: Transaction\_Ref\vspace{0.1cm} \\
\noindent \textbf{Description}: If the IsATransaction 
	attribute is False, then a Transaction MUST NOT be identified. If the 
	IsATransaction attribute is True, then a Transaction MUST be identified.
	The attributes of a Transaction can be found in "Transaction on page 
	277". 
	Note that Transactions that are in different Pools and are connected 
	through Message Flow MUST have the same TransactionId.\vspace{0.1cm} \\
\noindent \stepcounter{idaxiom} $AX\_\arabic{idaxiom}$ $\property{has\_sub\_process\_sub\_transaction\_ref} \mbox{ has domain }\concept{sub\_process}$\vspace{0.1cm} \\
\noindent \stepcounter{idaxiom} $AX\_\arabic{idaxiom}$ $\property{has\_sub\_process\_sub\_transaction\_ref} \mbox{ has range }\concept{transaction}$\vspace{0.1cm} \\
\vspace{0.4cm} \hrule \vspace{0.2cm} \noindent \textbf{Class}: $\concept{embedded\_sub\_process}$\vspace{0.2cm} \hrule \vspace{0.2cm} 
\noindent \textbf{Label}: Embedded Sub-process\vspace{0.1cm} \\
\noindent \textbf{Description}: \vspace{0.1cm} \\
\par \vspace{-0.1cm} \noindent \textbf{Property}: $\property{has\_embedded\_sub\_process\_sub\_graphical\_elements}$\vspace{0.1cm} \\
\noindent \textbf{Label}: GraphicalElements\vspace{0.1cm} \\
\noindent \textbf{Description}: The 
	GraphicalElements attribute identifies all of the objects (e.g., 
	Events, Activities, Gateways, and Artifacts) that are contained within 
	the Embedded Sub-Process.\vspace{0.1cm} \\
\noindent \stepcounter{idaxiom} $AX\_\arabic{idaxiom}$ $\property{has\_embedded\_sub\_process\_sub\_graphical\_elements} \mbox{ has domain }\concept{embedded\_sub\_process}$\vspace{0.1cm} \\
\noindent \stepcounter{idaxiom} $AX\_\arabic{idaxiom}$ $\property{has\_embedded\_sub\_process\_sub\_graphical\_elements} \mbox{ has range }\concept{graphical\_element}$\vspace{0.1cm} \\
\noindent \stepcounter{idaxiom} $AX\_\arabic{idaxiom}$ $\concept{embedded\_sub\_process} \sqsubseteq  (=1) \property{has\_embedded\_sub\_process\_ad\_hoc}$\vspace{0.1cm} \\
\par \vspace{-0.1cm} \noindent \textbf{Property}: $\property{has\_embedded\_sub\_process\_ad\_hoc}$\vspace{0.1cm} \\
\noindent \textbf{Label}: Ad\_hoc\vspace{0.1cm} \\
\noindent \textbf{Description}: AdHoc is a boolean attribute, which has a default of False. This specifies whether
	the embedded\_sub\_process is Ad Hoc or not. The activities within an Ad 
	Hoc embedded\_sub\_process are not controlled or sequenced in a particular
	order, their performance is determined by the performers of the 
	activities. If set to True, then the Ad Hoc marker SHALL be placed at 
	the bottom center of the embedded\_sub\_process or the 
	Sub-embedded\_sub\_process shape for Ad Hoc embedded\_sub\_processes.\vspace{0.1cm} \\
\noindent \stepcounter{idaxiom} $AX\_\arabic{idaxiom}$ $\property{has\_embedded\_sub\_process\_ad\_hoc} \mbox{ has domain }\concept{embedded\_sub\_process}$\vspace{0.1cm} \\
\noindent \stepcounter{idaxiom} $AX\_\arabic{idaxiom}$ $\property{has\_embedded\_sub\_process\_ad\_hoc} \mbox{ has range }\datatype{xsd:boolean}$\vspace{0.1cm} \\
\noindent \stepcounter{idaxiom} $AX\_\arabic{idaxiom}$ $\concept{embedded\_sub\_process} \sqsubseteq (\exists\property{has\_embedded\_sub\_process\_ad\_hoc}.\{"\datainstance{false}"\}) \sqcup \\ (\exists\property{has\_embedded\_sub\_process\_ad\_hoc}.\{"\datainstance{true}"\} \sqcap  (=1) \property{has\_embedded\_sub\_process\_ad\_hoc\_ordering} \sqcap \\ (=1) \property{has\_embedded\_sub\_process\_ad\_hoc\_completion\_condition})$\vspace{0.1cm} \\
\par \vspace{-0.1cm} \noindent \textbf{Property}: $\property{has\_embedded\_sub\_process\_ad\_hoc\_ordering}$\vspace{0.1cm} \\
\noindent \textbf{Label}: AdHocOrdering\vspace{0.1cm} \\
\noindent \textbf{Description}: If the 
        embedded\_sub\_process is Ad Hoc (the AdHoc attribute is True), then the 
	AdHocOrdering attribute MUST be included. This attribute defines if the
	activities within the embedded\_sub\_process can be performed in Parallel
	or must be performed sequentially. The default setting is Parallel and
	the setting of Sequential is a restriction on the performance that may 
	be required due to shared resources.\vspace{0.1cm} \\
\noindent \stepcounter{idaxiom} $AX\_\arabic{idaxiom}$ $\property{has\_embedded\_sub\_process\_ad\_hoc\_ordering} \mbox{ has domain }\concept{embedded\_sub\_process}$\vspace{0.1cm} \\
\noindent \stepcounter{idaxiom} $AX\_\arabic{idaxiom}$ $\property{has\_embedded\_sub\_process\_ad\_hoc\_ordering} \mbox{ has range }\datatype{xsd:string}\{"\datainstance{Sequential}","\datainstance{Parallel}"\}$\vspace{0.1cm} \\
\par \vspace{-0.1cm} \noindent \textbf{Property}: $\property{has\_embedded\_sub\_process\_ad\_hoc\_completion\_condition}$\vspace{0.1cm} \\
\noindent \textbf{Label}: AdHocCompletionCondition\vspace{0.1cm} \\
\noindent \textbf{Description}: If the 
	embedded\_sub\_process is Ad Hoc (the AdHoc attribute is True), then the
	AdHocCompletionCondition attribute MUST be included. This attribute 
	defines the conditions when the embedded\_sub\_process will end.\vspace{0.1cm} \\
\noindent \stepcounter{idaxiom} $AX\_\arabic{idaxiom}$ $\property{has\_embedded\_sub\_process\_ad\_hoc\_completion\_condition} \mbox{ has domain }\concept{embedded\_sub\_process}$\vspace{0.1cm} \\
\noindent \stepcounter{idaxiom} $AX\_\arabic{idaxiom}$ $\property{has\_embedded\_sub\_process\_ad\_hoc\_completion\_condition} \mbox{ has range }\concept{expression}$\vspace{0.1cm} \\
\vspace{0.4cm} \hrule \vspace{0.2cm} \noindent \textbf{Class}: $\concept{reusable\_sub\_process}$\vspace{0.2cm} \hrule \vspace{0.2cm} 
\noindent \textbf{Label}: Reusable Sub-process\vspace{0.1cm} \\
\noindent \textbf{Description}: \vspace{0.1cm} \\
\noindent \stepcounter{idaxiom} $AX\_\arabic{idaxiom}$ $\concept{reusable\_sub\_process} \sqsubseteq  (=1) \property{has\_reusable\_sub\_process\_sub\_diagram\_ref}$\vspace{0.1cm} \\
\par \vspace{-0.1cm} \noindent \textbf{Property}: $\property{has\_reusable\_sub\_process\_sub\_diagram\_ref}$\vspace{0.1cm} \\
\noindent \textbf{Label}: DiagramRef\vspace{0.1cm} \\
\noindent \textbf{Description}: The BPD MUST be 
	identified. The attributes of a BPD can be found in "Business Process 
	Diagram Attributes on page 31."\vspace{0.1cm} \\
\noindent \stepcounter{idaxiom} $AX\_\arabic{idaxiom}$ $\property{has\_reusable\_sub\_process\_sub\_diagram\_ref} \mbox{ has domain }\concept{reusable\_sub\_process}$\vspace{0.1cm} \\
\noindent \stepcounter{idaxiom} $AX\_\arabic{idaxiom}$ $\property{has\_reusable\_sub\_process\_sub\_diagram\_ref} \mbox{ has range }\concept{business\_process\_diagram}$\vspace{0.1cm} \\
\noindent \stepcounter{idaxiom} $AX\_\arabic{idaxiom}$ $\concept{reusable\_sub\_process} \sqsubseteq  (=1) \property{has\_reusable\_sub\_process\_sub\_process\_ref}$\vspace{0.1cm} \\
\par \vspace{-0.1cm} \noindent \textbf{Property}: $\property{has\_reusable\_sub\_process\_sub\_process\_ref}$\vspace{0.1cm} \\
\noindent \textbf{Label}: ProcessRef\vspace{0.1cm} \\
\noindent \textbf{Description}: A Process MUST be 
	identified. The attributes of a Process can be found in "Processes on 
	page 32"\vspace{0.1cm} \\
\noindent \stepcounter{idaxiom} $AX\_\arabic{idaxiom}$ $\property{has\_reusable\_sub\_process\_sub\_process\_ref} \mbox{ has domain }\concept{reusable\_sub\_process}$\vspace{0.1cm} \\
\noindent \stepcounter{idaxiom} $AX\_\arabic{idaxiom}$ $\property{has\_reusable\_sub\_process\_sub\_process\_ref} \mbox{ has range }\concept{process}$\vspace{0.1cm} \\
\par \vspace{-0.1cm} \noindent \textbf{Property}: $\property{has\_reusable\_sub\_process\_sub\_input\_maps}$\vspace{0.1cm} \\
\noindent \textbf{Label}: InputMaps\vspace{0.1cm} \\
\noindent \textbf{Description}: Multiple input 
        mappings MAY be made between the Reusable Sub-Process and the Process 
	referenced by this object. These mappings are in the form of an 
	expression. A specific mapping expression MUST specify the mapping of
	Properties between the two Processes OR the mapping of Artifacts between
	the two Processes.\vspace{0.1cm} \\
\noindent \stepcounter{idaxiom} $AX\_\arabic{idaxiom}$ $\property{has\_reusable\_sub\_process\_sub\_input\_maps} \mbox{ has domain }\concept{reusable\_sub\_process}$\vspace{0.1cm} \\
\noindent \stepcounter{idaxiom} $AX\_\arabic{idaxiom}$ $\property{has\_reusable\_sub\_process\_sub\_input\_maps} \mbox{ has range }\concept{expression}$\vspace{0.1cm} \\
\par \vspace{-0.1cm} \noindent \textbf{Property}: $\property{has\_reusable\_sub\_process\_sub\_output\_maps}$\vspace{0.1cm} \\
\noindent \textbf{Label}: OutputMaps\vspace{0.1cm} \\
\noindent \textbf{Description}: Multiple output 
	mappings MAY be made between the Reusable Sub-Process and the Process 
	referenced by this object. These mappings are in the form of an 
	expression. A specific mapping expression MUST specify the mapping of
	Properties between the two Processes OR the mapping of Artifacts between
	the two Processes.\vspace{0.1cm} \\
\noindent \stepcounter{idaxiom} $AX\_\arabic{idaxiom}$ $\property{has\_reusable\_sub\_process\_sub\_output\_maps} \mbox{ has domain }\concept{reusable\_sub\_process}$\vspace{0.1cm} \\
\noindent \stepcounter{idaxiom} $AX\_\arabic{idaxiom}$ $\property{has\_reusable\_sub\_process\_sub\_output\_maps} \mbox{ has range }\concept{expression}$\vspace{0.1cm} \\
\vspace{0.4cm} \hrule \vspace{0.2cm} \noindent \textbf{Class}: $\concept{reference\_sub\_process}$\vspace{0.2cm} \hrule \vspace{0.2cm} 
\noindent \textbf{Label}: Reference Sub-process\vspace{0.1cm} \\
\noindent \textbf{Description}: \vspace{0.1cm} \\
\noindent \stepcounter{idaxiom} $AX\_\arabic{idaxiom}$ $\concept{reference\_sub\_process} \sqsubseteq  (=1) \property{has\_reference\_sub\_process\_sub\_sub\_process\_ref}$\vspace{0.1cm} \\
\par \vspace{-0.1cm} \noindent \textbf{Property}: $\property{has\_reference\_sub\_process\_sub\_sub\_process\_ref}$\vspace{0.1cm} \\
\noindent \textbf{Label}: SubProcessRef\vspace{0.1cm} \\
\noindent \textbf{Description}: The Sub-Process
	being referenced MUST be identified. The attributes for the Sub-Process
	element can be found in Table B.12.\vspace{0.1cm} \\
\noindent \stepcounter{idaxiom} $AX\_\arabic{idaxiom}$ $\property{has\_reference\_sub\_process\_sub\_sub\_process\_ref} \mbox{ has domain }\concept{reference\_sub\_process}$\vspace{0.1cm} \\
\noindent \stepcounter{idaxiom} $AX\_\arabic{idaxiom}$ $\property{has\_reference\_sub\_process\_sub\_sub\_process\_ref} \mbox{ has range }\concept{sub\_process}$\vspace{0.1cm} \\
\vspace{0.4cm} \hrule \vspace{0.2cm} \noindent \textbf{Class}: $\concept{task}$\vspace{0.2cm} \hrule \vspace{0.2cm} 
\noindent \textbf{Label}: Task [Atomic]\vspace{0.1cm} \\
\noindent \textbf{Description}: A Task is an atomic activity that is included within a 
	Process. A Task is used when the work in the Process is not broken down
	to a finer level of Process Model detail.\vspace{0.1cm} \\
\noindent \stepcounter{idaxiom} $AX\_\arabic{idaxiom}$ $\concept{task} \sqsubseteq (\geq1) \property{has\_task\_task\_type}$\vspace{0.1cm} \\
\par \vspace{-0.1cm} \noindent \textbf{Property}: $\property{has\_task\_task\_type}$\vspace{0.1cm} \\
\noindent \textbf{Label}: TaskType\vspace{0.1cm} \\
\noindent \textbf{Description}: TaskType is an attribute that has a default
	of None, but MAY be set to Send, Receive, User, Script, Abstract, 
	Manual, Reference, or Service. The TaskType will be impacted by the 
	Message Flow to and/or from the Task, if Message Flow are used. A 
	TaskType of Receive MUST NOT have an outgoing Message Flow. A TaskType
	of Send MUST NOT have an incoming Message Flow. A TaskType of Script or
	Manual MUST NOT have an incoming or an outgoing Message Flow.
	The TaskType list MAY be extended to include new types. 
	The attributes for specific settings of TaskType can be found in Table 
	B.17 through Table B.22.\vspace{0.1cm} \\
\noindent \stepcounter{idaxiom} $AX\_\arabic{idaxiom}$ $\property{has\_task\_task\_type} \mbox{ has domain }\concept{task}$\vspace{0.1cm} \\
\noindent \stepcounter{idaxiom} $AX\_\arabic{idaxiom}$ $\property{has\_task\_task\_type} \mbox{ has range }\concept{task\_types}$\vspace{0.1cm} \\
\noindent \stepcounter{idaxiom} $AX\_\arabic{idaxiom}$ $\concept{task\_types} \equiv \{\instance{service\_task\_type}, \instance{receive\_task\_type}, \instance{send\_task\_type}, \instance{user\_task\_type}, \instance{script\_task\_type}, \\ \instance{abstract\_task\_type}, \instance{manual\_task\_type}, \instance{reference\_task\_type}\}$\vspace{0.1cm} \\
\par \vspace{-0.1cm} \noindent \textbf{Instance}: $\instance{service\_task\_type}$\vspace{0.1cm} \\
\noindent \textbf{Label}: Service\vspace{0.1cm} \\
\par \vspace{-0.1cm} \noindent \textbf{Instance}: $\instance{receive\_task\_type}$\vspace{0.1cm} \\
\noindent \textbf{Label}: Receive\vspace{0.1cm} \\
\par \vspace{-0.1cm} \noindent \textbf{Instance}: $\instance{send\_task\_type}$\vspace{0.1cm} \\
\noindent \textbf{Label}: Send\vspace{0.1cm} \\
\par \vspace{-0.1cm} \noindent \textbf{Instance}: $\instance{user\_task\_type}$\vspace{0.1cm} \\
\noindent \textbf{Label}: User\vspace{0.1cm} \\
\par \vspace{-0.1cm} \noindent \textbf{Instance}: $\instance{script\_task\_type}$\vspace{0.1cm} \\
\noindent \textbf{Label}: Script\vspace{0.1cm} \\
\par \vspace{-0.1cm} \noindent \textbf{Instance}: $\instance{abstract\_task\_type}$\vspace{0.1cm} \\
\noindent \textbf{Label}: Abstract\vspace{0.1cm} \\
\par \vspace{-0.1cm} \noindent \textbf{Instance}: $\instance{manual\_task\_type}$\vspace{0.1cm} \\
\noindent \textbf{Label}: Manual\vspace{0.1cm} \\
\par \vspace{-0.1cm} \noindent \textbf{Instance}: $\instance{reference\_task\_type}$\vspace{0.1cm} \\
\noindent \textbf{Label}: Reference\vspace{0.1cm} \\
\noindent \stepcounter{idaxiom} $AX\_\arabic{idaxiom}$ $\concept{service\_task} \equiv \concept{task} \sqcap \exists\property{has\_task\_task\_type}.\{\instance{service\_task\_type}\}$\vspace{0.1cm} \\
\noindent \stepcounter{idaxiom} $AX\_\arabic{idaxiom}$ $\concept{receive\_task} \equiv \concept{task} \sqcap \exists\property{has\_task\_task\_type}.\{\instance{receive\_task\_type}\}$\vspace{0.1cm} \\
\noindent \stepcounter{idaxiom} $AX\_\arabic{idaxiom}$ $\concept{send\_task} \equiv \concept{task} \sqcap \exists\property{has\_task\_task\_type}.\{\instance{send\_task\_type}\}$\vspace{0.1cm} \\
\noindent \stepcounter{idaxiom} $AX\_\arabic{idaxiom}$ $\concept{user\_task} \equiv \concept{task} \sqcap \exists\property{has\_task\_task\_type}.\{\instance{user\_task\_type}\}$\vspace{0.1cm} \\
\noindent \stepcounter{idaxiom} $AX\_\arabic{idaxiom}$ $\concept{script\_task} \equiv \concept{task} \sqcap \exists\property{has\_task\_task\_type}.\{\instance{script\_task\_type}\}$\vspace{0.1cm} \\
\noindent \stepcounter{idaxiom} $AX\_\arabic{idaxiom}$ $\concept{abstract\_task} \equiv \concept{task} \sqcap \exists\property{has\_task\_task\_type}.\{\instance{abstract\_task\_type}\}$\vspace{0.1cm} \\
\noindent \stepcounter{idaxiom} $AX\_\arabic{idaxiom}$ $\concept{manual\_task} \equiv \concept{task} \sqcap \exists\property{has\_task\_task\_type}.\{\instance{manual\_task\_type}\}$\vspace{0.1cm} \\
\noindent \stepcounter{idaxiom} $AX\_\arabic{idaxiom}$ $\concept{reference\_task} \equiv \concept{task} \sqcap \exists\property{has\_task\_task\_type}.\{\instance{reference\_task\_type}\}$\vspace{0.1cm} \\
\noindent \stepcounter{idaxiom} $AX\_\arabic{idaxiom}$ $\concept{service\_task} \sqsubseteq  \neg \concept{receive\_task}$\vspace{0.1cm} \\
\noindent \stepcounter{idaxiom} $AX\_\arabic{idaxiom}$ $\concept{service\_task} \sqsubseteq  \neg \concept{send\_task}$\vspace{0.1cm} \\
\noindent \stepcounter{idaxiom} $AX\_\arabic{idaxiom}$ $\concept{service\_task} \sqsubseteq  \neg \concept{user\_task}$\vspace{0.1cm} \\
\noindent \stepcounter{idaxiom} $AX\_\arabic{idaxiom}$ $\concept{service\_task} \sqsubseteq  \neg \concept{script\_task}$\vspace{0.1cm} \\
\noindent \stepcounter{idaxiom} $AX\_\arabic{idaxiom}$ $\concept{service\_task} \sqsubseteq  \neg \concept{abstract\_task}$\vspace{0.1cm} \\
\noindent \stepcounter{idaxiom} $AX\_\arabic{idaxiom}$ $\concept{service\_task} \sqsubseteq  \neg \concept{manual\_task}$\vspace{0.1cm} \\
\noindent \stepcounter{idaxiom} $AX\_\arabic{idaxiom}$ $\concept{service\_task} \sqsubseteq  \neg \concept{reference\_task}$\vspace{0.1cm} \\
\noindent \stepcounter{idaxiom} $AX\_\arabic{idaxiom}$ $\concept{receive\_task} \sqsubseteq  \neg \concept{send\_task}$\vspace{0.1cm} \\
\noindent \stepcounter{idaxiom} $AX\_\arabic{idaxiom}$ $\concept{receive\_task} \sqsubseteq  \neg \concept{user\_task}$\vspace{0.1cm} \\
\noindent \stepcounter{idaxiom} $AX\_\arabic{idaxiom}$ $\concept{receive\_task} \sqsubseteq  \neg \concept{script\_task}$\vspace{0.1cm} \\
\noindent \stepcounter{idaxiom} $AX\_\arabic{idaxiom}$ $\concept{receive\_task} \sqsubseteq  \neg \concept{abstract\_task}$\vspace{0.1cm} \\
\noindent \stepcounter{idaxiom} $AX\_\arabic{idaxiom}$ $\concept{receive\_task} \sqsubseteq  \neg \concept{manual\_task}$\vspace{0.1cm} \\
\noindent \stepcounter{idaxiom} $AX\_\arabic{idaxiom}$ $\concept{receive\_task} \sqsubseteq  \neg \concept{reference\_task}$\vspace{0.1cm} \\
\noindent \stepcounter{idaxiom} $AX\_\arabic{idaxiom}$ $\concept{send\_task} \sqsubseteq  \neg \concept{user\_task}$\vspace{0.1cm} \\
\noindent \stepcounter{idaxiom} $AX\_\arabic{idaxiom}$ $\concept{send\_task} \sqsubseteq  \neg \concept{script\_task}$\vspace{0.1cm} \\
\noindent \stepcounter{idaxiom} $AX\_\arabic{idaxiom}$ $\concept{send\_task} \sqsubseteq  \neg \concept{abstract\_task}$\vspace{0.1cm} \\
\noindent \stepcounter{idaxiom} $AX\_\arabic{idaxiom}$ $\concept{send\_task} \sqsubseteq  \neg \concept{manual\_task}$\vspace{0.1cm} \\
\noindent \stepcounter{idaxiom} $AX\_\arabic{idaxiom}$ $\concept{send\_task} \sqsubseteq  \neg \concept{reference\_task}$\vspace{0.1cm} \\
\noindent \stepcounter{idaxiom} $AX\_\arabic{idaxiom}$ $\concept{user\_task} \sqsubseteq  \neg \concept{script\_task}$\vspace{0.1cm} \\
\noindent \stepcounter{idaxiom} $AX\_\arabic{idaxiom}$ $\concept{user\_task} \sqsubseteq  \neg \concept{abstract\_task}$\vspace{0.1cm} \\
\noindent \stepcounter{idaxiom} $AX\_\arabic{idaxiom}$ $\concept{user\_task} \sqsubseteq  \neg \concept{manual\_task}$\vspace{0.1cm} \\
\noindent \stepcounter{idaxiom} $AX\_\arabic{idaxiom}$ $\concept{user\_task} \sqsubseteq  \neg \concept{reference\_task}$\vspace{0.1cm} \\
\noindent \stepcounter{idaxiom} $AX\_\arabic{idaxiom}$ $\concept{script\_task} \sqsubseteq  \neg \concept{abstract\_task}$\vspace{0.1cm} \\
\noindent \stepcounter{idaxiom} $AX\_\arabic{idaxiom}$ $\concept{script\_task} \sqsubseteq  \neg \concept{manual\_task}$\vspace{0.1cm} \\
\noindent \stepcounter{idaxiom} $AX\_\arabic{idaxiom}$ $\concept{script\_task} \sqsubseteq  \neg \concept{reference\_task}$\vspace{0.1cm} \\
\noindent \stepcounter{idaxiom} $AX\_\arabic{idaxiom}$ $\concept{abstract\_task} \sqsubseteq  \neg \concept{manual\_task}$\vspace{0.1cm} \\
\noindent \stepcounter{idaxiom} $AX\_\arabic{idaxiom}$ $\concept{abstract\_task} \sqsubseteq  \neg \concept{reference\_task}$\vspace{0.1cm} \\
\noindent \stepcounter{idaxiom} $AX\_\arabic{idaxiom}$ $\concept{manual\_task} \sqsubseteq  \neg \concept{reference\_task}$\vspace{0.1cm} \\
\vspace{0.4cm} \hrule \vspace{0.2cm} \noindent \textbf{Class}: $\concept{service\_task}$\vspace{0.2cm} \hrule \vspace{0.2cm} 
\noindent \textbf{Label}: Service Task\vspace{0.1cm} \\
\noindent \textbf{Description}: \vspace{0.1cm} \\
\noindent \stepcounter{idaxiom} $AX\_\arabic{idaxiom}$ $\concept{service\_task} \sqsubseteq  (=1) \property{has\_service\_task\_in\_message\_ref}$\vspace{0.1cm} \\
\par \vspace{-0.1cm} \noindent \textbf{Property}: $\property{has\_service\_task\_in\_message\_ref}$\vspace{0.1cm} \\
\noindent \textbf{Label}: InMessageRef\vspace{0.1cm} \\
\noindent \textbf{Description}: A Message for the InMessageRef
	attribute MUST be entered. This indicates that the Message will be 
	received at the start of the Task, after the availability of any defined
	InputSets. One or more corresponding incoming Message Flows MAY be shown
	on the diagram. However, the display of the Message Flow is not 
	required. The Message is applied to all incoming Message Flow, but can 
	arrive for only one of the incoming Message Flow for a single instance 
	of the Task.\vspace{0.1cm} \\
\noindent \stepcounter{idaxiom} $AX\_\arabic{idaxiom}$ $\property{has\_service\_task\_in\_message\_ref} \mbox{ has domain }\concept{service\_task}$\vspace{0.1cm} \\
\noindent \stepcounter{idaxiom} $AX\_\arabic{idaxiom}$ $\property{has\_service\_task\_in\_message\_ref} \mbox{ has range }\concept{message}$\vspace{0.1cm} \\
\noindent \stepcounter{idaxiom} $AX\_\arabic{idaxiom}$ $\concept{service\_task} \sqsubseteq  (=1) \property{has\_service\_task\_out\_message\_ref}$\vspace{0.1cm} \\
\par \vspace{-0.1cm} \noindent \textbf{Property}: $\property{has\_service\_task\_out\_message\_ref}$\vspace{0.1cm} \\
\noindent \textbf{Label}: OutMessageRef\vspace{0.1cm} \\
\noindent \textbf{Description}: A Message for the 
	OutMessageRef attribute MUST be entered. The sending of this message
	marks the completion of the Task, which may cause the production of an
	OutputSet. One or more corresponding outgoing Message Flow MAY be shown
	on the diagram. However, the display of the Message Flow is not 
	required. The Message is applied to all outgoing Message Flow and the
	Message will be sent down all outgoing Message Flow at the completion of
	a single instance of the Task.\vspace{0.1cm} \\
\noindent \stepcounter{idaxiom} $AX\_\arabic{idaxiom}$ $\property{has\_service\_task\_out\_message\_ref} \mbox{ has domain }\concept{service\_task}$\vspace{0.1cm} \\
\noindent \stepcounter{idaxiom} $AX\_\arabic{idaxiom}$ $\property{has\_service\_task\_out\_message\_ref} \mbox{ has range }\concept{message}$\vspace{0.1cm} \\
\par \vspace{-0.1cm} \noindent \textbf{Property}: $\property{has\_service\_task\_implementation}$\vspace{0.1cm} \\
\noindent \textbf{Label}: Implementation\vspace{0.1cm} \\
\noindent \textbf{Description}: This attribute specifies the
	technology that will be used to send or receive the message. A Web 
	service is the default technology.\vspace{0.1cm} \\
\noindent \stepcounter{idaxiom} $AX\_\arabic{idaxiom}$ $\property{has\_service\_task\_implementation} \mbox{ has domain }\concept{service\_task}$\vspace{0.1cm} \\
\noindent \stepcounter{idaxiom} $AX\_\arabic{idaxiom}$ $\property{has\_service\_task\_implementation} \mbox{ has range }\datatype{xsd:string}\{"\datainstance{Web\_Service}","\datainstance{Other}","\datainstance{Unspecified}"\}$\vspace{0.1cm} \\
\vspace{0.4cm} \hrule \vspace{0.2cm} \noindent \textbf{Class}: $\concept{receive\_task}$\vspace{0.2cm} \hrule \vspace{0.2cm} 
\noindent \textbf{Label}: Receive Task\vspace{0.1cm} \\
\noindent \textbf{Description}: \vspace{0.1cm} \\
\noindent \stepcounter{idaxiom} $AX\_\arabic{idaxiom}$ $\concept{receive\_task} \sqsubseteq  (=1) \property{has\_receive\_task\_message\_ref}$\vspace{0.1cm} \\
\par \vspace{-0.1cm} \noindent \textbf{Property}: $\property{has\_receive\_task\_message\_ref}$\vspace{0.1cm} \\
\noindent \textbf{Label}: MessageRef\vspace{0.1cm} \\
\noindent \textbf{Description}: A Message for the MessageRef
	attribute MUST be entered. This indicates that the Message will be
	received by the Task. The Message in this context is equivalent to an 
	in-only message pattern (Web service). One or more corresponding 
	incoming Message Flow MAY be shown on the diagram. However, the display
	of the Message Flow is not required. The Message is applied to all
	incoming Message Flow, but can arrive for only one of the incoming
	Message Flow for a single instance of the Task.\vspace{0.1cm} \\
\noindent \stepcounter{idaxiom} $AX\_\arabic{idaxiom}$ $\property{has\_receive\_task\_message\_ref} \mbox{ has domain }\concept{receive\_task}$\vspace{0.1cm} \\
\noindent \stepcounter{idaxiom} $AX\_\arabic{idaxiom}$ $\property{has\_receive\_task\_message\_ref} \mbox{ has range }\concept{message}$\vspace{0.1cm} \\
\noindent \stepcounter{idaxiom} $AX\_\arabic{idaxiom}$ $\concept{receive\_task} \sqsubseteq  (=1) \property{has\_receive\_task\_instantiate}$\vspace{0.1cm} \\
\par \vspace{-0.1cm} \noindent \textbf{Property}: $\property{has\_receive\_task\_instantiate}$\vspace{0.1cm} \\
\noindent \textbf{Label}: Instantiate\vspace{0.1cm} \\
\noindent \textbf{Description}: Receive Tasks can be defined as
	the instantiation mechanism for the Process with the Instantiate 
	attribute. This attribute MAY be set to true if the Task is the first
	activity after the Start Event or a starting Task if there is no Start 
	Event. Multiple Tasks MAY have this attribute set to True.\vspace{0.1cm} \\
\noindent \stepcounter{idaxiom} $AX\_\arabic{idaxiom}$ $\property{has\_receive\_task\_instantiate} \mbox{ has domain }\concept{receive\_task}$\vspace{0.1cm} \\
\noindent \stepcounter{idaxiom} $AX\_\arabic{idaxiom}$ $\property{has\_receive\_task\_instantiate} \mbox{ has range }\datatype{xsd:boolean}$\vspace{0.1cm} \\
\par \vspace{-0.1cm} \noindent \textbf{Property}: $\property{has\_receive\_task\_implementation}$\vspace{0.1cm} \\
\noindent \textbf{Label}: Implementation\vspace{0.1cm} \\
\noindent \textbf{Description}: This attribute specifies the
	technology that will be used to receive the message. A Web service is 
	the default technology.\vspace{0.1cm} \\
\noindent \stepcounter{idaxiom} $AX\_\arabic{idaxiom}$ $\property{has\_receive\_task\_implementation} \mbox{ has domain }\concept{receive\_task}$\vspace{0.1cm} \\
\noindent \stepcounter{idaxiom} $AX\_\arabic{idaxiom}$ $\property{has\_receive\_task\_implementation} \mbox{ has range }\datatype{xsd:string}\{"\datainstance{Web\_Service}","\datainstance{Other}","\datainstance{Unspecified}"\}$\vspace{0.1cm} \\
\vspace{0.4cm} \hrule \vspace{0.2cm} \noindent \textbf{Class}: $\concept{send\_task}$\vspace{0.2cm} \hrule \vspace{0.2cm} 
\noindent \textbf{Label}: Send Task\vspace{0.1cm} \\
\noindent \textbf{Description}: \vspace{0.1cm} \\
\noindent \stepcounter{idaxiom} $AX\_\arabic{idaxiom}$ $\concept{send\_task} \sqsubseteq  (=1) \property{has\_send\_task\_message\_ref}$\vspace{0.1cm} \\
\par \vspace{-0.1cm} \noindent \textbf{Property}: $\property{has\_send\_task\_message\_ref}$\vspace{0.1cm} \\
\noindent \textbf{Label}: MessageRef\vspace{0.1cm} \\
\noindent \textbf{Description}: A Message for the MessageRef
	attribute MUST be entered. This indicates that the Message will be sent
	by the Task. The Message in this context is equivalent to an out-only 
	message pattern (Web service). One or more corresponding outgoing Message
	Flow MAY be shown on the diagram. However, the display of the Message 
	Flow is not required. The Message is applied to all outgoing Message
	Flow and the Message will be sent down all outgoing Message Flow at the
	completion of a single instance of the Task.\vspace{0.1cm} \\
\noindent \stepcounter{idaxiom} $AX\_\arabic{idaxiom}$ $\property{has\_send\_task\_message\_ref} \mbox{ has domain }\concept{send\_task}$\vspace{0.1cm} \\
\noindent \stepcounter{idaxiom} $AX\_\arabic{idaxiom}$ $\property{has\_send\_task\_message\_ref} \mbox{ has range }\concept{message}$\vspace{0.1cm} \\
\par \vspace{-0.1cm} \noindent \textbf{Property}: $\property{has\_send\_task\_implementation}$\vspace{0.1cm} \\
\noindent \textbf{Label}: Implementation\vspace{0.1cm} \\
\noindent \textbf{Description}: This attribute specifies the
	technology that will be used to send the message. A Web service
	is the default technology.\vspace{0.1cm} \\
\noindent \stepcounter{idaxiom} $AX\_\arabic{idaxiom}$ $\property{has\_send\_task\_implementation} \mbox{ has domain }\concept{send\_task}$\vspace{0.1cm} \\
\noindent \stepcounter{idaxiom} $AX\_\arabic{idaxiom}$ $\property{has\_send\_task\_implementation} \mbox{ has range }\datatype{xsd:string}\{"\datainstance{Web\_Service}","\datainstance{Other}","\datainstance{Unspecified}"\}$\vspace{0.1cm} \\
\vspace{0.4cm} \hrule \vspace{0.2cm} \noindent \textbf{Class}: $\concept{user\_task}$\vspace{0.2cm} \hrule \vspace{0.2cm} 
\noindent \textbf{Label}: User Task\vspace{0.1cm} \\
\noindent \textbf{Description}: \vspace{0.1cm} \\
\noindent \stepcounter{idaxiom} $AX\_\arabic{idaxiom}$ $\concept{user\_task} \sqsubseteq  (=1) \property{has\_user\_task\_in\_message\_ref}$\vspace{0.1cm} \\
\par \vspace{-0.1cm} \noindent \textbf{Property}: $\property{has\_user\_task\_in\_message\_ref}$\vspace{0.1cm} \\
\noindent \textbf{Label}: InMessageRef\vspace{0.1cm} \\
\noindent \textbf{Description}: A Message for the InMessageRef
	attribute MUST be entered. This indicates that the Message will be
	received at the start of the Task, after the availability of any defined
	InputSets. One or more corresponding incoming Message Flows MAY be shown
	on the diagram. However, the display of the Message Flow is not 
	required.
	The Message is applied to all incoming Message Flow, but can arrive for
	only one of the incoming Message Flow for a single instance of the 
	Task.\vspace{0.1cm} \\
\noindent \stepcounter{idaxiom} $AX\_\arabic{idaxiom}$ $\property{has\_user\_task\_in\_message\_ref} \mbox{ has domain }\concept{user\_task}$\vspace{0.1cm} \\
\noindent \stepcounter{idaxiom} $AX\_\arabic{idaxiom}$ $\property{has\_user\_task\_in\_message\_ref} \mbox{ has range }\concept{message}$\vspace{0.1cm} \\
\noindent \stepcounter{idaxiom} $AX\_\arabic{idaxiom}$ $\concept{user\_task} \sqsubseteq  (=1) \property{has\_user\_task\_out\_message\_ref}$\vspace{0.1cm} \\
\par \vspace{-0.1cm} \noindent \textbf{Property}: $\property{has\_user\_task\_out\_message\_ref}$\vspace{0.1cm} \\
\noindent \textbf{Label}: OutMessageRef\vspace{0.1cm} \\
\noindent \textbf{Description}: A Message for the OutMessageRef
	attribute MUST be entered. The sending of this message marks the
	completion of the Task, which may cause the production of an OutputSet.
	One or more corresponding outgoing Message Flow MAY be shown on the
	diagram. However, the display of the Message Flow is not required.
	The Message is applied to all outgoing Message Flow and the Message will
	be sent down all outgoing Message Flow at the completion of a single
	instance of the Task.\vspace{0.1cm} \\
\noindent \stepcounter{idaxiom} $AX\_\arabic{idaxiom}$ $\property{has\_user\_task\_out\_message\_ref} \mbox{ has domain }\concept{user\_task}$\vspace{0.1cm} \\
\noindent \stepcounter{idaxiom} $AX\_\arabic{idaxiom}$ $\property{has\_user\_task\_out\_message\_ref} \mbox{ has range }\concept{message}$\vspace{0.1cm} \\
\par \vspace{-0.1cm} \noindent \textbf{Property}: $\property{has\_user\_task\_implementation}$\vspace{0.1cm} \\
\noindent \textbf{Label}: Implementation\vspace{0.1cm} \\
\noindent \textbf{Description}: This attribute specifies the
	technology that will be used by the Performers to perform the task. 
	A Web service is the default technology.\vspace{0.1cm} \\
\noindent \stepcounter{idaxiom} $AX\_\arabic{idaxiom}$ $\property{has\_user\_task\_implementation} \mbox{ has domain }\concept{user\_task}$\vspace{0.1cm} \\
\noindent \stepcounter{idaxiom} $AX\_\arabic{idaxiom}$ $\property{has\_user\_task\_implementation} \mbox{ has range }\datatype{xsd:string}\{"\datainstance{Web\_Service}","\datainstance{Other}","\datainstance{Unspecified}"\}$\vspace{0.1cm} \\
\vspace{0.4cm} \hrule \vspace{0.2cm} \noindent \textbf{Class}: $\concept{script\_task}$\vspace{0.2cm} \hrule \vspace{0.2cm} 
\noindent \textbf{Label}: Script Task\vspace{0.1cm} \\
\noindent \textbf{Description}: \vspace{0.1cm} \\
\noindent \stepcounter{idaxiom} $AX\_\arabic{idaxiom}$ $\concept{script\_task} \sqsubseteq (\geq1) \property{has\_script\_task\_script}$\vspace{0.1cm} \\
\par \vspace{-0.1cm} \noindent \textbf{Property}: $\property{has\_script\_task\_script}$\vspace{0.1cm} \\
\noindent \textbf{Label}: Script\vspace{0.1cm} \\
\noindent \textbf{Description}: The modeler MAY include a script that 
	can be run when the Task is performed. If a script is not included, 
	then the Task will act equivalent to a TaskType of None.\vspace{0.1cm} \\
\noindent \stepcounter{idaxiom} $AX\_\arabic{idaxiom}$ $\property{has\_script\_task\_script} \mbox{ has domain }\concept{script\_task}$\vspace{0.1cm} \\
\noindent \stepcounter{idaxiom} $AX\_\arabic{idaxiom}$ $\property{has\_script\_task\_script} \mbox{ has range }\datatype{xsd:string}$\vspace{0.1cm} \\
\vspace{0.4cm} \hrule \vspace{0.2cm} \noindent \textbf{Class}: $\concept{reference\_task}$\vspace{0.2cm} \hrule \vspace{0.2cm} 
\noindent \textbf{Label}: Reference Task\vspace{0.1cm} \\
\noindent \textbf{Description}: \vspace{0.1cm} \\
\noindent \stepcounter{idaxiom} $AX\_\arabic{idaxiom}$ $\concept{reference\_task} \sqsubseteq  (=1) \property{has\_reference\_task\_task\_ref}$\vspace{0.1cm} \\
\par \vspace{-0.1cm} \noindent \textbf{Property}: $\property{has\_reference\_task\_task\_ref}$\vspace{0.1cm} \\
\noindent \textbf{Label}: TaskRef\vspace{0.1cm} \\
\noindent \textbf{Description}: The Task being referenced MUST be 
	identified. The attributes for the Task element can be found in Table 
	B.16.\vspace{0.1cm} \\
\noindent \stepcounter{idaxiom} $AX\_\arabic{idaxiom}$ $\property{has\_reference\_task\_task\_ref} \mbox{ has domain }\concept{reference\_task}$\vspace{0.1cm} \\
\noindent \stepcounter{idaxiom} $AX\_\arabic{idaxiom}$ $\property{has\_reference\_task\_task\_ref} \mbox{ has range }\concept{task}$\vspace{0.1cm} \\
\vspace{0.4cm} \hrule \vspace{0.2cm} \noindent \textbf{Class}: $\concept{gateway}$\vspace{0.2cm} \hrule \vspace{0.2cm} 
\noindent \textbf{Label}: Gateway\vspace{0.1cm} \\
\noindent \textbf{Description}: A Gateway is used to control the divergence and 
	convergence of Sequence Flow. Thus, it will determine branching, 
	forking, merging, and joining of paths. Internal Markers will indicate
	the type of behavior control.\vspace{0.1cm} \\
\noindent \stepcounter{idaxiom} $AX\_\arabic{idaxiom}$ $\concept{gateway} \sqsubseteq  (=1) \property{has\_gateway\_gateway\_type}$\vspace{0.1cm} \\
\par \vspace{-0.1cm} \noindent \textbf{Property}: $\property{has\_gateway\_gateway\_type}$\vspace{0.1cm} \\
\noindent \textbf{Label}: GatewayType\vspace{0.1cm} \\
\noindent \textbf{Description}: GatewayType is by default Exclusive.
	The GatewayType MAY be set to Inclusive, Complex, or Parallel. The 
	GatewayType will determine the behavior of the Gateway, both for
	incoming and outgoing Sequence Flow, and will determine the internal 
	indicator (as shown in Figure 9.17).\vspace{0.1cm} \\
\noindent \stepcounter{idaxiom} $AX\_\arabic{idaxiom}$ $\property{has\_gateway\_gateway\_type} \mbox{ has domain }\concept{gateway}$\vspace{0.1cm} \\
\noindent \stepcounter{idaxiom} $AX\_\arabic{idaxiom}$ $\property{has\_gateway\_gateway\_type} \mbox{ has range }\concept{gateway\_types}$\vspace{0.1cm} \\
\vspace{0.4cm} \hrule \vspace{0.2cm} \noindent \textbf{Class}: $\concept{gateway\_types}$\vspace{0.2cm} \hrule \vspace{0.2cm} 
\noindent \textbf{Label}: Gateway Types\vspace{0.1cm} \\
\noindent \textbf{Description}: Icons within the diamond shape will
               indicate the type of flow control behavior.
	       The types of control include: 
	            1. exclusive -- exclusive decision and
	                merging. Both Data-Based and Event-Based. Data-Based 
			can be shown with or without the "X" marker.
 		    2. esclusive -- inclusive decision and merging
		    3. complex -- complex conditions and situations (e.g., 
		        3 out of 5)
  		    4. parallel -- forking and joining
               Each type of control affects both the
	       incoming and outgoing Flow.\vspace{0.1cm} \\
\noindent \stepcounter{idaxiom} $AX\_\arabic{idaxiom}$ $\concept{gateway\_types} \equiv \{\instance{exclusive}, \instance{inclusive}, \instance{complex}, \instance{parallel}\}$\vspace{0.1cm} \\
\par \vspace{-0.1cm} \noindent \textbf{Instance}: $\instance{exclusive}$\vspace{0.1cm} \\
\noindent \textbf{Label}: exclusive\vspace{0.1cm} \\
\noindent \textbf{Description}: exclusive -- exclusive decision parallel merging. 
	Data-Based or Event-Based - can be shown with inclusive without the 
	"X" marker.\vspace{0.1cm} \\
\par \vspace{-0.1cm} \noindent \textbf{Instance}: $\instance{inclusive}$\vspace{0.1cm} \\
\noindent \textbf{Label}: inclusive\vspace{0.1cm} \\
\noindent \textbf{Description}: inclusive -- inclusive decision parallel merging\vspace{0.1cm} \\
\par \vspace{-0.1cm} \noindent \textbf{Instance}: $\instance{complex}$\vspace{0.1cm} \\
\noindent \textbf{Label}: complex\vspace{0.1cm} \\
\noindent \textbf{Description}: Complex -- complex conditions parallel situations 
	(e.g., 3 out of 5)\vspace{0.1cm} \\
\par \vspace{-0.1cm} \noindent \textbf{Instance}: $\instance{parallel}$\vspace{0.1cm} \\
\noindent \textbf{Label}: parallel\vspace{0.1cm} \\
\noindent \textbf{Description}: parallel -- forking parallel joining\vspace{0.1cm} \\
\noindent \stepcounter{idaxiom} $AX\_\arabic{idaxiom}$ $( \neg \{\instance{exclusive}\})(\instance{inclusive})$\vspace{0.1cm} \\
\noindent \stepcounter{idaxiom} $AX\_\arabic{idaxiom}$ $( \neg \{\instance{exclusive}\})(\instance{complex})$\vspace{0.1cm} \\
\noindent \stepcounter{idaxiom} $AX\_\arabic{idaxiom}$ $( \neg \{\instance{exclusive}\})(\instance{parallel})$\vspace{0.1cm} \\
\noindent \stepcounter{idaxiom} $AX\_\arabic{idaxiom}$ $( \neg \{\instance{inclusive}\})(\instance{complex})$\vspace{0.1cm} \\
\noindent \stepcounter{idaxiom} $AX\_\arabic{idaxiom}$ $( \neg \{\instance{inclusive}\})(\instance{parallel})$\vspace{0.1cm} \\
\noindent \stepcounter{idaxiom} $AX\_\arabic{idaxiom}$ $( \neg \{\instance{complex}\})(\instance{parallel})$\vspace{0.1cm} \\
\noindent \stepcounter{idaxiom} $AX\_\arabic{idaxiom}$ $\concept{exclusive\_gateway} \equiv \concept{gateway} \sqcap \exists\property{has\_gateway\_gateway\_type}.\{\instance{exclusive}\}$\vspace{0.1cm} \\
\noindent \stepcounter{idaxiom} $AX\_\arabic{idaxiom}$ $\concept{inclusive\_gateway} \equiv \concept{gateway} \sqcap \exists\property{has\_gateway\_gateway\_type}.\{\instance{inclusive}\}$\vspace{0.1cm} \\
\noindent \stepcounter{idaxiom} $AX\_\arabic{idaxiom}$ $\concept{parallel\_gateway} \equiv \concept{gateway} \sqcap \exists\property{has\_gateway\_gateway\_type}.\{\instance{parallel}\}$\vspace{0.1cm} \\
\noindent \stepcounter{idaxiom} $AX\_\arabic{idaxiom}$ $\concept{complex\_gateway} \equiv \concept{gateway} \sqcap \exists\property{has\_gateway\_gateway\_type}.\{\instance{complex}\}$\vspace{0.1cm} \\
\par \vspace{-0.1cm} \noindent \textbf{Property}: $\property{has\_gateway\_gate}$\vspace{0.1cm} \\
\noindent \textbf{Label}: Gates\vspace{0.1cm} \\
\noindent \textbf{Description}: There MAY be zero or more Gates (except where
	noted below). Zero Gates are allowed if the Gateway is last object in a 
	process flow and there are no Start or 	End Events for the Process.
	If there are zero or only one incoming Sequence Flow, then there MUST be
	at least two Gates.
	For Exclusive Data-Based Gateways: When two Gates are required, one of
	them MAY be the DefaultGate.
	For Exclusive Event-Based Gateways: There MUST be two or more Gates. 
	(Note that this type of Gateway does not act only as a Merge--it is 
	always a Decision, at least.) 
	For Inclusive Gateways: When two Gates are required, one of them MAY be 
	the DefaultGate.\vspace{0.1cm} \\
\noindent \stepcounter{idaxiom} $AX\_\arabic{idaxiom}$ $\property{has\_gateway\_gate} \mbox{ has domain }\concept{gateway}$\vspace{0.1cm} \\
\noindent \stepcounter{idaxiom} $AX\_\arabic{idaxiom}$ $\property{has\_gateway\_gate} \mbox{ has range }\concept{gate}$\vspace{0.1cm} \\
\vspace{0.4cm} \hrule \vspace{0.2cm} \noindent \textbf{Class}: $\concept{exclusive\_gateway}$\vspace{0.2cm} \hrule \vspace{0.2cm} 
\noindent \textbf{Label}: Exclusive Gateway\vspace{0.1cm} \\
\noindent \textbf{Description}: Exclusive Gateway\vspace{0.1cm} \\
\noindent \stepcounter{idaxiom} $AX\_\arabic{idaxiom}$ $\concept{exclusive\_gateway} \sqsubseteq  (=1) \property{has\_exclusive\_gateway\_exclusive\_type}$\vspace{0.1cm} \\
\par \vspace{-0.1cm} \noindent \textbf{Property}: $\property{has\_exclusive\_gateway\_exclusive\_type}$\vspace{0.1cm} \\
\noindent \textbf{Label}: ExclusiveType\vspace{0.1cm} \\
\noindent \textbf{Description}: ExclusiveType is by 
	default Data. The ExclusiveType MAY be set to Event. Since Data-Based 
	Exclusive Gateways is the subject of this section, the attribute MUST
	be set to Data for the attributes and behavior defined in this section
	to apply to the Gateway.\vspace{0.1cm} \\
\noindent \stepcounter{idaxiom} $AX\_\arabic{idaxiom}$ $\property{has\_exclusive\_gateway\_exclusive\_type} \mbox{ has domain }\concept{exclusive\_gateway}$\vspace{0.1cm} \\
\noindent \stepcounter{idaxiom} $AX\_\arabic{idaxiom}$ $\property{has\_exclusive\_gateway\_exclusive\_type} \mbox{ has range }\concept{exclusive\_types}$\vspace{0.1cm} \\
\vspace{0.4cm} \hrule \vspace{0.2cm} \noindent \textbf{Class}: $\concept{exclusive\_types}$\vspace{0.2cm} \hrule \vspace{0.2cm} 
\noindent \textbf{Label}: Exclusive Types\vspace{0.1cm} \\
\noindent \textbf{Description}: \vspace{0.1cm} \\
\noindent \stepcounter{idaxiom} $AX\_\arabic{idaxiom}$ $\concept{exclusive\_types} \equiv \{\instance{data\_exclusive\_type}, \instance{event\_exclusive\_type}\}$\vspace{0.1cm} \\
\par \vspace{-0.1cm} \noindent \textbf{Instance}: $\instance{data\_exclusive\_type}$\vspace{0.1cm} \\
\noindent \textbf{Label}: data\vspace{0.1cm} \\
\noindent \textbf{Description}: data -- Data-Based\vspace{0.1cm} \\
\par \vspace{-0.1cm} \noindent \textbf{Instance}: $\instance{event\_exclusive\_type}$\vspace{0.1cm} \\
\noindent \textbf{Label}: event\vspace{0.1cm} \\
\noindent \textbf{Description}: event -- Event-based\vspace{0.1cm} \\
\noindent \stepcounter{idaxiom} $AX\_\arabic{idaxiom}$ $( \neg \{\instance{data\_exclusive\_type}\})(\instance{event\_exclusive\_type})$\vspace{0.1cm} \\
\noindent \stepcounter{idaxiom} $AX\_\arabic{idaxiom}$ $\concept{data\_based\_exclusive\_gateway} \equiv \concept{exclusive\_gateway} \sqcap \\ 
\exists\property{has\_exclusive\_gateway\_exclusive\_type}.\{\instance{data\_exclusive\_type}\}$\vspace{0.1cm} \\
\noindent \stepcounter{idaxiom} $AX\_\arabic{idaxiom}$ $\concept{event\_based\_exclusive\_gateway} \equiv \concept{exclusive\_gateway} \sqcap \\ 
\exists\property{has\_exclusive\_gateway\_exclusive\_type}.\{\instance{event\_exclusive\_type}\}$\vspace{0.1cm} \\
\vspace{0.4cm} \hrule \vspace{0.2cm} \noindent \textbf{Class}: $\concept{data\_based\_exclusive\_gateway}$\vspace{0.2cm} \hrule \vspace{0.2cm} 
\noindent \textbf{Label}: Data Based Exclusive Gateway\vspace{0.1cm} \\
\noindent \textbf{Description}: Data Based Exclusive Gateway\vspace{0.1cm} \\
\noindent \stepcounter{idaxiom} $AX\_\arabic{idaxiom}$ $\concept{data\_based\_exclusive\_gateway} \sqsubseteq  (=1) \property{has\_data\_based\_exclusive\_gateway\_marker\_visible}$\vspace{0.1cm} \\
\par \vspace{-0.1cm} \noindent \textbf{Property}: $\property{has\_data\_based\_exclusive\_gateway\_marker\_visible}$\vspace{0.1cm} \\
\noindent \textbf{Label}: MarkerVisible\vspace{0.1cm} \\
\noindent \textbf{Description}: This attribute
	determines if the Exclusive Marker is displayed in the center of the
	Gateway diamond (an "X"). The marker is displayed if the attribute is 
	True and it is not displayed if the attribute is False. By default, the
	marker is not displayed.\vspace{0.1cm} \\
\noindent \stepcounter{idaxiom} $AX\_\arabic{idaxiom}$ $\property{has\_data\_based\_exclusive\_gateway\_marker\_visible} \mbox{ has domain }\concept{data\_based\_exclusive\_gateway}$\vspace{0.1cm} \\
\noindent \stepcounter{idaxiom} $AX\_\arabic{idaxiom}$ $\property{has\_data\_based\_exclusive\_gateway\_marker\_visible} \mbox{ has range }\datatype{xsd:boolean}$\vspace{0.1cm} \\
\noindent \stepcounter{idaxiom} $AX\_\arabic{idaxiom}$ $\concept{data\_based\_exclusive\_gateway} \sqsubseteq (\geq1) \property{has\_data\_based\_exclusive\_gateway\_default\_gate}$\vspace{0.1cm} \\
\par \vspace{-0.1cm} \noindent \textbf{Property}: $\property{has\_data\_based\_exclusive\_gateway\_default\_gate}$\vspace{0.1cm} \\
\noindent \textbf{Label}: DefaultGate\vspace{0.1cm} \\
\noindent \textbf{Description}: A Default Gate 
	MAY be specified (see Section B.11.9, "Gate," on page 274).\vspace{0.1cm} \\
\noindent \stepcounter{idaxiom} $AX\_\arabic{idaxiom}$ $\property{has\_data\_based\_exclusive\_gateway\_default\_gate} \mbox{ has domain }\concept{data\_based\_exclusive\_gateway}$\vspace{0.1cm} \\
\noindent \stepcounter{idaxiom} $AX\_\arabic{idaxiom}$ $\property{has\_data\_based\_exclusive\_gateway\_default\_gate} \mbox{ has range }\concept{gate}$\vspace{0.1cm} \\
\vspace{0.4cm} \hrule \vspace{0.2cm} \noindent \textbf{Class}: $\concept{event\_based\_exclusive\_gateway}$\vspace{0.2cm} \hrule \vspace{0.2cm} 
\noindent \textbf{Label}: Event Based Exclusive Gateway\vspace{0.1cm} \\
\noindent \textbf{Description}: Event Based Exclusive Gateway\vspace{0.1cm} \\
\noindent \stepcounter{idaxiom} $AX\_\arabic{idaxiom}$ $\concept{event\_based\_exclusive\_gateway} \sqsubseteq  (=1) \property{has\_event\_based\_exclusive\_gateway\_instantiate}$\vspace{0.1cm} \\
\par \vspace{-0.1cm} \noindent \textbf{Property}: $\property{has\_event\_based\_exclusive\_gateway\_instantiate}$\vspace{0.1cm} \\
\noindent \textbf{Label}: MarkerVisible\vspace{0.1cm} \\
\noindent \textbf{Description}: Event-Based
	Gateways can be defined as the instantiation mechanism for the Process 
	with the Instantiate attribute. This attribute MAY be set to true if the
	Gateway is the first element after the Start Event or a starting Gateway
	if there is no Start Event (i.e., there are no incoming Sequence Flow).\vspace{0.1cm} \\
\noindent \stepcounter{idaxiom} $AX\_\arabic{idaxiom}$ $\property{has\_event\_based\_exclusive\_gateway\_instantiate} \mbox{ has domain }\concept{event\_based\_exclusive\_gateway}$\vspace{0.1cm} \\
\noindent \stepcounter{idaxiom} $AX\_\arabic{idaxiom}$ $\property{has\_event\_based\_exclusive\_gateway\_instantiate} \mbox{ has range }\datatype{xsd:boolean}$\vspace{0.1cm} \\
\vspace{0.4cm} \hrule \vspace{0.2cm} \noindent \textbf{Class}: $\concept{inclusive\_gateway}$\vspace{0.2cm} \hrule \vspace{0.2cm} 
\noindent \textbf{Label}: Inclusive Gateway\vspace{0.1cm} \\
\noindent \textbf{Description}: Inclusive Gateway\vspace{0.1cm} \\
\noindent \stepcounter{idaxiom} $AX\_\arabic{idaxiom}$ $\concept{inclusive\_gateway} \sqsubseteq (\geq1) \property{has\_inclusive\_gateway\_default\_gate}$\vspace{0.1cm} \\
\par \vspace{-0.1cm} \noindent \textbf{Property}: $\property{has\_inclusive\_gateway\_default\_gate}$\vspace{0.1cm} \\
\noindent \textbf{Label}: DefaultGate\vspace{0.1cm} \\
\noindent \textbf{Description}: A Default Gate MAY be 
	specified (see Section B.11.9, "Gate," on page 274).\vspace{0.1cm} \\
\noindent \stepcounter{idaxiom} $AX\_\arabic{idaxiom}$ $\property{has\_inclusive\_gateway\_default\_gate} \mbox{ has domain }\concept{inclusive\_gateway}$\vspace{0.1cm} \\
\noindent \stepcounter{idaxiom} $AX\_\arabic{idaxiom}$ $\property{has\_inclusive\_gateway\_default\_gate} \mbox{ has range }\concept{gate}$\vspace{0.1cm} \\
\vspace{0.4cm} \hrule \vspace{0.2cm} \noindent \textbf{Class}: $\concept{complex\_gateway}$\vspace{0.2cm} \hrule \vspace{0.2cm} 
\noindent \textbf{Label}: Complex Gateway\vspace{0.1cm} \\
\noindent \textbf{Description}: Complex Gateway\vspace{0.1cm} \\
\noindent \stepcounter{idaxiom} $AX\_\arabic{idaxiom}$ $\concept{complex\_gateway} \sqsubseteq (\geq1) \property{has\_complex\_gateway\_incoming\_condition}$\vspace{0.1cm} \\
\noindent \stepcounter{idaxiom} $AX\_\arabic{idaxiom}$ $\concept{complex\_gateway} \sqsubseteq (\geq1) \property{has\_sequence\_flow\_target\_ref\_inv} \sqcup ((\leq2) \property{has\_sequence\_flow\_target\_ref\_inv} \sqcap \exists\property{has\_complex\_gateway\_incoming\_condition}.\concept{expression})$\vspace{0.1cm} \\
\par \vspace{-0.1cm} \noindent \textbf{Property}: $\property{has\_complex\_gateway\_incoming\_condition}$\vspace{0.1cm} \\
\noindent \textbf{Label}: Incoming Condition\vspace{0.1cm} \\
\noindent \textbf{Description}: If there are Multiple 
	incoming Sequence Flow, an IncomingCondition expression MUST be set by 
	the modeler. This will consist of an expression that can reference
	Sequence Flow names and or Process Properties (Data).\vspace{0.1cm} \\
\noindent \stepcounter{idaxiom} $AX\_\arabic{idaxiom}$ $\property{has\_complex\_gateway\_incoming\_condition} \mbox{ has domain }\concept{complex\_gateway}$\vspace{0.1cm} \\
\noindent \stepcounter{idaxiom} $AX\_\arabic{idaxiom}$ $\property{has\_complex\_gateway\_incoming\_condition} \mbox{ has range }\concept{expression}$\vspace{0.1cm} \\
\noindent \stepcounter{idaxiom} $AX\_\arabic{idaxiom}$ $\concept{complex\_gateway} \sqsubseteq (\geq1) \property{has\_complex\_gateway\_outgoing\_condition}$\vspace{0.1cm} \\
\noindent \stepcounter{idaxiom} $AX\_\arabic{idaxiom}$ $\concept{complex\_gateway} \sqsubseteq (\geq1) \property{has\_sequence\_flow\_source\_ref\_inv} \sqcup ((\leq2) \property{has\_sequence\_flow\_source\_ref\_inv} \sqcap \exists\property{has\_complex\_gateway\_outgoing\_condition}.\concept{expression})$\vspace{0.1cm} \\
\par \vspace{-0.1cm} \noindent \textbf{Property}: $\property{has\_complex\_gateway\_outgoing\_condition}$\vspace{0.1cm} \\
\noindent \textbf{Label}: Outgoing Condition\vspace{0.1cm} \\
\noindent \textbf{Description}: If there are Multiple 
	outgoing Sequence Flow, an OutgoingCondition expression MUST be set by 
	the modeler. This will consist of an expression that can reference
	(outgoing) Sequence Flow Ids and or Process Properties (Data).\vspace{0.1cm} \\
\noindent \stepcounter{idaxiom} $AX\_\arabic{idaxiom}$ $\property{has\_complex\_gateway\_outgoing\_condition} \mbox{ has domain }\concept{complex\_gateway}$\vspace{0.1cm} \\
\noindent \stepcounter{idaxiom} $AX\_\arabic{idaxiom}$ $\property{has\_complex\_gateway\_outgoing\_condition} \mbox{ has range }\concept{expression}$\vspace{0.1cm} \\
\vspace{0.4cm} \hrule \vspace{0.2cm} \noindent \textbf{Class}: $\concept{parallel\_gateway}$\vspace{0.2cm} \hrule \vspace{0.2cm} 
\noindent \textbf{Label}: Parallel Gateway\vspace{0.1cm} \\
\noindent \textbf{Description}: Parallel Gateway\vspace{0.1cm} \\
\vspace{0.4cm} \hrule \vspace{0.2cm} \noindent \textbf{Class}: $\concept{swimlane}$\vspace{0.2cm} \hrule \vspace{0.2cm} 
\noindent \textbf{Label}: Swimlane\vspace{0.1cm} \\
\noindent \textbf{Description}: There are two ways of grouping the primary modeling
	elements through "swimlane": Pools and Lanes\vspace{0.1cm} \\
\noindent \stepcounter{idaxiom} $AX\_\arabic{idaxiom}$ $\concept{swimlane} \equiv \concept{pool} \sqcup \concept{lane}$\vspace{0.1cm} \\
\noindent \stepcounter{idaxiom} $AX\_\arabic{idaxiom}$ $\concept{pool} \sqsubseteq  \neg \concept{lane}$\vspace{0.1cm} \\
\noindent \stepcounter{idaxiom} $AX\_\arabic{idaxiom}$ $\concept{swimlane} \sqsubseteq  (=1) \property{has\_swimlane\_name}$\vspace{0.1cm} \\
\par \vspace{-0.1cm} \noindent \textbf{Property}: $\property{has\_swimlane\_name}$\vspace{0.1cm} \\
\noindent \textbf{Label}: Name\vspace{0.1cm} \\
\noindent \textbf{Description}: Name is an attribute that is text 
	description of the Swimlane.\vspace{0.1cm} \\
\noindent \stepcounter{idaxiom} $AX\_\arabic{idaxiom}$ $\property{has\_swimlane\_name} \mbox{ has domain }\concept{swimlane}$\vspace{0.1cm} \\
\noindent \stepcounter{idaxiom} $AX\_\arabic{idaxiom}$ $\property{has\_swimlane\_name} \mbox{ has range }\datatype{xsd:string}$\vspace{0.1cm} \\
\vspace{0.4cm} \hrule \vspace{0.2cm} \noindent \textbf{Class}: $\concept{pool}$\vspace{0.2cm} \hrule \vspace{0.2cm} 
\noindent \textbf{Label}: Pool\vspace{0.1cm} \\
\noindent \textbf{Description}: A Pool represents a Participant in a Process. It is
                      also acts as a "swimlane" and a graphical container
                      for partitioning a set of activities from other Pools,
                      usually in the context of B2B situations.\vspace{0.1cm} \\
\noindent \stepcounter{idaxiom} $AX\_\arabic{idaxiom}$ $\concept{pool} \sqsubseteq (\geq1) \property{has\_pool\_process\_ref}$\vspace{0.1cm} \\
\par \vspace{-0.1cm} \noindent \textbf{Property}: $\property{has\_pool\_process\_ref}$\vspace{0.1cm} \\
\noindent \textbf{Label}: ProcessRef\vspace{0.1cm} \\
\noindent \textbf{Description}: The ProcessRef attribute defines the 
	Process that is contained within the Pool. Each Pool MAY have a Process.
	The attributes for a Process can be found in "These attributes are used 
	for Graphical Elements (which are Flow Objects (Section B.4,"Common Flow
	Object Attributes," on page 243), Connecting Objects (Section B.10, 
	"Graphical Connecting Objects," on page 263), Swimlanes (Section B.8, 
	"Swimlanes (Pools and Lanes)," on page 259), and Artifacts (Section B.9,
	"Artifacts," on page 260)), and Supporting Elements (Section B.11,
	"Supporting Elements," on page 266). on page 241."\vspace{0.1cm} \\
\noindent \stepcounter{idaxiom} $AX\_\arabic{idaxiom}$ $\property{has\_pool\_process\_ref} \mbox{ has domain }\concept{pool}$\vspace{0.1cm} \\
\noindent \stepcounter{idaxiom} $AX\_\arabic{idaxiom}$ $\property{has\_pool\_process\_ref} \mbox{ has range }\concept{process}$\vspace{0.1cm} \\
\noindent \stepcounter{idaxiom} $AX\_\arabic{idaxiom}$ $\concept{pool} \sqsubseteq  (=1) \property{has\_pool\_participant\_ref}$\vspace{0.1cm} \\
\par \vspace{-0.1cm} \noindent \textbf{Property}: $\property{has\_pool\_participant\_ref}$\vspace{0.1cm} \\
\noindent \textbf{Label}: ParticipantRef\vspace{0.1cm} \\
\noindent \textbf{Description}: The Modeler MUST define the 
	Participant for a Pool. The Participant can be either a Role or an
	Entity. The attributes for a Participant can be found in "Participant on
	page 276."\vspace{0.1cm} \\
\noindent \stepcounter{idaxiom} $AX\_\arabic{idaxiom}$ $\property{has\_pool\_participant\_ref} \mbox{ has domain }\concept{pool}$\vspace{0.1cm} \\
\noindent \stepcounter{idaxiom} $AX\_\arabic{idaxiom}$ $\property{has\_pool\_participant\_ref} \mbox{ has range }\concept{participant}$\vspace{0.1cm} \\
\noindent \stepcounter{idaxiom} $AX\_\arabic{idaxiom}$ $\concept{pool} \sqsubseteq (\leq1) \property{has\_pool\_lanes}$\vspace{0.1cm} \\
\par \vspace{-0.1cm} \noindent \textbf{Property}: $\property{has\_pool\_lanes}$\vspace{0.1cm} \\
\noindent \textbf{Label}: Lanes\vspace{0.1cm} \\
\noindent \textbf{Description}: There MUST one or more Lanes within a Pool. If 
	there is only one Lane, then that Lane shares the name of the Pool and 
	only the Pool name is displayed. If there is more than one Lane, then 
	each Lane has to have its own name and all names are displayed. The 
	attributes for a Lane can be found in "Lane on page 89."\vspace{0.1cm} \\
\noindent \stepcounter{idaxiom} $AX\_\arabic{idaxiom}$ $\property{has\_pool\_lanes} \mbox{ has domain }\concept{pool}$\vspace{0.1cm} \\
\noindent \stepcounter{idaxiom} $AX\_\arabic{idaxiom}$ $\property{has\_pool\_lanes} \mbox{ has range }\concept{lane}$\vspace{0.1cm} \\
\noindent \stepcounter{idaxiom} $AX\_\arabic{idaxiom}$ $\concept{pool} \sqsubseteq  (=1) \property{has\_pool\_boundary\_visible}$\vspace{0.1cm} \\
\par \vspace{-0.1cm} \noindent \textbf{Property}: $\property{has\_pool\_boundary\_visible}$\vspace{0.1cm} \\
\noindent \textbf{Label}: boundary\_visible\vspace{0.1cm} \\
\noindent \textbf{Description}: This attribute defines if the 
	rectangular boundary for the Pool is visible. Only one Pool in the 
	Diagram MAY have the attribute set to False.\vspace{0.1cm} \\
\noindent \stepcounter{idaxiom} $AX\_\arabic{idaxiom}$ $\property{has\_pool\_boundary\_visible} \mbox{ has domain }\concept{pool}$\vspace{0.1cm} \\
\noindent \stepcounter{idaxiom} $AX\_\arabic{idaxiom}$ $\property{has\_pool\_boundary\_visible} \mbox{ has range }\datatype{xsd:boolean}$\vspace{0.1cm} \\
\noindent \stepcounter{idaxiom} $AX\_\arabic{idaxiom}$ $\concept{pool} \sqsubseteq  (=1) \property{has\_pool\_main\_pool}$\vspace{0.1cm} \\
\par \vspace{-0.1cm} \noindent \textbf{Property}: $\property{has\_pool\_main\_pool}$\vspace{0.1cm} \\
\noindent \textbf{Label}: main\_pool\vspace{0.1cm} \\
\noindent \textbf{Description}: This attribute defines if the Pool is the 
	"main" Pool or the focus of the diagram. Only one Pool in the Diagram 
	MAY have the attribute set to True.\vspace{0.1cm} \\
\noindent \stepcounter{idaxiom} $AX\_\arabic{idaxiom}$ $\property{has\_pool\_main\_pool} \mbox{ has domain }\concept{pool}$\vspace{0.1cm} \\
\noindent \stepcounter{idaxiom} $AX\_\arabic{idaxiom}$ $\property{has\_pool\_main\_pool} \mbox{ has range }\datatype{xsd:boolean}$\vspace{0.1cm} \\
\vspace{0.4cm} \hrule \vspace{0.2cm} \noindent \textbf{Class}: $\concept{lane}$\vspace{0.2cm} \hrule \vspace{0.2cm} 
\noindent \textbf{Label}: Lane\vspace{0.1cm} \\
\noindent \textbf{Description}: A Lane is a sub-partition within a Pool and will extend 
	the entire length of the Pool, either vertically or horizontally. Lanes 
	are used to organize and categorize activities.\vspace{0.1cm} \\
\par \vspace{-0.1cm} \noindent \textbf{Property}: $\property{has\_lane\_lanes}$\vspace{0.1cm} \\
\noindent \textbf{Label}: Lanes\vspace{0.1cm} \\
\noindent \textbf{Description}: This attribute identifies any Lanes that are 
	nested within the current Lane.\vspace{0.1cm} \\
\noindent \stepcounter{idaxiom} $AX\_\arabic{idaxiom}$ $\property{has\_lane\_lanes} \mbox{ has domain }\concept{lane}$\vspace{0.1cm} \\
\noindent \stepcounter{idaxiom} $AX\_\arabic{idaxiom}$ $\property{has\_lane\_lanes} \mbox{ has range }\concept{lane}$\vspace{0.1cm} \\
\vspace{0.4cm} \hrule \vspace{0.2cm} \noindent \textbf{Class}: $\concept{artifact}$\vspace{0.2cm} \hrule \vspace{0.2cm} 
\noindent \textbf{Label}: Artifact\vspace{0.1cm} \\
\noindent \textbf{Description}: Artifacts are used to provide additional information 
	about the Process. There are three standardized Artifacts, but modelers
	or modeling tools are free to add as many Artifacts as required. There 
	may be addition BPMN efforts to standardize a larger set of Artifacts 
	for general use or for vertical markets. The current set of Artifacts 
	include: Data Object, Group, Annotation\vspace{0.1cm} \\
\noindent \stepcounter{idaxiom} $AX\_\arabic{idaxiom}$ $\concept{artifact} \equiv \concept{data\_object} \sqcup (\concept{group} \sqcup \concept{annotation})$\vspace{0.1cm} \\
\noindent \stepcounter{idaxiom} $AX\_\arabic{idaxiom}$ $\concept{data\_object} \sqsubseteq  \neg \concept{group}$\vspace{0.1cm} \\
\noindent \stepcounter{idaxiom} $AX\_\arabic{idaxiom}$ $\concept{data\_object} \sqsubseteq  \neg \concept{annotation}$\vspace{0.1cm} \\
\noindent \stepcounter{idaxiom} $AX\_\arabic{idaxiom}$ $\concept{group} \sqsubseteq  \neg \concept{annotation}$\vspace{0.1cm} \\
\noindent \stepcounter{idaxiom} $AX\_\arabic{idaxiom}$ $\concept{artifact} \sqsubseteq  (=1) \property{has\_artifact\_type}$\vspace{0.1cm} \\
\par \vspace{-0.1cm} \noindent \textbf{Property}: $\property{has\_artifact\_type}$\vspace{0.1cm} \\
\noindent \textbf{Label}: Name\vspace{0.1cm} \\
\noindent \textbf{Description}: The ArtifactType MAY be set to DataObject, 
	Group, or Annotation. 
	The ArtifactType list MAY be extended to include new types.\vspace{0.1cm} \\
\noindent \stepcounter{idaxiom} $AX\_\arabic{idaxiom}$ $\property{has\_artifact\_type} \mbox{ has domain }\concept{artifact}$\vspace{0.1cm} \\
\noindent \stepcounter{idaxiom} $AX\_\arabic{idaxiom}$ $\property{has\_artifact\_type} \mbox{ has range }\concept{artifact\_types}$\vspace{0.1cm} \\
\noindent \stepcounter{idaxiom} $AX\_\arabic{idaxiom}$ $\concept{artifact\_types} \equiv \{\instance{data\_object\_artifact\_type}, \instance{group\_artifact\_type}, \instance{annotation\_artifact\_type}\}$\vspace{0.1cm} \\
\par \vspace{-0.1cm} \noindent \textbf{Instance}: $\instance{data\_object\_artifact\_type}$\vspace{0.1cm} \\
\noindent \textbf{Label}: Data Object\vspace{0.1cm} \\
\par \vspace{-0.1cm} \noindent \textbf{Instance}: $\instance{group\_artifact\_type}$\vspace{0.1cm} \\
\noindent \textbf{Label}: Group\vspace{0.1cm} \\
\par \vspace{-0.1cm} \noindent \textbf{Instance}: $\instance{annotation\_artifact\_type}$\vspace{0.1cm} \\
\noindent \textbf{Label}: Annotation\vspace{0.1cm} \\
\noindent \stepcounter{idaxiom} $AX\_\arabic{idaxiom}$ $\concept{data\_object} \equiv \concept{artifact} \sqcap \exists\property{has\_artifact\_type}.\{\instance{data\_object\_artifact\_type}\}$\vspace{0.1cm} \\
\noindent \stepcounter{idaxiom} $AX\_\arabic{idaxiom}$ $\concept{group} \equiv \concept{artifact} \sqcap \exists\property{has\_artifact\_type}.\{\instance{group\_artifact\_type}\}$\vspace{0.1cm} \\
\noindent \stepcounter{idaxiom} $AX\_\arabic{idaxiom}$ $\concept{annotation} \equiv \concept{artifact} \sqcap \exists\property{has\_artifact\_type}.\{\instance{annotation\_artifact\_type}\}$\vspace{0.1cm} \\
\vspace{0.4cm} \hrule \vspace{0.2cm} \noindent \textbf{Class}: $\concept{data\_object}$\vspace{0.2cm} \hrule \vspace{0.2cm} 
\noindent \textbf{Label}: Data Object\vspace{0.1cm} \\
\noindent \textbf{Description}: Data Objects are considered Artifacts because they
	do not have any direct effect on the Sequence Flow or Message Flow of 
	the Process, but they do provide information about what activities 
	require to be performed and/or what they produce.\vspace{0.1cm} \\
\noindent \stepcounter{idaxiom} $AX\_\arabic{idaxiom}$ $\concept{data\_object} \sqsubseteq  (=1) \property{has\_data\_object\_name}$\vspace{0.1cm} \\
\par \vspace{-0.1cm} \noindent \textbf{Property}: $\property{has\_data\_object\_name}$\vspace{0.1cm} \\
\noindent \textbf{Label}: Name\vspace{0.1cm} \\
\noindent \textbf{Description}: Name is an attribute that is text 
	description of the object.\vspace{0.1cm} \\
\noindent \stepcounter{idaxiom} $AX\_\arabic{idaxiom}$ $\property{has\_data\_object\_name} \mbox{ has domain }\concept{data\_object}$\vspace{0.1cm} \\
\noindent \stepcounter{idaxiom} $AX\_\arabic{idaxiom}$ $\property{has\_data\_object\_name} \mbox{ has range }\datatype{xsd:string}$\vspace{0.1cm} \\
\noindent \stepcounter{idaxiom} $AX\_\arabic{idaxiom}$ $\concept{data\_object} \sqsubseteq (\geq1) \property{has\_data\_object\_state}$\vspace{0.1cm} \\
\par \vspace{-0.1cm} \noindent \textbf{Property}: $\property{has\_data\_object\_state}$\vspace{0.1cm} \\
\noindent \textbf{Label}: State\vspace{0.1cm} \\
\noindent \textbf{Description}: State is an optional attribute that 
	indicates the impact the Process has had on the Data Object. Multiple 
	Data Objects with the same name MAY share the same state within one 
	Process.\vspace{0.1cm} \\
\noindent \stepcounter{idaxiom} $AX\_\arabic{idaxiom}$ $\property{has\_data\_object\_state} \mbox{ has domain }\concept{data\_object}$\vspace{0.1cm} \\
\noindent \stepcounter{idaxiom} $AX\_\arabic{idaxiom}$ $\property{has\_data\_object\_state} \mbox{ has range }\datatype{xsd:string}$\vspace{0.1cm} \\
\par \vspace{-0.1cm} \noindent \textbf{Property}: $\property{has\_data\_object\_properties}$\vspace{0.1cm} \\
\noindent \textbf{Label}: Properties\vspace{0.1cm} \\
\noindent \textbf{Description}: Modeler-defined Properties MAY be 
	added to a Data Object. The fully delineated name of these properties 
	are "process name.task name.property name" (e.g., 
	"Add Customer.Review Credit Report.Score"). Further details about the
	definition of a Property can be found in "Property on page 276."\vspace{0.1cm} \\
\noindent \stepcounter{idaxiom} $AX\_\arabic{idaxiom}$ $\property{has\_data\_object\_properties} \mbox{ has domain }\concept{data\_object}$\vspace{0.1cm} \\
\noindent \stepcounter{idaxiom} $AX\_\arabic{idaxiom}$ $\property{has\_data\_object\_properties} \mbox{ has range }\concept{property}$\vspace{0.1cm} \\
\vspace{0.4cm} \hrule \vspace{0.2cm} \noindent \textbf{Class}: $\concept{annotation}$\vspace{0.2cm} \hrule \vspace{0.2cm} 
\noindent \textbf{Label}: Annotation\vspace{0.1cm} \\
\noindent \textbf{Description}: Text Annotations are a mechanism for a modeler to
	(attached with an provide additional information for the reader of a 
	Association) BPMN Diagram.\vspace{0.1cm} \\
\noindent \stepcounter{idaxiom} $AX\_\arabic{idaxiom}$ $\concept{annotation} \sqsubseteq  (=1) \property{has\_annotation\_text}$\vspace{0.1cm} \\
\par \vspace{-0.1cm} \noindent \textbf{Property}: $\property{has\_annotation\_text}$\vspace{0.1cm} \\
\noindent \textbf{Label}: Text\vspace{0.1cm} \\
\noindent \textbf{Description}: Text is an attribute that is text that 
	the modeler wishes to communicate to the reader	of the Diagram.\vspace{0.1cm} \\
\noindent \stepcounter{idaxiom} $AX\_\arabic{idaxiom}$ $\property{has\_annotation\_text} \mbox{ has domain }\concept{annotation}$\vspace{0.1cm} \\
\noindent \stepcounter{idaxiom} $AX\_\arabic{idaxiom}$ $\property{has\_annotation\_text} \mbox{ has range }\datatype{xsd:string}$\vspace{0.1cm} \\
\vspace{0.4cm} \hrule \vspace{0.2cm} \noindent \textbf{Class}: $\concept{group}$\vspace{0.2cm} \hrule \vspace{0.2cm} 
\noindent \textbf{Label}: Group\vspace{0.1cm} \\
\noindent \textbf{Description}: A grouping of activities that does not affect the 
	Sequence Flow. The grouping can be used for documentation or analysis 
	purposes. Groups can also be used to identify the activities of a
	distributed transaction that is shown across Pools.\vspace{0.1cm} \\
\noindent \stepcounter{idaxiom} $AX\_\arabic{idaxiom}$ $\concept{group} \sqsubseteq  (=1) \property{has\_group\_category\_ref}$\vspace{0.1cm} \\
\par \vspace{-0.1cm} \noindent \textbf{Property}: $\property{has\_group\_category\_ref}$\vspace{0.1cm} \\
\noindent \textbf{Label}: CategoryRef\vspace{0.1cm} \\
\noindent \textbf{Description}: CategoryRef specifies the Category that
	the Group represents (Further details about the definition of a Category
	can be found in "Category on page 269."). The name of the Category 
	provides the label for the Group. The graphical elements within the 
	boundaries of the Group will be assigned the Category.\vspace{0.1cm} \\
\noindent \stepcounter{idaxiom} $AX\_\arabic{idaxiom}$ $\property{has\_group\_category\_ref} \mbox{ has domain }\concept{group}$\vspace{0.1cm} \\
\noindent \stepcounter{idaxiom} $AX\_\arabic{idaxiom}$ $\property{has\_group\_category\_ref} \mbox{ has range }\concept{category}$\vspace{0.1cm} \\
\par \vspace{-0.1cm} \noindent \textbf{Property}: $\property{has\_group\_graphical\_element}$\vspace{0.1cm} \\
\noindent \textbf{Label}: GraphicalElement\vspace{0.1cm} \\
\noindent \textbf{Description}: The GraphicalElements attribute 
	identifies all of the graphical elements (e.g., Events, Activities, 
	Gateways, and Artifacts) that are within the boundaries of the Group.\vspace{0.1cm} \\
\noindent \stepcounter{idaxiom} $AX\_\arabic{idaxiom}$ $\property{has\_group\_graphical\_element} \mbox{ has domain }\concept{group}$\vspace{0.1cm} \\
\noindent \stepcounter{idaxiom} $AX\_\arabic{idaxiom}$ $\property{has\_group\_graphical\_element} \mbox{ has range }\concept{graphical\_element}$\vspace{0.1cm} \\
\vspace{0.4cm} \hrule \vspace{0.2cm} \noindent \textbf{Class}: $\concept{connecting\_object}$\vspace{0.2cm} \hrule \vspace{0.2cm} 
\noindent \textbf{Label}: Connecting object\vspace{0.1cm} \\
\noindent \textbf{Description}: There are three ways of connecting the Flow 
	Objects to each other or other information. There are three Connecting
	Objects: Sequence Flow, Message Flow, and Association\vspace{0.1cm} \\
\noindent \stepcounter{idaxiom} $AX\_\arabic{idaxiom}$ $\concept{connecting\_object} \equiv \concept{sequence\_flow} \sqcup (\concept{message\_flow} \sqcup \concept{association})$\vspace{0.1cm} \\
\noindent \stepcounter{idaxiom} $AX\_\arabic{idaxiom}$ $\concept{sequence\_flow} \sqsubseteq  \neg \concept{message\_flow}$\vspace{0.1cm} \\
\noindent \stepcounter{idaxiom} $AX\_\arabic{idaxiom}$ $\concept{sequence\_flow} \sqsubseteq  \neg \concept{association}$\vspace{0.1cm} \\
\noindent \stepcounter{idaxiom} $AX\_\arabic{idaxiom}$ $\concept{message\_flow} \sqsubseteq  \neg \concept{association}$\vspace{0.1cm} \\
\noindent \stepcounter{idaxiom} $AX\_\arabic{idaxiom}$ $\concept{connecting\_object} \sqsubseteq (\geq1) \property{has\_connecting\_object\_name}$\vspace{0.1cm} \\
\par \vspace{-0.1cm} \noindent \textbf{Property}: $\property{has\_connecting\_object\_name}$\vspace{0.1cm} \\
\noindent \textbf{Label}: Name\vspace{0.1cm} \\
\noindent \textbf{Description}: Name is an attribute that is text 
	description of the object.\vspace{0.1cm} \\
\noindent \stepcounter{idaxiom} $AX\_\arabic{idaxiom}$ $\property{has\_connecting\_object\_name} \mbox{ has domain }\concept{connecting\_object}$\vspace{0.1cm} \\
\noindent \stepcounter{idaxiom} $AX\_\arabic{idaxiom}$ $\property{has\_connecting\_object\_name} \mbox{ has range }\datatype{xsd:string}$\vspace{0.1cm} \\
\noindent \stepcounter{idaxiom} $AX\_\arabic{idaxiom}$ $\concept{connecting\_object} \sqsubseteq  (=1) \property{has\_connecting\_object\_source\_ref}$\vspace{0.1cm} \\
\par \vspace{-0.1cm} \noindent \textbf{Property}: $\property{has\_connecting\_object\_source\_ref}$\vspace{0.1cm} \\
\noindent \textbf{Label}: SourceRef\vspace{0.1cm} \\
\noindent \textbf{Description}: SourceRef is an attribute 
	that identifies which Graphical Element the Connecting Object is 
	connected from. Note: there are restrictions as to what objects Sequence
	Flow and Message Flow can connect. Refer to the Sequence Flow 
	Connections section and the Message Flow Connections section for each 
	Flow Object, Swimlane, and Artifact.\vspace{0.1cm} \\
\noindent \stepcounter{idaxiom} $AX\_\arabic{idaxiom}$ $\property{has\_connecting\_object\_source\_ref} \mbox{ has domain }\concept{connecting\_object}$\vspace{0.1cm} \\
\noindent \stepcounter{idaxiom} $AX\_\arabic{idaxiom}$ $\property{has\_connecting\_object\_source\_ref} \mbox{ has range }\concept{graphical\_element}$\vspace{0.1cm} \\
\noindent \stepcounter{idaxiom} $AX\_\arabic{idaxiom}$ $\concept{connecting\_object} \sqsubseteq  (=1) \property{has\_connecting\_object\_target\_ref}$\vspace{0.1cm} \\
\par \vspace{-0.1cm} \noindent \textbf{Property}: $\property{has\_connecting\_object\_target\_ref}$\vspace{0.1cm} \\
\noindent \textbf{Label}: TargetRef\vspace{0.1cm} \\
\noindent \textbf{Description}: Target is an attribute that 
	identifies which Graphical Element the Connecting Object is connected 
	to. Note: there are restrictions as to what objects Sequence Flow and
	Message Flow can connect. Refer to the Sequence Flow Connections section
	and the Message Flow Connections section for each Flow Object, Swimlane,
	and Artifact.\vspace{0.1cm} \\
\noindent \stepcounter{idaxiom} $AX\_\arabic{idaxiom}$ $\property{has\_connecting\_object\_target\_ref} \mbox{ has domain }\concept{connecting\_object}$\vspace{0.1cm} \\
\noindent \stepcounter{idaxiom} $AX\_\arabic{idaxiom}$ $\property{has\_connecting\_object\_target\_ref} \mbox{ has range }\concept{graphical\_element}$\vspace{0.1cm} \\
\noindent \stepcounter{idaxiom} $AX\_\arabic{idaxiom}$ $\property{has\_connecting\_object\_source\_ref\_inv} =\property{has\_connecting\_object\_source\_ref}^{-1}$\vspace{0.1cm} \\
\noindent \stepcounter{idaxiom} $AX\_\arabic{idaxiom}$ $\property{has\_connecting\_object\_target\_ref\_inv} =\property{has\_connecting\_object\_target\_ref}^{-1}$\vspace{0.1cm} \\
\vspace{0.4cm} \hrule \vspace{0.2cm} \noindent \textbf{Class}: $\concept{sequence\_flow}$\vspace{0.2cm} \hrule \vspace{0.2cm} 
\noindent \textbf{Label}: Sequence Flow\vspace{0.1cm} \\
\noindent \textbf{Description}: A Sequence Flow is used to show the order that
              activities will be performed in a Process.\vspace{0.1cm} \\
\noindent \stepcounter{idaxiom} $AX\_\arabic{idaxiom}$ $\concept{sequence\_flow} \sqsubseteq  (=1) \property{has\_sequence\_flow\_condition\_type}$\vspace{0.1cm} \\
\par \vspace{-0.1cm} \noindent \textbf{Property}: $\property{has\_sequence\_flow\_condition\_type}$\vspace{0.1cm} \\
\noindent \textbf{Label}: Condition Type\vspace{0.1cm} \\
\noindent \textbf{Description}: By default, the ConditionType
	of a Sequence Flow is None. This means that there is no evaluation at 
	runtime to determine whether or not the Sequence Flow will be used. Once
	a Token is ready to traverse the Sequence Flow (i.e., the Source is an
	activity that has completed), then the Token will do so. The normal, 
	uncontrolled use of Sequence Flow, in a sequence of activities, will 
	have a None ConditionType (see Figure 10.1). A None ConditionType MUST 
	NOT be used if the Source of the Sequence Flow is an Exclusive 
	Data-Based or Inclusive Gateway. 
	The ConditionType attribute MAY be set to Expression if the Source of 
	the Sequence Flow is a Task, a Sub-Process, or a Gateway of type 
	Exclusive-Data-Based or Inclusive.
	If the ConditionType attribute is set to Expression, then a condition 
	marker SHALL be added to the line if the Sequence Flow is outgoing from 
	an activity (see Figure 10.2). However, a condition indicator MUST NOT 
	be added to the line if the Sequence Flow is outgoing from a Gateway.
	An Expression ConditionType MUST NOT be used if the Source of the 
	Sequence Flow is an Event-Based Exclusive Gateway, a Complex Gateway, a 
	Parallel Gateway, a Start Event, or an Intermediate Event. In addition, 
	an Expression ConditionType MUST NOT be used if the Sequence Flow is 
	associated with the Default Gate of a Gateway.
	The ConditionType attribute MAY be set to Default only if the Source of 
	the Sequence Flow is an activity or an Exclusive Data-Based Gateway. If
	the ConditionType is Default, then the Default marker SHALL be displayed
	(see Figure 10.3).\vspace{0.1cm} \\
\noindent \stepcounter{idaxiom} $AX\_\arabic{idaxiom}$ $\property{has\_sequence\_flow\_condition\_type} \mbox{ has domain }\concept{sequence\_flow}$\vspace{0.1cm} \\
\noindent \stepcounter{idaxiom} $AX\_\arabic{idaxiom}$ $\property{has\_sequence\_flow\_condition\_type} \mbox{ has range }\datatype{xsd:string}\{"\datainstance{None}","\datainstance{Expression}","\datainstance{Default}"\}$\vspace{0.1cm} \\
\noindent \stepcounter{idaxiom} $AX\_\arabic{idaxiom}$ $\concept{sequence\_flow} \sqsubseteq ( \neg \exists\property{has\_sequence\_flow\_condition\_type}.\{"\datainstance{Expression}"\}) \sqcup \\ ((\exists\property{has\_sequence\_flow\_condition\_type}.\{"\datainstance{Expression}"\}) \sqcap ( (=1) \property{has\_sequence\_flow\_condition\_expression}))$\vspace{0.1cm} \\
\par \vspace{-0.1cm} \noindent \textbf{Property}: $\property{has\_sequence\_flow\_condition\_expression}$\vspace{0.1cm} \\
\noindent \textbf{Label}: Condition Expression\vspace{0.1cm} \\
\noindent \textbf{Description}: If the ConditionType 
	attribute is set to Expression, then the ConditionExpression attribute 
	MUST be defined as a valid expression. The expression will be evaluated
	at runtime. If the result of the evaluation is TRUE, then a Token will 
	be generated and will traverse the Sequence--Subject to any constraints 
	imposed by a Source that is a Gateway.\vspace{0.1cm} \\
\noindent \stepcounter{idaxiom} $AX\_\arabic{idaxiom}$ $\property{has\_sequence\_flow\_condition\_expression} \mbox{ has domain }\concept{sequence\_flow}$\vspace{0.1cm} \\
\noindent \stepcounter{idaxiom} $AX\_\arabic{idaxiom}$ $\property{has\_sequence\_flow\_condition\_expression} \mbox{ has range }\concept{expression}$\vspace{0.1cm} \\
\noindent \stepcounter{idaxiom} $AX\_\arabic{idaxiom}$ $\property{has\_sequence\_flow\_source\_ref} \sqsubseteq\property{has\_connecting\_object\_source\_ref}$\vspace{0.1cm} \\
\noindent \stepcounter{idaxiom} $AX\_\arabic{idaxiom}$ $\property{has\_sequence\_flow\_target\_ref} \sqsubseteq\property{has\_connecting\_object\_target\_ref}$\vspace{0.1cm} \\
\par \vspace{-0.1cm} \noindent \textbf{Property}: $\property{has\_sequence\_flow\_source\_ref}$\vspace{0.1cm} \\
\noindent \textbf{Label}: SequenceFlow\_SourceRef\vspace{0.1cm} \\
\noindent \textbf{Description}: SourceRef is an attribute 
	that identifies which Graphical Element the Connecting Object is 
	connected from. Note: there are restrictions as to what objects Sequence
	Flow and Message Flow can connect. Refer to the Sequence Flow 
	Connections section and the Message Flow Connections section for each 
	Flow Object, Swimlane, and Artifact.\vspace{0.1cm} \\
\noindent \stepcounter{idaxiom} $AX\_\arabic{idaxiom}$ $\property{has\_sequence\_flow\_source\_ref} \mbox{ has domain }\concept{sequence\_flow}$\vspace{0.1cm} \\
\par \vspace{-0.1cm} \noindent \textbf{Property}: $\property{has\_sequence\_flow\_target\_ref}$\vspace{0.1cm} \\
\noindent \textbf{Label}: SequenceFlow\_TargetRef\vspace{0.1cm} \\
\noindent \textbf{Description}: Target is an attribute that 
	identifies which Graphical Element the Connecting Object is connected 
	to. Note: there are restrictions as to what objects Sequence Flow and
	Message Flow can connect. Refer to the Sequence Flow Connections section
	and the Message Flow Connections section for each Flow Object, Swimlane,
	and Artifact.\vspace{0.1cm} \\
\noindent \stepcounter{idaxiom} $AX\_\arabic{idaxiom}$ $\property{has\_sequence\_flow\_target\_ref} \mbox{ has domain }\concept{sequence\_flow}$\vspace{0.1cm} \\
\noindent \stepcounter{idaxiom} $AX\_\arabic{idaxiom}$ $\property{has\_sequence\_flow\_source\_ref\_inv} =\property{has\_sequence\_flow\_source\_ref}^{-1}$\vspace{0.1cm} \\
\noindent \stepcounter{idaxiom} $AX\_\arabic{idaxiom}$ $\property{has\_sequence\_flow\_target\_ref\_inv} =\property{has\_sequence\_flow\_target\_ref}^{-1}$\vspace{0.1cm} \\
\vspace{0.4cm} \hrule \vspace{0.2cm} \noindent \textbf{Class}: $\concept{message\_flow}$\vspace{0.2cm} \hrule \vspace{0.2cm} 
\noindent \textbf{Label}: Message Flow\vspace{0.1cm} \\
\noindent \textbf{Description}: A Message Flow is used to show the flow of
             messages between two participants that are
             prepared to send and receive them. In BPMN, two
             separate Pools in the Diagram will represent the two
             participants (e.g., business entities or business
             roles).\vspace{0.1cm} \\
\noindent \stepcounter{idaxiom} $AX\_\arabic{idaxiom}$ $\concept{message\_flow} \sqsubseteq (\geq1) \property{has\_message\_flow\_message\_ref}$\vspace{0.1cm} \\
\par \vspace{-0.1cm} \noindent \textbf{Property}: $\property{has\_message\_flow\_message\_ref}$\vspace{0.1cm} \\
\noindent \textbf{Label}: MessageRef\vspace{0.1cm} \\
\noindent \textbf{Description}: MessageRef is an optional 
	attribute that identifies the Message that is being sent. The attributes
	of a Message can be found in "Message on page 275."\vspace{0.1cm} \\
\noindent \stepcounter{idaxiom} $AX\_\arabic{idaxiom}$ $\property{has\_message\_flow\_message\_ref} \mbox{ has domain }\concept{message\_flow}$\vspace{0.1cm} \\
\noindent \stepcounter{idaxiom} $AX\_\arabic{idaxiom}$ $\property{has\_message\_flow\_message\_ref} \mbox{ has range }\concept{message}$\vspace{0.1cm} \\
\noindent \stepcounter{idaxiom} $AX\_\arabic{idaxiom}$ $\property{has\_message\_flow\_source\_ref} \sqsubseteq\property{has\_connecting\_object\_source\_ref}$\vspace{0.1cm} \\
\noindent \stepcounter{idaxiom} $AX\_\arabic{idaxiom}$ $\property{has\_message\_flow\_target\_ref} \sqsubseteq\property{has\_connecting\_object\_target\_ref}$\vspace{0.1cm} \\
\par \vspace{-0.1cm} \noindent \textbf{Property}: $\property{has\_message\_flow\_source\_ref}$\vspace{0.1cm} \\
\noindent \textbf{Label}: MessageFlow\_SourceRef\vspace{0.1cm} \\
\noindent \textbf{Description}: SourceRef is an attribute 
	that identifies which Graphical Element the Connecting Object is 
	connected from. Note: there are restrictions as to what objects Sequence
	Flow and Message Flow can connect. Refer to the Sequence Flow 
	Connections section and the Message Flow Connections section for each 
	Flow Object, Swimlane, and Artifact.\vspace{0.1cm} \\
\noindent \stepcounter{idaxiom} $AX\_\arabic{idaxiom}$ $\property{has\_message\_flow\_source\_ref} \mbox{ has domain }\concept{message\_flow}$\vspace{0.1cm} \\
\par \vspace{-0.1cm} \noindent \textbf{Property}: $\property{has\_message\_flow\_target\_ref}$\vspace{0.1cm} \\
\noindent \textbf{Label}: MessageFlow\_TargetRef\vspace{0.1cm} \\
\noindent \textbf{Description}: Target is an attribute that 
	identifies which Graphical Element the Connecting Object is connected 
	to. Note: there are restrictions as to what objects Sequence Flow and
	Message Flow can connect. Refer to the Sequence Flow Connections section
	and the Message Flow Connections section for each Flow Object, Swimlane,
	and Artifact.\vspace{0.1cm} \\
\noindent \stepcounter{idaxiom} $AX\_\arabic{idaxiom}$ $\property{has\_message\_flow\_target\_ref} \mbox{ has domain }\concept{message\_flow}$\vspace{0.1cm} \\
\noindent \stepcounter{idaxiom} $AX\_\arabic{idaxiom}$ $\property{has\_message\_flow\_source\_ref\_inv} =\property{has\_message\_flow\_source\_ref}^{-1}$\vspace{0.1cm} \\
\noindent \stepcounter{idaxiom} $AX\_\arabic{idaxiom}$ $\property{has\_message\_flow\_target\_ref\_inv} =\property{has\_message\_flow\_target\_ref}^{-1}$\vspace{0.1cm} \\
\vspace{0.4cm} \hrule \vspace{0.2cm} \noindent \textbf{Class}: $\concept{association}$\vspace{0.2cm} \hrule \vspace{0.2cm} 
\noindent \textbf{Label}: Association\vspace{0.1cm} \\
\noindent \textbf{Description}: An Association is used to associate information
            with Flow Objects. Text and graphical non-Flow
            Objects can be associated with the Flow Objects.\vspace{0.1cm} \\
\noindent \stepcounter{idaxiom} $AX\_\arabic{idaxiom}$ $\concept{association} \sqsubseteq  (=1) \property{has\_association\_direction}$\vspace{0.1cm} \\
\par \vspace{-0.1cm} \noindent \textbf{Property}: $\property{has\_association\_direction}$\vspace{0.1cm} \\
\noindent \textbf{Label}: Direction\vspace{0.1cm} \\
\noindent \textbf{Description}: Direction is an attribute that 
	defines whether or not the Association shows any directionality with an 
	arrowhead. The default is None (no arrowhead). A value of One means that
	the arrowhead SHALL be at the Target Object. A value of Both means that 
	there SHALL be an arrowhead at both ends of the Association line.\vspace{0.1cm} \\
\noindent \stepcounter{idaxiom} $AX\_\arabic{idaxiom}$ $\property{has\_association\_direction} \mbox{ has domain }\concept{association}$\vspace{0.1cm} \\
\noindent \stepcounter{idaxiom} $AX\_\arabic{idaxiom}$ $\property{has\_association\_direction} \mbox{ has range }\datatype{xsd:string}\{"\datainstance{None}","\datainstance{One}","\datainstance{Both}"\}$\vspace{0.1cm} \\
\vspace{0.4cm} \hrule \vspace{0.2cm} \noindent \textbf{Class}: $\concept{supporting\_element}$\vspace{0.2cm} \hrule \vspace{0.2cm} 
\noindent \textbf{Label}: Supporting Element\vspace{0.1cm} \\
\noindent \textbf{Description}: Supporting Element is one of two main 
	elements that are of type BPMN Element (see Figure B.1). The other is 
	Graphical Element. There are 16 types, and a few subtypes, of Support 
	Element. These are: These are: Assignments (see Section B.11.3 on page 
	269), Categories (see Section B.11.4 on page 269), Entities (see Section
	B.11.5 on page 269), Event Details (see Section B.11.7 on page 270), 
	Expressions (see Section B.11.8 on page 273), Gates (see Section B.11.9
	on page 274), Inputs (see Section B.11.10 on page 274), Messages (see 
	Section B.11.11 on page 275), Outputs (see Section B.11.13 on page 275),
	Participants (see Section B.11.14 on page 276), Processes (see Section 
	B.3 on page 242), Properties (see Section B.11.15 on page 276), Roles 
	(see Section B.11.16 on page 276), Conditions (see Section B.11.5 on 
	page 269), Transactions (see Section B.11.19 on page 277), and Web 
	Services (see Section B.11.20 on page 277).\vspace{0.1cm} \\
\noindent \stepcounter{idaxiom} $AX\_\arabic{idaxiom}$ $\concept{supporting\_element} \equiv \concept{process} \sqcup \concept{message} \sqcup \concept{condition} \sqcup \concept{event\_detail} \sqcup \concept{assignment} \sqcup \concept{expression} \sqcup \concept{property} \sqcup \concept{transaction} \sqcup \concept{gate} \sqcup \concept{web\_service} \sqcup \concept{role} \sqcup \concept{entity} \sqcup \concept{participant} \sqcup \concept{category} \sqcup \concept{output\_set} \sqcup \concept{input\_set}$\vspace{0.1cm} \\
\noindent \stepcounter{idaxiom} $AX\_\arabic{idaxiom}$ $\concept{process} \sqsubseteq  \neg \concept{message}$\vspace{0.1cm} \\
\noindent \stepcounter{idaxiom} $AX\_\arabic{idaxiom}$ $\concept{process} \sqsubseteq  \neg \concept{condition}$\vspace{0.1cm} \\
\noindent \stepcounter{idaxiom} $AX\_\arabic{idaxiom}$ $\concept{process} \sqsubseteq  \neg \concept{event\_detail}$\vspace{0.1cm} \\
\noindent \stepcounter{idaxiom} $AX\_\arabic{idaxiom}$ $\concept{process} \sqsubseteq  \neg \concept{assignment}$\vspace{0.1cm} \\
\noindent \stepcounter{idaxiom} $AX\_\arabic{idaxiom}$ $\concept{process} \sqsubseteq  \neg \concept{expression}$\vspace{0.1cm} \\
\noindent \stepcounter{idaxiom} $AX\_\arabic{idaxiom}$ $\concept{process} \sqsubseteq  \neg \concept{property}$\vspace{0.1cm} \\
\noindent \stepcounter{idaxiom} $AX\_\arabic{idaxiom}$ $\concept{process} \sqsubseteq  \neg \concept{transaction}$\vspace{0.1cm} \\
\noindent \stepcounter{idaxiom} $AX\_\arabic{idaxiom}$ $\concept{process} \sqsubseteq  \neg \concept{gate}$\vspace{0.1cm} \\
\noindent \stepcounter{idaxiom} $AX\_\arabic{idaxiom}$ $\concept{process} \sqsubseteq  \neg \concept{web\_service}$\vspace{0.1cm} \\
\noindent \stepcounter{idaxiom} $AX\_\arabic{idaxiom}$ $\concept{process} \sqsubseteq  \neg \concept{role}$\vspace{0.1cm} \\
\noindent \stepcounter{idaxiom} $AX\_\arabic{idaxiom}$ $\concept{process} \sqsubseteq  \neg \concept{entity}$\vspace{0.1cm} \\
\noindent \stepcounter{idaxiom} $AX\_\arabic{idaxiom}$ $\concept{process} \sqsubseteq  \neg \concept{participant}$\vspace{0.1cm} \\
\noindent \stepcounter{idaxiom} $AX\_\arabic{idaxiom}$ $\concept{process} \sqsubseteq  \neg \concept{category}$\vspace{0.1cm} \\
\noindent \stepcounter{idaxiom} $AX\_\arabic{idaxiom}$ $\concept{process} \sqsubseteq  \neg \concept{output\_set}$\vspace{0.1cm} \\
\noindent \stepcounter{idaxiom} $AX\_\arabic{idaxiom}$ $\concept{process} \sqsubseteq  \neg \concept{input\_set}$\vspace{0.1cm} \\
\noindent \stepcounter{idaxiom} $AX\_\arabic{idaxiom}$ $\concept{message} \sqsubseteq  \neg \concept{condition}$\vspace{0.1cm} \\
\noindent \stepcounter{idaxiom} $AX\_\arabic{idaxiom}$ $\concept{message} \sqsubseteq  \neg \concept{event\_detail}$\vspace{0.1cm} \\
\noindent \stepcounter{idaxiom} $AX\_\arabic{idaxiom}$ $\concept{message} \sqsubseteq  \neg \concept{assignment}$\vspace{0.1cm} \\
\noindent \stepcounter{idaxiom} $AX\_\arabic{idaxiom}$ $\concept{message} \sqsubseteq  \neg \concept{expression}$\vspace{0.1cm} \\
\noindent \stepcounter{idaxiom} $AX\_\arabic{idaxiom}$ $\concept{message} \sqsubseteq  \neg \concept{property}$\vspace{0.1cm} \\
\noindent \stepcounter{idaxiom} $AX\_\arabic{idaxiom}$ $\concept{message} \sqsubseteq  \neg \concept{transaction}$\vspace{0.1cm} \\
\noindent \stepcounter{idaxiom} $AX\_\arabic{idaxiom}$ $\concept{message} \sqsubseteq  \neg \concept{gate}$\vspace{0.1cm} \\
\noindent \stepcounter{idaxiom} $AX\_\arabic{idaxiom}$ $\concept{message} \sqsubseteq  \neg \concept{web\_service}$\vspace{0.1cm} \\
\noindent \stepcounter{idaxiom} $AX\_\arabic{idaxiom}$ $\concept{message} \sqsubseteq  \neg \concept{role}$\vspace{0.1cm} \\
\noindent \stepcounter{idaxiom} $AX\_\arabic{idaxiom}$ $\concept{message} \sqsubseteq  \neg \concept{entity}$\vspace{0.1cm} \\
\noindent \stepcounter{idaxiom} $AX\_\arabic{idaxiom}$ $\concept{message} \sqsubseteq  \neg \concept{participant}$\vspace{0.1cm} \\
\noindent \stepcounter{idaxiom} $AX\_\arabic{idaxiom}$ $\concept{message} \sqsubseteq  \neg \concept{category}$\vspace{0.1cm} \\
\noindent \stepcounter{idaxiom} $AX\_\arabic{idaxiom}$ $\concept{message} \sqsubseteq  \neg \concept{output\_set}$\vspace{0.1cm} \\
\noindent \stepcounter{idaxiom} $AX\_\arabic{idaxiom}$ $\concept{message} \sqsubseteq  \neg \concept{input\_set}$\vspace{0.1cm} \\
\noindent \stepcounter{idaxiom} $AX\_\arabic{idaxiom}$ $\concept{condition} \sqsubseteq  \neg \concept{event\_detail}$\vspace{0.1cm} \\
\noindent \stepcounter{idaxiom} $AX\_\arabic{idaxiom}$ $\concept{condition} \sqsubseteq  \neg \concept{assignment}$\vspace{0.1cm} \\
\noindent \stepcounter{idaxiom} $AX\_\arabic{idaxiom}$ $\concept{condition} \sqsubseteq  \neg \concept{expression}$\vspace{0.1cm} \\
\noindent \stepcounter{idaxiom} $AX\_\arabic{idaxiom}$ $\concept{condition} \sqsubseteq  \neg \concept{property}$\vspace{0.1cm} \\
\noindent \stepcounter{idaxiom} $AX\_\arabic{idaxiom}$ $\concept{condition} \sqsubseteq  \neg \concept{transaction}$\vspace{0.1cm} \\
\noindent \stepcounter{idaxiom} $AX\_\arabic{idaxiom}$ $\concept{condition} \sqsubseteq  \neg \concept{gate}$\vspace{0.1cm} \\
\noindent \stepcounter{idaxiom} $AX\_\arabic{idaxiom}$ $\concept{condition} \sqsubseteq  \neg \concept{web\_service}$\vspace{0.1cm} \\
\noindent \stepcounter{idaxiom} $AX\_\arabic{idaxiom}$ $\concept{condition} \sqsubseteq  \neg \concept{role}$\vspace{0.1cm} \\
\noindent \stepcounter{idaxiom} $AX\_\arabic{idaxiom}$ $\concept{condition} \sqsubseteq  \neg \concept{entity}$\vspace{0.1cm} \\
\noindent \stepcounter{idaxiom} $AX\_\arabic{idaxiom}$ $\concept{condition} \sqsubseteq  \neg \concept{participant}$\vspace{0.1cm} \\
\noindent \stepcounter{idaxiom} $AX\_\arabic{idaxiom}$ $\concept{condition} \sqsubseteq  \neg \concept{category}$\vspace{0.1cm} \\
\noindent \stepcounter{idaxiom} $AX\_\arabic{idaxiom}$ $\concept{condition} \sqsubseteq  \neg \concept{output\_set}$\vspace{0.1cm} \\
\noindent \stepcounter{idaxiom} $AX\_\arabic{idaxiom}$ $\concept{condition} \sqsubseteq  \neg \concept{input\_set}$\vspace{0.1cm} \\
\noindent \stepcounter{idaxiom} $AX\_\arabic{idaxiom}$ $\concept{event\_detail} \sqsubseteq  \neg \concept{assignment}$\vspace{0.1cm} \\
\noindent \stepcounter{idaxiom} $AX\_\arabic{idaxiom}$ $\concept{event\_detail} \sqsubseteq  \neg \concept{expression}$\vspace{0.1cm} \\
\noindent \stepcounter{idaxiom} $AX\_\arabic{idaxiom}$ $\concept{event\_detail} \sqsubseteq  \neg \concept{property}$\vspace{0.1cm} \\
\noindent \stepcounter{idaxiom} $AX\_\arabic{idaxiom}$ $\concept{event\_detail} \sqsubseteq  \neg \concept{transaction}$\vspace{0.1cm} \\
\noindent \stepcounter{idaxiom} $AX\_\arabic{idaxiom}$ $\concept{event\_detail} \sqsubseteq  \neg \concept{gate}$\vspace{0.1cm} \\
\noindent \stepcounter{idaxiom} $AX\_\arabic{idaxiom}$ $\concept{event\_detail} \sqsubseteq  \neg \concept{web\_service}$\vspace{0.1cm} \\
\noindent \stepcounter{idaxiom} $AX\_\arabic{idaxiom}$ $\concept{event\_detail} \sqsubseteq  \neg \concept{role}$\vspace{0.1cm} \\
\noindent \stepcounter{idaxiom} $AX\_\arabic{idaxiom}$ $\concept{event\_detail} \sqsubseteq  \neg \concept{entity}$\vspace{0.1cm} \\
\noindent \stepcounter{idaxiom} $AX\_\arabic{idaxiom}$ $\concept{event\_detail} \sqsubseteq  \neg \concept{participant}$\vspace{0.1cm} \\
\noindent \stepcounter{idaxiom} $AX\_\arabic{idaxiom}$ $\concept{event\_detail} \sqsubseteq  \neg \concept{category}$\vspace{0.1cm} \\
\noindent \stepcounter{idaxiom} $AX\_\arabic{idaxiom}$ $\concept{event\_detail} \sqsubseteq  \neg \concept{output\_set}$\vspace{0.1cm} \\
\noindent \stepcounter{idaxiom} $AX\_\arabic{idaxiom}$ $\concept{event\_detail} \sqsubseteq  \neg \concept{input\_set}$\vspace{0.1cm} \\
\noindent \stepcounter{idaxiom} $AX\_\arabic{idaxiom}$ $\concept{assignment} \sqsubseteq  \neg \concept{expression}$\vspace{0.1cm} \\
\noindent \stepcounter{idaxiom} $AX\_\arabic{idaxiom}$ $\concept{assignment} \sqsubseteq  \neg \concept{property}$\vspace{0.1cm} \\
\noindent \stepcounter{idaxiom} $AX\_\arabic{idaxiom}$ $\concept{assignment} \sqsubseteq  \neg \concept{transaction}$\vspace{0.1cm} \\
\noindent \stepcounter{idaxiom} $AX\_\arabic{idaxiom}$ $\concept{assignment} \sqsubseteq  \neg \concept{gate}$\vspace{0.1cm} \\
\noindent \stepcounter{idaxiom} $AX\_\arabic{idaxiom}$ $\concept{assignment} \sqsubseteq  \neg \concept{web\_service}$\vspace{0.1cm} \\
\noindent \stepcounter{idaxiom} $AX\_\arabic{idaxiom}$ $\concept{assignment} \sqsubseteq  \neg \concept{role}$\vspace{0.1cm} \\
\noindent \stepcounter{idaxiom} $AX\_\arabic{idaxiom}$ $\concept{assignment} \sqsubseteq  \neg \concept{entity}$\vspace{0.1cm} \\
\noindent \stepcounter{idaxiom} $AX\_\arabic{idaxiom}$ $\concept{assignment} \sqsubseteq  \neg \concept{participant}$\vspace{0.1cm} \\
\noindent \stepcounter{idaxiom} $AX\_\arabic{idaxiom}$ $\concept{assignment} \sqsubseteq  \neg \concept{category}$\vspace{0.1cm} \\
\noindent \stepcounter{idaxiom} $AX\_\arabic{idaxiom}$ $\concept{assignment} \sqsubseteq  \neg \concept{output\_set}$\vspace{0.1cm} \\
\noindent \stepcounter{idaxiom} $AX\_\arabic{idaxiom}$ $\concept{assignment} \sqsubseteq  \neg \concept{input\_set}$\vspace{0.1cm} \\
\noindent \stepcounter{idaxiom} $AX\_\arabic{idaxiom}$ $\concept{expression} \sqsubseteq  \neg \concept{property}$\vspace{0.1cm} \\
\noindent \stepcounter{idaxiom} $AX\_\arabic{idaxiom}$ $\concept{expression} \sqsubseteq  \neg \concept{transaction}$\vspace{0.1cm} \\
\noindent \stepcounter{idaxiom} $AX\_\arabic{idaxiom}$ $\concept{expression} \sqsubseteq  \neg \concept{gate}$\vspace{0.1cm} \\
\noindent \stepcounter{idaxiom} $AX\_\arabic{idaxiom}$ $\concept{expression} \sqsubseteq  \neg \concept{web\_service}$\vspace{0.1cm} \\
\noindent \stepcounter{idaxiom} $AX\_\arabic{idaxiom}$ $\concept{expression} \sqsubseteq  \neg \concept{role}$\vspace{0.1cm} \\
\noindent \stepcounter{idaxiom} $AX\_\arabic{idaxiom}$ $\concept{expression} \sqsubseteq  \neg \concept{entity}$\vspace{0.1cm} \\
\noindent \stepcounter{idaxiom} $AX\_\arabic{idaxiom}$ $\concept{expression} \sqsubseteq  \neg \concept{participant}$\vspace{0.1cm} \\
\noindent \stepcounter{idaxiom} $AX\_\arabic{idaxiom}$ $\concept{expression} \sqsubseteq  \neg \concept{category}$\vspace{0.1cm} \\
\noindent \stepcounter{idaxiom} $AX\_\arabic{idaxiom}$ $\concept{expression} \sqsubseteq  \neg \concept{output\_set}$\vspace{0.1cm} \\
\noindent \stepcounter{idaxiom} $AX\_\arabic{idaxiom}$ $\concept{expression} \sqsubseteq  \neg \concept{input\_set}$\vspace{0.1cm} \\
\noindent \stepcounter{idaxiom} $AX\_\arabic{idaxiom}$ $\concept{property} \sqsubseteq  \neg \concept{transaction}$\vspace{0.1cm} \\
\noindent \stepcounter{idaxiom} $AX\_\arabic{idaxiom}$ $\concept{property} \sqsubseteq  \neg \concept{gate}$\vspace{0.1cm} \\
\noindent \stepcounter{idaxiom} $AX\_\arabic{idaxiom}$ $\concept{property} \sqsubseteq  \neg \concept{web\_service}$\vspace{0.1cm} \\
\noindent \stepcounter{idaxiom} $AX\_\arabic{idaxiom}$ $\concept{property} \sqsubseteq  \neg \concept{role}$\vspace{0.1cm} \\
\noindent \stepcounter{idaxiom} $AX\_\arabic{idaxiom}$ $\concept{property} \sqsubseteq  \neg \concept{entity}$\vspace{0.1cm} \\
\noindent \stepcounter{idaxiom} $AX\_\arabic{idaxiom}$ $\concept{property} \sqsubseteq  \neg \concept{participant}$\vspace{0.1cm} \\
\noindent \stepcounter{idaxiom} $AX\_\arabic{idaxiom}$ $\concept{property} \sqsubseteq  \neg \concept{category}$\vspace{0.1cm} \\
\noindent \stepcounter{idaxiom} $AX\_\arabic{idaxiom}$ $\concept{property} \sqsubseteq  \neg \concept{output\_set}$\vspace{0.1cm} \\
\noindent \stepcounter{idaxiom} $AX\_\arabic{idaxiom}$ $\concept{property} \sqsubseteq  \neg \concept{input\_set}$\vspace{0.1cm} \\
\noindent \stepcounter{idaxiom} $AX\_\arabic{idaxiom}$ $\concept{transaction} \sqsubseteq  \neg \concept{gate}$\vspace{0.1cm} \\
\noindent \stepcounter{idaxiom} $AX\_\arabic{idaxiom}$ $\concept{transaction} \sqsubseteq  \neg \concept{web\_service}$\vspace{0.1cm} \\
\noindent \stepcounter{idaxiom} $AX\_\arabic{idaxiom}$ $\concept{transaction} \sqsubseteq  \neg \concept{role}$\vspace{0.1cm} \\
\noindent \stepcounter{idaxiom} $AX\_\arabic{idaxiom}$ $\concept{transaction} \sqsubseteq  \neg \concept{entity}$\vspace{0.1cm} \\
\noindent \stepcounter{idaxiom} $AX\_\arabic{idaxiom}$ $\concept{transaction} \sqsubseteq  \neg \concept{participant}$\vspace{0.1cm} \\
\noindent \stepcounter{idaxiom} $AX\_\arabic{idaxiom}$ $\concept{transaction} \sqsubseteq  \neg \concept{category}$\vspace{0.1cm} \\
\noindent \stepcounter{idaxiom} $AX\_\arabic{idaxiom}$ $\concept{transaction} \sqsubseteq  \neg \concept{output\_set}$\vspace{0.1cm} \\
\noindent \stepcounter{idaxiom} $AX\_\arabic{idaxiom}$ $\concept{transaction} \sqsubseteq  \neg \concept{input\_set}$\vspace{0.1cm} \\
\noindent \stepcounter{idaxiom} $AX\_\arabic{idaxiom}$ $\concept{gate} \sqsubseteq  \neg \concept{web\_service}$\vspace{0.1cm} \\
\noindent \stepcounter{idaxiom} $AX\_\arabic{idaxiom}$ $\concept{gate} \sqsubseteq  \neg \concept{role}$\vspace{0.1cm} \\
\noindent \stepcounter{idaxiom} $AX\_\arabic{idaxiom}$ $\concept{gate} \sqsubseteq  \neg \concept{entity}$\vspace{0.1cm} \\
\noindent \stepcounter{idaxiom} $AX\_\arabic{idaxiom}$ $\concept{gate} \sqsubseteq  \neg \concept{participant}$\vspace{0.1cm} \\
\noindent \stepcounter{idaxiom} $AX\_\arabic{idaxiom}$ $\concept{gate} \sqsubseteq  \neg \concept{category}$\vspace{0.1cm} \\
\noindent \stepcounter{idaxiom} $AX\_\arabic{idaxiom}$ $\concept{gate} \sqsubseteq  \neg \concept{output\_set}$\vspace{0.1cm} \\
\noindent \stepcounter{idaxiom} $AX\_\arabic{idaxiom}$ $\concept{gate} \sqsubseteq  \neg \concept{input\_set}$\vspace{0.1cm} \\
\noindent \stepcounter{idaxiom} $AX\_\arabic{idaxiom}$ $\concept{web\_service} \sqsubseteq  \neg \concept{role}$\vspace{0.1cm} \\
\noindent \stepcounter{idaxiom} $AX\_\arabic{idaxiom}$ $\concept{web\_service} \sqsubseteq  \neg \concept{entity}$\vspace{0.1cm} \\
\noindent \stepcounter{idaxiom} $AX\_\arabic{idaxiom}$ $\concept{web\_service} \sqsubseteq  \neg \concept{participant}$\vspace{0.1cm} \\
\noindent \stepcounter{idaxiom} $AX\_\arabic{idaxiom}$ $\concept{web\_service} \sqsubseteq  \neg \concept{category}$\vspace{0.1cm} \\
\noindent \stepcounter{idaxiom} $AX\_\arabic{idaxiom}$ $\concept{web\_service} \sqsubseteq  \neg \concept{output\_set}$\vspace{0.1cm} \\
\noindent \stepcounter{idaxiom} $AX\_\arabic{idaxiom}$ $\concept{web\_service} \sqsubseteq  \neg \concept{input\_set}$\vspace{0.1cm} \\
\noindent \stepcounter{idaxiom} $AX\_\arabic{idaxiom}$ $\concept{role} \sqsubseteq  \neg \concept{entity}$\vspace{0.1cm} \\
\noindent \stepcounter{idaxiom} $AX\_\arabic{idaxiom}$ $\concept{role} \sqsubseteq  \neg \concept{participant}$\vspace{0.1cm} \\
\noindent \stepcounter{idaxiom} $AX\_\arabic{idaxiom}$ $\concept{role} \sqsubseteq  \neg \concept{category}$\vspace{0.1cm} \\
\noindent \stepcounter{idaxiom} $AX\_\arabic{idaxiom}$ $\concept{role} \sqsubseteq  \neg \concept{output\_set}$\vspace{0.1cm} \\
\noindent \stepcounter{idaxiom} $AX\_\arabic{idaxiom}$ $\concept{role} \sqsubseteq  \neg \concept{input\_set}$\vspace{0.1cm} \\
\noindent \stepcounter{idaxiom} $AX\_\arabic{idaxiom}$ $\concept{entity} \sqsubseteq  \neg \concept{participant}$\vspace{0.1cm} \\
\noindent \stepcounter{idaxiom} $AX\_\arabic{idaxiom}$ $\concept{entity} \sqsubseteq  \neg \concept{category}$\vspace{0.1cm} \\
\noindent \stepcounter{idaxiom} $AX\_\arabic{idaxiom}$ $\concept{entity} \sqsubseteq  \neg \concept{output\_set}$\vspace{0.1cm} \\
\noindent \stepcounter{idaxiom} $AX\_\arabic{idaxiom}$ $\concept{entity} \sqsubseteq  \neg \concept{input\_set}$\vspace{0.1cm} \\
\noindent \stepcounter{idaxiom} $AX\_\arabic{idaxiom}$ $\concept{participant} \sqsubseteq  \neg \concept{category}$\vspace{0.1cm} \\
\noindent \stepcounter{idaxiom} $AX\_\arabic{idaxiom}$ $\concept{participant} \sqsubseteq  \neg \concept{output\_set}$\vspace{0.1cm} \\
\noindent \stepcounter{idaxiom} $AX\_\arabic{idaxiom}$ $\concept{participant} \sqsubseteq  \neg \concept{input\_set}$\vspace{0.1cm} \\
\noindent \stepcounter{idaxiom} $AX\_\arabic{idaxiom}$ $\concept{category} \sqsubseteq  \neg \concept{output\_set}$\vspace{0.1cm} \\
\noindent \stepcounter{idaxiom} $AX\_\arabic{idaxiom}$ $\concept{category} \sqsubseteq  \neg \concept{input\_set}$\vspace{0.1cm} \\
\noindent \stepcounter{idaxiom} $AX\_\arabic{idaxiom}$ $\concept{output\_set} \sqsubseteq  \neg \concept{input\_set}$\vspace{0.1cm} \\
\vspace{0.4cm} \hrule \vspace{0.2cm} \noindent \textbf{Class}: $\concept{artifact\_input}$\vspace{0.2cm} \hrule \vspace{0.2cm} 
\noindent \textbf{Label}: ArtifactInput\vspace{0.1cm} \\
\noindent \textbf{Description}: artifact\_input, which is used in the definition
	of attributes for all graphical elements.\vspace{0.1cm} \\
\noindent \stepcounter{idaxiom} $AX\_\arabic{idaxiom}$ $\concept{artifact\_input} \sqsubseteq  (=1) \property{has\_artifact\_input\_artifact\_ref}$\vspace{0.1cm} \\
\par \vspace{-0.1cm} \noindent \textbf{Property}: $\property{has\_artifact\_input\_artifact\_ref}$\vspace{0.1cm} \\
\noindent \textbf{Label}: ArtifactRef\vspace{0.1cm} \\
\noindent \textbf{Description}: This attribute identifies an 
	Artifact that will be used as an input to an activity. The identified 
	Artifact will be part of an InputSet for an activity.\vspace{0.1cm} \\
\noindent \stepcounter{idaxiom} $AX\_\arabic{idaxiom}$ $\property{has\_artifact\_input\_artifact\_ref} \mbox{ has range }\concept{artifact}$\vspace{0.1cm} \\
\noindent \stepcounter{idaxiom} $AX\_\arabic{idaxiom}$ $\property{has\_artifact\_input\_artifact\_ref} \mbox{ has domain }\concept{artifact\_input}$\vspace{0.1cm} \\
\noindent \stepcounter{idaxiom} $AX\_\arabic{idaxiom}$ $\concept{artifact\_input} \sqsubseteq  (=1) \property{has\_artifact\_input\_required\_for\_start}$\vspace{0.1cm} \\
\par \vspace{-0.1cm} \noindent \textbf{Property}: $\property{has\_artifact\_input\_required\_for\_start}$\vspace{0.1cm} \\
\noindent \textbf{Label}: RequiredForStart\vspace{0.1cm} \\
\noindent \textbf{Description}: The default value for 
	this attribute is True. This means that the Input is required for an 
	activity to start. If set to False, then the activity MAY start within 
	the input if it is available, but MAY accept the input (more than once) 
	after the activity has started. An InputSet may have a some of 
	ArtifactInputs that have this attribute set to True and some that are 
	set to False.\vspace{0.1cm} \\
\noindent \stepcounter{idaxiom} $AX\_\arabic{idaxiom}$ $\property{has\_artifact\_input\_required\_for\_start} \mbox{ has range }\datatype{xsd:boolean}$\vspace{0.1cm} \\
\noindent \stepcounter{idaxiom} $AX\_\arabic{idaxiom}$ $\property{has\_artifact\_input\_required\_for\_start} \mbox{ has domain }\concept{artifact\_input}$\vspace{0.1cm} \\
\vspace{0.4cm} \hrule \vspace{0.2cm} \noindent \textbf{Class}: $\concept{artifact\_output}$\vspace{0.2cm} \hrule \vspace{0.2cm} 
\noindent \textbf{Label}: ArtifactOutput\vspace{0.1cm} \\
\noindent \textbf{Description}: artifact\_output, which is used in the 
	definition of attributes for all graphical elements.\vspace{0.1cm} \\
\noindent \stepcounter{idaxiom} $AX\_\arabic{idaxiom}$ $\concept{artifact\_output} \sqsubseteq  (=1) \property{has\_artifact\_output\_artifact\_ref}$\vspace{0.1cm} \\
\par \vspace{-0.1cm} \noindent \textbf{Property}: $\property{has\_artifact\_output\_artifact\_ref}$\vspace{0.1cm} \\
\noindent \textbf{Label}: ArtifactRef\vspace{0.1cm} \\
\noindent \textbf{Description}: This attribute identifies an 
	Artifact that will be used as an output from an activity.
	The identified Artifact will be part of an OutputSet for an activity.\vspace{0.1cm} \\
\noindent \stepcounter{idaxiom} $AX\_\arabic{idaxiom}$ $\property{has\_artifact\_output\_artifact\_ref} \mbox{ has range }\concept{artifact}$\vspace{0.1cm} \\
\noindent \stepcounter{idaxiom} $AX\_\arabic{idaxiom}$ $\property{has\_artifact\_output\_artifact\_ref} \mbox{ has domain }\concept{artifact\_output}$\vspace{0.1cm} \\
\noindent \stepcounter{idaxiom} $AX\_\arabic{idaxiom}$ $\concept{artifact\_output} \sqsubseteq  (=1) \property{has\_artifact\_output\_produce\_at\_completion}$\vspace{0.1cm} \\
\par \vspace{-0.1cm} \noindent \textbf{Property}: $\property{has\_artifact\_output\_produce\_at\_completion}$\vspace{0.1cm} \\
\noindent \textbf{Label}: ProduceAtCompletion\vspace{0.1cm} \\
\noindent \textbf{Description}: The default value 
	for this attribute is True. This means that the Output will be produced 
	when an activity has been completed. If set to False, then the activity
	MAY produce the output (more than once) before it has completed. An 
	OutputSet may have a some of ArtifactOutputs that have this attribute 
	set to True and some that are set to False.\vspace{0.1cm} \\
\noindent \stepcounter{idaxiom} $AX\_\arabic{idaxiom}$ $\property{has\_artifact\_output\_produce\_at\_completion} \mbox{ has range }\datatype{xsd:boolean}$\vspace{0.1cm} \\
\noindent \stepcounter{idaxiom} $AX\_\arabic{idaxiom}$ $\property{has\_artifact\_output\_produce\_at\_completion} \mbox{ has domain }\concept{artifact\_output}$\vspace{0.1cm} \\
\vspace{0.4cm} \hrule \vspace{0.2cm} \noindent \textbf{Class}: $\concept{assignment}$\vspace{0.2cm} \hrule \vspace{0.2cm} 
\noindent \textbf{Label}: Assignment\vspace{0.1cm} \\
\noindent \textbf{Description}: Assignment, which is used in the definition of 
	attributes for Process, Activities, Events, Gateways, and Gates, and 
	which extends the set of common BPMN Element attributes\vspace{0.1cm} \\
\noindent \stepcounter{idaxiom} $AX\_\arabic{idaxiom}$ $\concept{assignment} \sqsubseteq  (=1) \property{has\_assignment\_to}$\vspace{0.1cm} \\
\par \vspace{-0.1cm} \noindent \textbf{Property}: $\property{has\_assignment\_to}$\vspace{0.1cm} \\
\noindent \textbf{Label}: To\vspace{0.1cm} \\
\noindent \textbf{Description}: The target for the Assignment MUST be a 
	Property of the Process or the activity itself.\vspace{0.1cm} \\
\noindent \stepcounter{idaxiom} $AX\_\arabic{idaxiom}$ $\property{has\_assignment\_to} \mbox{ has domain }\concept{assignment}$\vspace{0.1cm} \\
\noindent \stepcounter{idaxiom} $AX\_\arabic{idaxiom}$ $\property{has\_assignment\_to} \mbox{ has range }\concept{property}$\vspace{0.1cm} \\
\noindent \stepcounter{idaxiom} $AX\_\arabic{idaxiom}$ $\concept{assignment} \sqsubseteq  (=1) \property{has\_assignment\_from}$\vspace{0.1cm} \\
\par \vspace{-0.1cm} \noindent \textbf{Property}: $\property{has\_assignment\_from}$\vspace{0.1cm} \\
\noindent \textbf{Label}: From\vspace{0.1cm} \\
\noindent \textbf{Description}: The Expression MUST be made up of a 
	combination of Values, Properties, and Attributes, which are separated 
	by operators such as add or multiply. The expression language is defined
	in the ExpressionLanguage attribute of the Business Process Diagram - 
	see "Business Process Diagram Attributes on page 241."\vspace{0.1cm} \\
\noindent \stepcounter{idaxiom} $AX\_\arabic{idaxiom}$ $\property{has\_assignment\_from} \mbox{ has domain }\concept{assignment}$\vspace{0.1cm} \\
\noindent \stepcounter{idaxiom} $AX\_\arabic{idaxiom}$ $\property{has\_assignment\_from} \mbox{ has range }\concept{expression}$\vspace{0.1cm} \\
\noindent \stepcounter{idaxiom} $AX\_\arabic{idaxiom}$ $\concept{assignment} \sqsubseteq (\geq1) \property{has\_assignment\_assign\_time}$\vspace{0.1cm} \\
\par \vspace{-0.1cm} \noindent \textbf{Property}: $\property{has\_assignment\_assign\_time}$\vspace{0.1cm} \\
\noindent \textbf{Label}: AssignTime\vspace{0.1cm} \\
\noindent \textbf{Description}: An Assignment MAY have a AssignTime
	setting. If the Object is an activity (Task, Sub-Process, or Process), 
	then the Assignment MUST have an AssignTime. A value of Start means that
	the assignment SHALL occur at the start of the activity. This can be 
	used to assign the higher-level (global) Properties of the Process to 
	the (local) Properties of the activity as an input to the activity. A 
	value of End means that the assignment SHALL occur at the end of the 
	activity. This can be used to assign the (local) Properties of the 
	activity to the higher-level (global) Properties of the Process as an 
	output to the activity.\vspace{0.1cm} \\
\noindent \stepcounter{idaxiom} $AX\_\arabic{idaxiom}$ $\property{has\_assignment\_assign\_time} \mbox{ has range }\datatype{xsd:string}\{"\datainstance{Start}","\datainstance{End}"\}$\vspace{0.1cm} \\
\noindent \stepcounter{idaxiom} $AX\_\arabic{idaxiom}$ $\property{has\_assignment\_assign\_time} \mbox{ has domain }\concept{assignment}$\vspace{0.1cm} \\
\vspace{0.4cm} \hrule \vspace{0.2cm} \noindent \textbf{Class}: $\concept{category}$\vspace{0.2cm} \hrule \vspace{0.2cm} 
\noindent \textbf{Label}: Category\vspace{0.1cm} \\
\noindent \textbf{Description}: Category, which is used in the definition of 
	attributes for all BPMN elements, and which extends the set of common 
	BPMN Element attributes (see Table B.2). Since a Category is also a 
	BPMN element, a Category can have Categories to create a hierarchical 
	structure of Categories.\vspace{0.1cm} \\
\noindent \stepcounter{idaxiom} $AX\_\arabic{idaxiom}$ $\concept{category} \sqsubseteq  (=1) \property{has\_category\_name}$\vspace{0.1cm} \\
\par \vspace{-0.1cm} \noindent \textbf{Property}: $\property{has\_category\_name}$\vspace{0.1cm} \\
\noindent \textbf{Label}: Name\vspace{0.1cm} \\
\noindent \textbf{Description}: Name is an attribute that is text 
	description of the Category and is used to visually distinguish the 
	category.\vspace{0.1cm} \\
\noindent \stepcounter{idaxiom} $AX\_\arabic{idaxiom}$ $\property{has\_category\_name} \mbox{ has domain }\concept{category}$\vspace{0.1cm} \\
\noindent \stepcounter{idaxiom} $AX\_\arabic{idaxiom}$ $\property{has\_category\_name} \mbox{ has range }\datatype{xsd:string}$\vspace{0.1cm} \\
\vspace{0.4cm} \hrule \vspace{0.2cm} \noindent \textbf{Class}: $\concept{condition}$\vspace{0.2cm} \hrule \vspace{0.2cm} 
\noindent \textbf{Label}: Condition\vspace{0.1cm} \\
\noindent \textbf{Description}: Condition, which is used in the definition of 
	attributes for Start Event and Intermediate Event, and which extends the
	set of common BPMN Element attributes (see Table B.2).\vspace{0.1cm} \\
\noindent \stepcounter{idaxiom} $AX\_\arabic{idaxiom}$ $\concept{condition} \sqsubseteq  (=1) \property{has\_condition\_name} \sqcup  (=1) \property{has\_condition\_condition\_expression}$\vspace{0.1cm} \\
\par \vspace{-0.1cm} \noindent \textbf{Property}: $\property{has\_condition\_name}$\vspace{0.1cm} \\
\noindent \textbf{Label}: Name\vspace{0.1cm} \\
\noindent \textbf{Description}: Name is an optional attribute that is text 
	description of the Condition. If a Name is not entered, then a 
	ConditionExpression MUST be entered.\vspace{0.1cm} \\
\noindent \stepcounter{idaxiom} $AX\_\arabic{idaxiom}$ $\property{has\_condition\_name} \mbox{ has domain }\concept{condition}$\vspace{0.1cm} \\
\noindent \stepcounter{idaxiom} $AX\_\arabic{idaxiom}$ $\property{has\_condition\_name} \mbox{ has range }\datatype{xsd:string}$\vspace{0.1cm} \\
\par \vspace{-0.1cm} \noindent \textbf{Property}: $\property{has\_condition\_condition\_expression}$\vspace{0.1cm} \\
\noindent \textbf{Label}: ConditionExpression\vspace{0.1cm} \\
\noindent \textbf{Description}: A ConditionExpression MAY 
	be entered. In some cases the Condition itself will be stored and 
	maintained in a separate application (e.g., a Rules Engine). If a 
	ConditionExpression is not entered, then a Name MUST be entered. The 
	attributes of an Expression can be found in "Expression on page 273."\vspace{0.1cm} \\
\noindent \stepcounter{idaxiom} $AX\_\arabic{idaxiom}$ $\property{has\_condition\_condition\_expression} \mbox{ has domain }\concept{condition}$\vspace{0.1cm} \\
\noindent \stepcounter{idaxiom} $AX\_\arabic{idaxiom}$ $\property{has\_condition\_condition\_expression} \mbox{ has range }\concept{expression}$\vspace{0.1cm} \\
\vspace{0.4cm} \hrule \vspace{0.2cm} \noindent \textbf{Class}: $\concept{entity}$\vspace{0.2cm} \hrule \vspace{0.2cm} 
\noindent \textbf{Label}: Entity\vspace{0.1cm} \\
\noindent \textbf{Description}: Entity, which is used in the definition of attributes 
	for a Participant, and which extends the set of common BPMN Element 
	attributes (see Table B.2).\vspace{0.1cm} \\
\noindent \stepcounter{idaxiom} $AX\_\arabic{idaxiom}$ $\concept{entity} \sqsubseteq  (=1) \property{has\_entity\_name}$\vspace{0.1cm} \\
\par \vspace{-0.1cm} \noindent \textbf{Property}: $\property{has\_entity\_name}$\vspace{0.1cm} \\
\noindent \textbf{Label}: Name\vspace{0.1cm} \\
\noindent \textbf{Description}: Name is an attribute that is text description 
	of the Entity.\vspace{0.1cm} \\
\noindent \stepcounter{idaxiom} $AX\_\arabic{idaxiom}$ $\property{has\_entity\_name} \mbox{ has domain }\concept{entity}$\vspace{0.1cm} \\
\noindent \stepcounter{idaxiom} $AX\_\arabic{idaxiom}$ $\property{has\_entity\_name} \mbox{ has range }\datatype{xsd:string}$\vspace{0.1cm} \\
\vspace{0.4cm} \hrule \vspace{0.2cm} \noindent \textbf{Class}: $\concept{event\_detail}$\vspace{0.2cm} \hrule \vspace{0.2cm} 
\noindent \textbf{Label}: Event Detail\vspace{0.1cm} \\
\noindent \textbf{Description}: present the attributes common to all Event 
	Details and the specific attributes for the Event Details that have 
	additional attributes. Note that the Cancel and Terminate Event Details
	do not have additional attributes\vspace{0.1cm} \\
\noindent \stepcounter{idaxiom} $AX\_\arabic{idaxiom}$ $\concept{event\_detail\_types} \equiv \{\instance{cancel\_event\_detail\_type}, \instance{compensation\_event\_detail\_type}, \instance{link\_event\_detail\_type}, \\ \instance{error\_event\_detail\_type}, \instance{conditional\_event\_detail\_type}, \instance{message\_event\_detail\_type}, \instance{terminate\_event\_detail\_type}, \\ \instance{timer\_event\_detail\_type}, \instance{signal\_event\_detail\_type}\}$\vspace{0.1cm} \\
\noindent \stepcounter{idaxiom} $AX\_\arabic{idaxiom}$ $\concept{event\_detail} \sqsubseteq  (=1) \property{has\_event\_detail\_type}$\vspace{0.1cm} \\
\par \vspace{-0.1cm} \noindent \textbf{Property}: $\property{has\_event\_detail\_type}$\vspace{0.1cm} \\
\noindent \textbf{Label}: Event Detail Type\vspace{0.1cm} \\
\noindent \textbf{Description}: The EventDetailType attribute defines 
	the type of trigger expected for an Event. The set of types includes 
	Message, Timer, Error, Conditional, Link, Signal, Compensate,
	Cancel, and Terminate. The EventTypes (Start, Intermediate, and End) 
	will each have a subset of the EventDetailTypes that can be used.
	The EventDetailType list MAY be extended to include new types. These new
	types MAY have a new modeler- or tool-defined Marker to fit within the 
	boundaries of the Event.\vspace{0.1cm} \\
\noindent \stepcounter{idaxiom} $AX\_\arabic{idaxiom}$ $\property{has\_event\_detail\_type} \mbox{ has domain }\concept{event\_detail}$\vspace{0.1cm} \\
\noindent \stepcounter{idaxiom} $AX\_\arabic{idaxiom}$ $\property{has\_event\_detail\_type} \mbox{ has range }\concept{event\_detail\_types}$\vspace{0.1cm} \\
\par \vspace{-0.1cm} \noindent \textbf{Instance}: $\instance{cancel\_event\_detail\_type}$\vspace{0.1cm} \\
\noindent \textbf{Label}: cancel\vspace{0.1cm} \\
\par \vspace{-0.1cm} \noindent \textbf{Instance}: $\instance{compensation\_event\_detail\_type}$\vspace{0.1cm} \\
\noindent \textbf{Label}: compensation\vspace{0.1cm} \\
\par \vspace{-0.1cm} \noindent \textbf{Instance}: $\instance{link\_event\_detail\_type}$\vspace{0.1cm} \\
\noindent \textbf{Label}: link\vspace{0.1cm} \\
\par \vspace{-0.1cm} \noindent \textbf{Instance}: $\instance{error\_event\_detail\_type}$\vspace{0.1cm} \\
\noindent \textbf{Label}: error\vspace{0.1cm} \\
\par \vspace{-0.1cm} \noindent \textbf{Instance}: $\instance{conditional\_event\_detail\_type}$\vspace{0.1cm} \\
\noindent \textbf{Label}: conditional\vspace{0.1cm} \\
\par \vspace{-0.1cm} \noindent \textbf{Instance}: $\instance{message\_event\_detail\_type}$\vspace{0.1cm} \\
\noindent \textbf{Label}: message\vspace{0.1cm} \\
\par \vspace{-0.1cm} \noindent \textbf{Instance}: $\instance{terminate\_event\_detail\_type}$\vspace{0.1cm} \\
\noindent \textbf{Label}: terminate\vspace{0.1cm} \\
\par \vspace{-0.1cm} \noindent \textbf{Instance}: $\instance{timer\_event\_detail\_type}$\vspace{0.1cm} \\
\noindent \textbf{Label}: timer\vspace{0.1cm} \\
\par \vspace{-0.1cm} \noindent \textbf{Instance}: $\instance{signal\_event\_detail\_type}$\vspace{0.1cm} \\
\noindent \textbf{Label}: signal\vspace{0.1cm} \\
\noindent \stepcounter{idaxiom} $AX\_\arabic{idaxiom}$ $( \neg \{\instance{cancel\_event\_detail\_type}\})(\instance{compensation\_event\_detail\_type})$\vspace{0.1cm} \\
\noindent \stepcounter{idaxiom} $AX\_\arabic{idaxiom}$ $( \neg \{\instance{cancel\_event\_detail\_type}\})(\instance{link\_event\_detail\_type})$\vspace{0.1cm} \\
\noindent \stepcounter{idaxiom} $AX\_\arabic{idaxiom}$ $( \neg \{\instance{cancel\_event\_detail\_type}\})(\instance{error\_event\_detail\_type})$\vspace{0.1cm} \\
\noindent \stepcounter{idaxiom} $AX\_\arabic{idaxiom}$ $( \neg \{\instance{cancel\_event\_detail\_type}\})(\instance{conditional\_event\_detail\_type})$\vspace{0.1cm} \\
\noindent \stepcounter{idaxiom} $AX\_\arabic{idaxiom}$ $( \neg \{\instance{cancel\_event\_detail\_type}\})(\instance{message\_event\_detail\_type})$\vspace{0.1cm} \\
\noindent \stepcounter{idaxiom} $AX\_\arabic{idaxiom}$ $( \neg \{\instance{cancel\_event\_detail\_type}\})(\instance{terminate\_event\_detail\_type})$\vspace{0.1cm} \\
\noindent \stepcounter{idaxiom} $AX\_\arabic{idaxiom}$ $( \neg \{\instance{cancel\_event\_detail\_type}\})(\instance{timer\_event\_detail\_type})$\vspace{0.1cm} \\
\noindent \stepcounter{idaxiom} $AX\_\arabic{idaxiom}$ $( \neg \{\instance{cancel\_event\_detail\_type}\})(\instance{signal\_event\_detail\_type})$\vspace{0.1cm} \\
\noindent \stepcounter{idaxiom} $AX\_\arabic{idaxiom}$ $( \neg \{\instance{compensation\_event\_detail\_type}\})(\instance{link\_event\_detail\_type})$\vspace{0.1cm} \\
\noindent \stepcounter{idaxiom} $AX\_\arabic{idaxiom}$ $( \neg \{\instance{compensation\_event\_detail\_type}\})(\instance{error\_event\_detail\_type})$\vspace{0.1cm} \\
\noindent \stepcounter{idaxiom} $AX\_\arabic{idaxiom}$ $( \neg \{\instance{compensation\_event\_detail\_type}\})(\instance{conditional\_event\_detail\_type})$\vspace{0.1cm} \\
\noindent \stepcounter{idaxiom} $AX\_\arabic{idaxiom}$ $( \neg \{\instance{compensation\_event\_detail\_type}\})(\instance{message\_event\_detail\_type})$\vspace{0.1cm} \\
\noindent \stepcounter{idaxiom} $AX\_\arabic{idaxiom}$ $( \neg \{\instance{compensation\_event\_detail\_type}\})(\instance{terminate\_event\_detail\_type})$\vspace{0.1cm} \\
\noindent \stepcounter{idaxiom} $AX\_\arabic{idaxiom}$ $( \neg \{\instance{compensation\_event\_detail\_type}\})(\instance{timer\_event\_detail\_type})$\vspace{0.1cm} \\
\noindent \stepcounter{idaxiom} $AX\_\arabic{idaxiom}$ $( \neg \{\instance{compensation\_event\_detail\_type}\})(\instance{signal\_event\_detail\_type})$\vspace{0.1cm} \\
\noindent \stepcounter{idaxiom} $AX\_\arabic{idaxiom}$ $( \neg \{\instance{link\_event\_detail\_type}\})(\instance{error\_event\_detail\_type})$\vspace{0.1cm} \\
\noindent \stepcounter{idaxiom} $AX\_\arabic{idaxiom}$ $( \neg \{\instance{link\_event\_detail\_type}\})(\instance{conditional\_event\_detail\_type})$\vspace{0.1cm} \\
\noindent \stepcounter{idaxiom} $AX\_\arabic{idaxiom}$ $( \neg \{\instance{link\_event\_detail\_type}\})(\instance{message\_event\_detail\_type})$\vspace{0.1cm} \\
\noindent \stepcounter{idaxiom} $AX\_\arabic{idaxiom}$ $( \neg \{\instance{link\_event\_detail\_type}\})(\instance{terminate\_event\_detail\_type})$\vspace{0.1cm} \\
\noindent \stepcounter{idaxiom} $AX\_\arabic{idaxiom}$ $( \neg \{\instance{link\_event\_detail\_type}\})(\instance{timer\_event\_detail\_type})$\vspace{0.1cm} \\
\noindent \stepcounter{idaxiom} $AX\_\arabic{idaxiom}$ $( \neg \{\instance{link\_event\_detail\_type}\})(\instance{signal\_event\_detail\_type})$\vspace{0.1cm} \\
\noindent \stepcounter{idaxiom} $AX\_\arabic{idaxiom}$ $( \neg \{\instance{error\_event\_detail\_type}\})(\instance{conditional\_event\_detail\_type})$\vspace{0.1cm} \\
\noindent \stepcounter{idaxiom} $AX\_\arabic{idaxiom}$ $( \neg \{\instance{error\_event\_detail\_type}\})(\instance{message\_event\_detail\_type})$\vspace{0.1cm} \\
\noindent \stepcounter{idaxiom} $AX\_\arabic{idaxiom}$ $( \neg \{\instance{error\_event\_detail\_type}\})(\instance{terminate\_event\_detail\_type})$\vspace{0.1cm} \\
\noindent \stepcounter{idaxiom} $AX\_\arabic{idaxiom}$ $( \neg \{\instance{error\_event\_detail\_type}\})(\instance{timer\_event\_detail\_type})$\vspace{0.1cm} \\
\noindent \stepcounter{idaxiom} $AX\_\arabic{idaxiom}$ $( \neg \{\instance{error\_event\_detail\_type}\})(\instance{signal\_event\_detail\_type})$\vspace{0.1cm} \\
\noindent \stepcounter{idaxiom} $AX\_\arabic{idaxiom}$ $( \neg \{\instance{conditional\_event\_detail\_type}\})(\instance{message\_event\_detail\_type})$\vspace{0.1cm} \\
\noindent \stepcounter{idaxiom} $AX\_\arabic{idaxiom}$ $( \neg \{\instance{conditional\_event\_detail\_type}\})(\instance{terminate\_event\_detail\_type})$\vspace{0.1cm} \\
\noindent \stepcounter{idaxiom} $AX\_\arabic{idaxiom}$ $( \neg \{\instance{conditional\_event\_detail\_type}\})(\instance{timer\_event\_detail\_type})$\vspace{0.1cm} \\
\noindent \stepcounter{idaxiom} $AX\_\arabic{idaxiom}$ $( \neg \{\instance{conditional\_event\_detail\_type}\})(\instance{signal\_event\_detail\_type})$\vspace{0.1cm} \\
\noindent \stepcounter{idaxiom} $AX\_\arabic{idaxiom}$ $( \neg \{\instance{message\_event\_detail\_type}\})(\instance{terminate\_event\_detail\_type})$\vspace{0.1cm} \\
\noindent \stepcounter{idaxiom} $AX\_\arabic{idaxiom}$ $( \neg \{\instance{message\_event\_detail\_type}\})(\instance{timer\_event\_detail\_type})$\vspace{0.1cm} \\
\noindent \stepcounter{idaxiom} $AX\_\arabic{idaxiom}$ $( \neg \{\instance{message\_event\_detail\_type}\})(\instance{signal\_event\_detail\_type})$\vspace{0.1cm} \\
\noindent \stepcounter{idaxiom} $AX\_\arabic{idaxiom}$ $( \neg \{\instance{terminate\_event\_detail\_type}\})(\instance{timer\_event\_detail\_type})$\vspace{0.1cm} \\
\noindent \stepcounter{idaxiom} $AX\_\arabic{idaxiom}$ $( \neg \{\instance{terminate\_event\_detail\_type}\})(\instance{signal\_event\_detail\_type})$\vspace{0.1cm} \\
\noindent \stepcounter{idaxiom} $AX\_\arabic{idaxiom}$ $( \neg \{\instance{timer\_event\_detail\_type}\})(\instance{signal\_event\_detail\_type})$\vspace{0.1cm} \\
\noindent \stepcounter{idaxiom} $AX\_\arabic{idaxiom}$ $\concept{cancel\_event\_detail} \equiv \concept{event\_detail} \sqcap \exists\property{has\_event\_detail\_type}.\{\instance{cancel\_event\_detail\_type}\}$\vspace{0.1cm} \\
\vspace{0.4cm} \hrule \vspace{0.2cm} \noindent \textbf{Class}: $\concept{cancel\_event\_detail}$\vspace{0.2cm} \hrule \vspace{0.2cm} 
\noindent \textbf{Label}: Cancel Event Detail\vspace{0.1cm} \\
\noindent \stepcounter{idaxiom} $AX\_\arabic{idaxiom}$ $\concept{conditional\_event\_detail} \equiv \concept{event\_detail} \sqcap \exists\property{has\_event\_detail\_type}.\{\instance{conditional\_event\_detail\_type}\}$\vspace{0.1cm} \\
\vspace{0.4cm} \hrule \vspace{0.2cm} \noindent \textbf{Class}: $\concept{conditional\_event\_detail}$\vspace{0.2cm} \hrule \vspace{0.2cm} 
\noindent \textbf{Label}: Conditional Event Detail\vspace{0.1cm} \\
\noindent \stepcounter{idaxiom} $AX\_\arabic{idaxiom}$ $\concept{conditional\_event\_detail} \sqsubseteq  (=1) \property{has\_conditional\_event\_condition\_ref}$\vspace{0.1cm} \\
\par \vspace{-0.1cm} \noindent \textbf{Property}: $\property{has\_conditional\_event\_condition\_ref}$\vspace{0.1cm} \\
\noindent \textbf{Label}: ConditionRef\vspace{0.1cm} \\
\noindent \textbf{Description}: If the Trigger is 
	Conditional, then a Condition MUST be entered. The attributes of
	a Condition can be found in Section B.11.5, "Condition," on page 269.\vspace{0.1cm} \\
\noindent \stepcounter{idaxiom} $AX\_\arabic{idaxiom}$ $\property{has\_conditional\_event\_condition\_ref} \mbox{ has domain }\concept{conditional\_event\_detail}$\vspace{0.1cm} \\
\noindent \stepcounter{idaxiom} $AX\_\arabic{idaxiom}$ $\property{has\_conditional\_event\_condition\_ref} \mbox{ has range }\concept{condition}$\vspace{0.1cm} \\
\noindent \stepcounter{idaxiom} $AX\_\arabic{idaxiom}$ $\concept{compensation\_event\_detail} \equiv \concept{event\_detail} \sqcap \exists\property{has\_event\_detail\_type}.\{\instance{compensation\_event\_detail\_type}\}$\vspace{0.1cm} \\
\vspace{0.4cm} \hrule \vspace{0.2cm} \noindent \textbf{Class}: $\concept{compensation\_event\_detail}$\vspace{0.2cm} \hrule \vspace{0.2cm} 
\noindent \textbf{Label}: Compensation Event Detail\vspace{0.1cm} \\
\noindent \stepcounter{idaxiom} $AX\_\arabic{idaxiom}$ $\concept{compensation\_event\_detail} \sqsubseteq (\geq1) \property{has\_activity\_ref}$\vspace{0.1cm} \\
\par \vspace{-0.1cm} \noindent \textbf{Property}: $\property{has\_activity\_ref}$\vspace{0.1cm} \\
\noindent \textbf{Label}: ActivityRef\vspace{0.1cm} \\
\noindent \textbf{Description}: For an End Event:
	If the Result is a Compensation, then the Activity that needs to be 
	compensated MAY be supplied. If an Activity is not supplied, then the 
	Event broadcast to all completed activities in the Process Instance.
	For an Intermediate Event within Normal Flow:
	If the Trigger is a Compensation, then the Activity that needs to be 
	compensated MAY be supplied. If an Activity is not supplied, then the 
	Event broadcast to all completed activities in the Process Instance. 
	This "throws" the compensation.
	For an Intermediate Event attached to the boundary of an Activity:
	This Event "catches" the compensation. No further information is 
	required. The Activity the Event is attached to will provide the Id 
	necessary to match the compensation event with the event that "threw" 
	the compensation or the compensation will be a broadcast.\vspace{0.1cm} \\
\noindent \stepcounter{idaxiom} $AX\_\arabic{idaxiom}$ $\property{has\_activity\_ref} \mbox{ has domain }\concept{compensation\_event\_detail}$\vspace{0.1cm} \\
\noindent \stepcounter{idaxiom} $AX\_\arabic{idaxiom}$ $\property{has\_activity\_ref} \mbox{ has range }\concept{activity}$\vspace{0.1cm} \\
\noindent \stepcounter{idaxiom} $AX\_\arabic{idaxiom}$ $\concept{error\_event\_detail} \equiv \concept{event\_detail} \sqcap \exists\property{has\_event\_detail\_type}.\{\instance{error\_event\_detail\_type}\}$\vspace{0.1cm} \\
\vspace{0.4cm} \hrule \vspace{0.2cm} \noindent \textbf{Class}: $\concept{error\_event\_detail}$\vspace{0.2cm} \hrule \vspace{0.2cm} 
\noindent \textbf{Label}: Error Event Detail\vspace{0.1cm} \\
\noindent \stepcounter{idaxiom} $AX\_\arabic{idaxiom}$ $\concept{error\_event\_detail} \sqsubseteq (\geq1) \property{has\_error\_detail\_error\_code}$\vspace{0.1cm} \\
\par \vspace{-0.1cm} \noindent \textbf{Property}: $\property{has\_error\_detail\_error\_code}$\vspace{0.1cm} \\
\noindent \textbf{Label}: ErrorCode\vspace{0.1cm} \\
\noindent \textbf{Description}: For an End Event:
	If the Result is an Error, then the ErrorCode MUST be supplied.This 
	"throws" the error.
	For an Intermediate Event within Normal Flow:
	If the Trigger is an Error, then the ErrorCode MUST be entered. This 
	"throws" the error.
	For an Intermediate Event attached to the boundary of an Activity:
	If the Trigger is an Error, then the ErrorCode MAY be entered. This 
	Event "catches" the error. If there is no ErrorCode, then any error 
	SHALL trigger the Event. If there is an ErrorCode, then only an error 
	that matches the ErrorCode SHALL trigger the Event.\vspace{0.1cm} \\
\noindent \stepcounter{idaxiom} $AX\_\arabic{idaxiom}$ $\property{has\_error\_detail\_error\_code} \mbox{ has domain }\concept{error\_event\_detail}$\vspace{0.1cm} \\
\noindent \stepcounter{idaxiom} $AX\_\arabic{idaxiom}$ $\property{has\_error\_detail\_error\_code} \mbox{ has range }\datatype{xsd:string}$\vspace{0.1cm} \\
\noindent \stepcounter{idaxiom} $AX\_\arabic{idaxiom}$ $\concept{link\_event\_detail} \equiv \concept{event\_detail} \sqcap \exists\property{has\_event\_detail\_type}.\{\instance{link\_event\_detail\_type}\}$\vspace{0.1cm} \\
\vspace{0.4cm} \hrule \vspace{0.2cm} \noindent \textbf{Class}: $\concept{link\_event\_detail}$\vspace{0.2cm} \hrule \vspace{0.2cm} 
\noindent \textbf{Label}: Link Event Detail\vspace{0.1cm} \\
\noindent \stepcounter{idaxiom} $AX\_\arabic{idaxiom}$ $\concept{link\_event\_detail} \sqsubseteq  (=1) \property{has\_link\_event\_name}$\vspace{0.1cm} \\
\par \vspace{-0.1cm} \noindent \textbf{Property}: $\property{has\_link\_event\_name}$\vspace{0.1cm} \\
\noindent \textbf{Label}: Name\vspace{0.1cm} \\
\noindent \textbf{Description}: If the Trigger is a Link, then the Name 
	MUST be entered.\vspace{0.1cm} \\
\noindent \stepcounter{idaxiom} $AX\_\arabic{idaxiom}$ $\property{has\_link\_event\_name} \mbox{ has domain }\concept{link\_event\_detail}$\vspace{0.1cm} \\
\noindent \stepcounter{idaxiom} $AX\_\arabic{idaxiom}$ $\property{has\_link\_event\_name} \mbox{ has range }\datatype{xsd:string}$\vspace{0.1cm} \\
\noindent \stepcounter{idaxiom} $AX\_\arabic{idaxiom}$ $\concept{message\_event\_detail} \equiv \concept{event\_detail} \sqcap \exists\property{has\_event\_detail\_type}.\{\instance{message\_event\_detail\_type}\}$\vspace{0.1cm} \\
\vspace{0.4cm} \hrule \vspace{0.2cm} \noindent \textbf{Class}: $\concept{message\_event\_detail}$\vspace{0.2cm} \hrule \vspace{0.2cm} 
\noindent \textbf{Label}: Message Event Detail\vspace{0.1cm} \\
\noindent \stepcounter{idaxiom} $AX\_\arabic{idaxiom}$ $\concept{message\_event\_detail} \sqsubseteq  (=1) \property{has\_message\_event\_message\_ref}$\vspace{0.1cm} \\
\par \vspace{-0.1cm} \noindent \textbf{Property}: $\property{has\_message\_event\_message\_ref}$\vspace{0.1cm} \\
\noindent \textbf{Label}: MessageRef\vspace{0.1cm} \\
\noindent \textbf{Description}: If the EventDetailType is a 
	MessageRef, then the a Message MUST be supplied. The attributes of a 
	Message can be found in Section B.11.11, "Message," on page 275.\vspace{0.1cm} \\
\noindent \stepcounter{idaxiom} $AX\_\arabic{idaxiom}$ $\property{has\_message\_event\_message\_ref} \mbox{ has domain }\concept{message\_event\_detail}$\vspace{0.1cm} \\
\noindent \stepcounter{idaxiom} $AX\_\arabic{idaxiom}$ $\property{has\_message\_event\_message\_ref} \mbox{ has range }\concept{message}$\vspace{0.1cm} \\
\noindent \stepcounter{idaxiom} $AX\_\arabic{idaxiom}$ $\concept{message\_event\_detail} \sqsubseteq  (=1) \property{has\_message\_event\_implementation}$\vspace{0.1cm} \\
\par \vspace{-0.1cm} \noindent \textbf{Property}: $\property{has\_message\_event\_implementation}$\vspace{0.1cm} \\
\noindent \textbf{Label}: Implementation\vspace{0.1cm} \\
\noindent \textbf{Description}: This attribute specifies the
	technology that will be used to send or receive the message. A Web 
	service is the default technology.\vspace{0.1cm} \\
\noindent \stepcounter{idaxiom} $AX\_\arabic{idaxiom}$ $\property{has\_message\_event\_implementation} \mbox{ has domain }\concept{message\_event\_detail}$\vspace{0.1cm} \\
\noindent \stepcounter{idaxiom} $AX\_\arabic{idaxiom}$ $\property{has\_message\_event\_implementation} \mbox{ has range }\datatype{xsd:string}\{"\datainstance{Web\_Service}","\datainstance{Other}","\datainstance{Unspecified}"\}$\vspace{0.1cm} \\
\noindent \stepcounter{idaxiom} $AX\_\arabic{idaxiom}$ $\concept{signal\_event\_detail} \equiv \concept{event\_detail} \sqcap \exists\property{has\_event\_detail\_type}.\{\instance{signal\_event\_detail\_type}\}$\vspace{0.1cm} \\
\vspace{0.4cm} \hrule \vspace{0.2cm} \noindent \textbf{Class}: $\concept{signal\_event\_detail}$\vspace{0.2cm} \hrule \vspace{0.2cm} 
\noindent \textbf{Label}: Signal Event Detail\vspace{0.1cm} \\
\noindent \stepcounter{idaxiom} $AX\_\arabic{idaxiom}$ $\concept{signal\_event\_detail} \sqsubseteq  (=1) \property{has\_signal\_event\_signal\_ref}$\vspace{0.1cm} \\
\par \vspace{-0.1cm} \noindent \textbf{Property}: $\property{has\_signal\_event\_signal\_ref}$\vspace{0.1cm} \\
\noindent \textbf{Label}: SignalRef\vspace{0.1cm} \\
\noindent \textbf{Description}: If the Trigger is a Signal, then 
	a Signal Shall be entered. The attributes of a Signal can be found in 
	Section B.11.17, "Signal," on page 277.\vspace{0.1cm} \\
\noindent \stepcounter{idaxiom} $AX\_\arabic{idaxiom}$ $\property{has\_signal\_event\_signal\_ref} \mbox{ has domain }\concept{signal\_event\_detail}$\vspace{0.1cm} \\
\noindent \stepcounter{idaxiom} $AX\_\arabic{idaxiom}$ $\property{has\_signal\_event\_signal\_ref} \mbox{ has range }\concept{signal}$\vspace{0.1cm} \\
\noindent \stepcounter{idaxiom} $AX\_\arabic{idaxiom}$ $\concept{terminate\_event\_detail} \equiv \concept{event\_detail} \sqcap \exists\property{has\_event\_detail\_type}.\{\instance{terminate\_event\_detail\_type}\}$\vspace{0.1cm} \\
\vspace{0.4cm} \hrule \vspace{0.2cm} \noindent \textbf{Class}: $\concept{terminate\_event\_detail}$\vspace{0.2cm} \hrule \vspace{0.2cm} 
\noindent \textbf{Label}: Terminate Event Detail\vspace{0.1cm} \\
\noindent \stepcounter{idaxiom} $AX\_\arabic{idaxiom}$ $\concept{timer\_event\_detail} \equiv \concept{event\_detail} \sqcap \exists\property{has\_event\_detail\_type}.\{\instance{timer\_event\_detail\_type}\}$\vspace{0.1cm} \\
\vspace{0.4cm} \hrule \vspace{0.2cm} \noindent \textbf{Class}: $\concept{timer\_event\_detail}$\vspace{0.2cm} \hrule \vspace{0.2cm} 
\noindent \textbf{Label}: Timer Event Detail\vspace{0.1cm} \\
\noindent \stepcounter{idaxiom} $AX\_\arabic{idaxiom}$ $\concept{timer\_event\_detail} \sqsubseteq  (=1) \property{has\_timer\_event\_time\_date} \sqcup  (=1) \property{has\_timer\_event\_time\_cycle}$\vspace{0.1cm} \\
\par \vspace{-0.1cm} \noindent \textbf{Property}: $\property{has\_timer\_event\_time\_date}$\vspace{0.1cm} \\
\noindent \textbf{Label}: TimeDate\vspace{0.1cm} \\
\noindent \textbf{Description}: If the Trigger is a Timer, then a 
	TimeDate MAY be entered. If a TimeDate is not entered, then a TimeCycle
	MUST be entered (see the attribute below). The attributes of a 
	TimeDateExpression can be found in Section B.11.18 on page 277\vspace{0.1cm} \\
\noindent \stepcounter{idaxiom} $AX\_\arabic{idaxiom}$ $\property{has\_timer\_event\_time\_date} \mbox{ has domain }\concept{timer\_event\_detail}$\vspace{0.1cm} \\
\noindent \stepcounter{idaxiom} $AX\_\arabic{idaxiom}$ $\property{has\_timer\_event\_time\_date} \mbox{ has range }\concept{time\_date\_expression}$\vspace{0.1cm} \\
\par \vspace{-0.1cm} \noindent \textbf{Property}: $\property{has\_timer\_event\_time\_cycle}$\vspace{0.1cm} \\
\noindent \textbf{Label}: TimeCycle\vspace{0.1cm} \\
\noindent \textbf{Description}: If the Trigger is a Timer, then a 
	TimeCycle MAY be entered. If a TimeCycle is not entered, then a 
	TimeDate MUST be entered (see the attribute above).\vspace{0.1cm} \\
\noindent \stepcounter{idaxiom} $AX\_\arabic{idaxiom}$ $\property{has\_timer\_event\_time\_cycle} \mbox{ has domain }\concept{timer\_event\_detail}$\vspace{0.1cm} \\
\noindent \stepcounter{idaxiom} $AX\_\arabic{idaxiom}$ $\property{has\_timer\_event\_time\_cycle} \mbox{ has range }\concept{time\_date\_expression}$\vspace{0.1cm} \\
\vspace{0.4cm} \hrule \vspace{0.2cm} \noindent \textbf{Class}: $\concept{expression}$\vspace{0.2cm} \hrule \vspace{0.2cm} 
\noindent \textbf{Label}: Expression\vspace{0.1cm} \\
\noindent \textbf{Description}: Expression, which is used in the definition of 
	attributes for Start Event, Intermediate Event, Activity, Complex 
	Gateway, and Sequence Flow, and which extends the set of common BPMN 
	Element attributes (see Table B.2).\vspace{0.1cm} \\
\noindent \stepcounter{idaxiom} $AX\_\arabic{idaxiom}$ $\concept{expression} \sqsubseteq  (=1) \property{has\_expression\_expression\_body}$\vspace{0.1cm} \\
\par \vspace{-0.1cm} \noindent \textbf{Property}: $\property{has\_expression\_expression\_body}$\vspace{0.1cm} \\
\noindent \textbf{Label}: ExpressionBody\vspace{0.1cm} \\
\noindent \textbf{Description}: An ExpressionBody MUST be 
	entered to provide the text of the expression, which will be written in 
	the language defined by the ExpressionLanguage attribute.\vspace{0.1cm} \\
\noindent \stepcounter{idaxiom} $AX\_\arabic{idaxiom}$ $\property{has\_expression\_expression\_body} \mbox{ has domain }\concept{expression}$\vspace{0.1cm} \\
\noindent \stepcounter{idaxiom} $AX\_\arabic{idaxiom}$ $\property{has\_expression\_expression\_body} \mbox{ has range }\datatype{xsd:string}$\vspace{0.1cm} \\
\noindent \stepcounter{idaxiom} $AX\_\arabic{idaxiom}$ $\concept{expression} \sqsubseteq  (=1) \property{has\_expression\_expression\_language}$\vspace{0.1cm} \\
\par \vspace{-0.1cm} \noindent \textbf{Property}: $\property{has\_expression\_expression\_language}$\vspace{0.1cm} \\
\noindent \textbf{Label}: ExpressionLanguage\vspace{0.1cm} \\
\noindent \textbf{Description}: A Language MUST be provided
	to identify the language of the ExpressionBody. The value of the 
	ExpressionLanguage should follow the naming conventions for the version 
	of the specified language.\vspace{0.1cm} \\
\noindent \stepcounter{idaxiom} $AX\_\arabic{idaxiom}$ $\property{has\_expression\_expression\_language} \mbox{ has domain }\concept{expression}$\vspace{0.1cm} \\
\noindent \stepcounter{idaxiom} $AX\_\arabic{idaxiom}$ $\property{has\_expression\_expression\_language} \mbox{ has range }\datatype{xsd:string}$\vspace{0.1cm} \\
\noindent \stepcounter{idaxiom} $AX\_\arabic{idaxiom}$ $\concept{time\_date\_expression} \sqsubseteq \concept{expression}$\vspace{0.1cm} \\
\vspace{0.4cm} \hrule \vspace{0.2cm} \noindent \textbf{Class}: $\concept{time\_date\_expression}$\vspace{0.2cm} \hrule \vspace{0.2cm} 
\noindent \textbf{Label}: TimeDate Expression\vspace{0.1cm} \\
\noindent \textbf{Description}: The TimeDateExpression supporting 
	element is a sub-type of the Expression Element (Expression on page 273)
	and uses all the attributes of the Expression Element.\vspace{0.1cm} \\
\vspace{0.4cm} \hrule \vspace{0.2cm} \noindent \textbf{Class}: $\concept{gate}$\vspace{0.2cm} \hrule \vspace{0.2cm} 
\noindent \textbf{Label}: Gate\vspace{0.1cm} \\
\noindent \textbf{Description}: Gate, which is used in the definition of attributes for 
	Gateways, and which extends the set of common BPMN Element attributes 
	(see Table B.2).\vspace{0.1cm} \\
\noindent \stepcounter{idaxiom} $AX\_\arabic{idaxiom}$ $\concept{gate} \sqsubseteq  (=1) \property{has\_gate\_outgoing\_sequence\_flow\_ref}$\vspace{0.1cm} \\
\par \vspace{-0.1cm} \noindent \textbf{Property}: $\property{has\_gate\_outgoing\_sequence\_flow\_ref}$\vspace{0.1cm} \\
\noindent \textbf{Label}: OutgoingSequenceFlowRef\vspace{0.1cm} \\
\noindent \textbf{Description}: Each Gate MUST 
	have an associated (outgoing) Sequence Flow. The attributes of a 
	Sequence Flow can be found in the Section B.10.2 on page 264.
	For Exclusive Event-Based, Complex, and Parallel Gateways:
		The Sequence Flow MUST have its Condition attribute set to None 
		(there is not an evaluation of a condition expression).
	For Exclusive Data-Based, and Inclusive Gateways:
		The Sequence Flow MUST have its Condition attribute set to 
		Expression and MUST have a valid ConditionExpression. The 
		ConditionExpression MUST be unique for all the Gates within the 
		Gateway. If there is only one Gate (i.e., the Gateway is acting 
		only as a Merge), then Sequence Flow MUST have its Condition 
		set to None.
	For DefaultGates:
		The Sequence Flow MUST have its Condition attribute set to 
		Otherwise\vspace{0.1cm} \\
\noindent \stepcounter{idaxiom} $AX\_\arabic{idaxiom}$ $\property{has\_gate\_outgoing\_sequence\_flow\_ref} \mbox{ has domain }\concept{gate}$\vspace{0.1cm} \\
\noindent \stepcounter{idaxiom} $AX\_\arabic{idaxiom}$ $\property{has\_gate\_outgoing\_sequence\_flow\_ref} \mbox{ has range }\concept{sequence\_flow}$\vspace{0.1cm} \\
\par \vspace{-0.1cm} \noindent \textbf{Property}: $\property{has\_gate\_assignments}$\vspace{0.1cm} \\
\noindent \textbf{Label}: Assignments\vspace{0.1cm} \\
\noindent \textbf{Description}: One or more assignment expressions MAY be
	made for each Gate. The Assignment SHALL be performed when the Gate is 
	selected. The Assignment is defined in the Section B.11.3 on page 269.\vspace{0.1cm} \\
\noindent \stepcounter{idaxiom} $AX\_\arabic{idaxiom}$ $\property{has\_gate\_assignments} \mbox{ has domain }\concept{gate}$\vspace{0.1cm} \\
\noindent \stepcounter{idaxiom} $AX\_\arabic{idaxiom}$ $\property{has\_gate\_assignments} \mbox{ has range }\concept{assignment}$\vspace{0.1cm} \\
\vspace{0.4cm} \hrule \vspace{0.2cm} \noindent \textbf{Class}: $\concept{input\_set}$\vspace{0.2cm} \hrule \vspace{0.2cm} 
\noindent \textbf{Label}: Input Set\vspace{0.1cm} \\
\noindent \textbf{Description}: InputSet, which is used in the definition of common
	attributes for Activities and for attributes of a Process, and which 
	extends the set of common BPMN Element attributes (see Table B.2).\vspace{0.1cm} \\
\noindent \stepcounter{idaxiom} $AX\_\arabic{idaxiom}$ $\concept{input\_set} \sqsubseteq (\exists\property{has\_input\_set\_artifact\_input}.\concept{artifact\_input}) \sqcup (\exists\property{has\_input\_set\_property\_input}.\concept{property})$\vspace{0.1cm} \\
\par \vspace{-0.1cm} \noindent \textbf{Property}: $\property{has\_input\_set\_artifact\_input}$\vspace{0.1cm} \\
\noindent \textbf{Label}: ArtifactInput\vspace{0.1cm} \\
\noindent \textbf{Description}: Zero or more ArtifactInputs MAY 
	be defined for each InputSet. For the combination of ArtifactInputs and 
	PropertyInputs, there MUST be at least one item defined for the 
	InputSet. An ArtifactInput is an Artifact, usually a Data Object. Note 
	that the Artifacts MAY also be displayed on the diagram and MAY be 
	connected to the activity through an Association--however, it is not 
	required for them to be displayed. Further details about the definition
	of an ArtifactInput can be found in Section B.11.1 on page 268.\vspace{0.1cm} \\
\noindent \stepcounter{idaxiom} $AX\_\arabic{idaxiom}$ $\property{has\_input\_set\_artifact\_input} \mbox{ has domain }\concept{input\_set}$\vspace{0.1cm} \\
\noindent \stepcounter{idaxiom} $AX\_\arabic{idaxiom}$ $\property{has\_input\_set\_artifact\_input} \mbox{ has range }\concept{artifact\_input}$\vspace{0.1cm} \\
\par \vspace{-0.1cm} \noindent \textbf{Property}: $\property{has\_input\_set\_property\_input}$\vspace{0.1cm} \\
\noindent \textbf{Label}: PropertyInput\vspace{0.1cm} \\
\noindent \textbf{Description}: Zero or more PropertyInputs MAY 
	be defined for each InputSet. For the combination of ArtifactInputs and
	PropertyInputs, there MUST be at least one item defined for the 
	InputSet.\vspace{0.1cm} \\
\noindent \stepcounter{idaxiom} $AX\_\arabic{idaxiom}$ $\property{has\_input\_set\_property\_input} \mbox{ has domain }\concept{input\_set}$\vspace{0.1cm} \\
\noindent \stepcounter{idaxiom} $AX\_\arabic{idaxiom}$ $\property{has\_input\_set\_property\_input} \mbox{ has range }\concept{property}$\vspace{0.1cm} \\
\vspace{0.4cm} \hrule \vspace{0.2cm} \noindent \textbf{Class}: $\concept{message}$\vspace{0.2cm} \hrule \vspace{0.2cm} 
\noindent \textbf{Label}: Message\vspace{0.1cm} \\
\noindent \textbf{Description}: Message, which is used in the definition of attributes
	for a Start Event, End Event, Intermediate Event, Task, and Message 
	Flow, and which extends the set of common BPMN Element attributes (see 
	Table B.2)\vspace{0.1cm} \\
\noindent \stepcounter{idaxiom} $AX\_\arabic{idaxiom}$ $\concept{message} \sqsubseteq  (=1) \property{has\_message\_name}$\vspace{0.1cm} \\
\par \vspace{-0.1cm} \noindent \textbf{Property}: $\property{has\_message\_name}$\vspace{0.1cm} \\
\noindent \textbf{Label}: Name\vspace{0.1cm} \\
\noindent \textbf{Description}: Name is an attribute that is text description
	of the Message.\vspace{0.1cm} \\
\noindent \stepcounter{idaxiom} $AX\_\arabic{idaxiom}$ $\property{has\_message\_name} \mbox{ has domain }\concept{message}$\vspace{0.1cm} \\
\noindent \stepcounter{idaxiom} $AX\_\arabic{idaxiom}$ $\property{has\_message\_name} \mbox{ has range }\datatype{xsd:string}$\vspace{0.1cm} \\
\par \vspace{-0.1cm} \noindent \textbf{Property}: $\property{has\_message\_property}$\vspace{0.1cm} \\
\noindent \textbf{Label}: Property\vspace{0.1cm} \\
\noindent \textbf{Description}: Multiple Properties MAY entered for the 
	Message. The attributes of a Property can be found in "Property on 
	page 276."\vspace{0.1cm} \\
\noindent \stepcounter{idaxiom} $AX\_\arabic{idaxiom}$ $\property{has\_message\_property} \mbox{ has domain }\concept{message}$\vspace{0.1cm} \\
\noindent \stepcounter{idaxiom} $AX\_\arabic{idaxiom}$ $\property{has\_message\_property} \mbox{ has range }\concept{property}$\vspace{0.1cm} \\
\noindent \stepcounter{idaxiom} $AX\_\arabic{idaxiom}$ $\concept{message} \sqsubseteq  (=1) \property{has\_message\_from\_ref}$\vspace{0.1cm} \\
\par \vspace{-0.1cm} \noindent \textbf{Property}: $\property{has\_message\_from\_ref}$\vspace{0.1cm} \\
\noindent \textbf{Label}: FromRef\vspace{0.1cm} \\
\noindent \textbf{Description}: This defines the source of the Message. 
	The attributes for a Participant can be found in "Participant on page 
	276."\vspace{0.1cm} \\
\noindent \stepcounter{idaxiom} $AX\_\arabic{idaxiom}$ $\property{has\_message\_from\_ref} \mbox{ has domain }\concept{message}$\vspace{0.1cm} \\
\noindent \stepcounter{idaxiom} $AX\_\arabic{idaxiom}$ $\property{has\_message\_from\_ref} \mbox{ has range }\concept{participant}$\vspace{0.1cm} \\
\noindent \stepcounter{idaxiom} $AX\_\arabic{idaxiom}$ $\concept{message} \sqsubseteq  (=1) \property{has\_message\_to\_ref}$\vspace{0.1cm} \\
\par \vspace{-0.1cm} \noindent \textbf{Property}: $\property{has\_message\_to\_ref}$\vspace{0.1cm} \\
\noindent \textbf{Label}: ToRef\vspace{0.1cm} \\
\noindent \textbf{Description}: This defines the source of the Message. 
	The attributes for a Participant can be found in "Participant on page 
	276."\vspace{0.1cm} \\
\noindent \stepcounter{idaxiom} $AX\_\arabic{idaxiom}$ $\property{has\_message\_to\_ref} \mbox{ has domain }\concept{message}$\vspace{0.1cm} \\
\noindent \stepcounter{idaxiom} $AX\_\arabic{idaxiom}$ $\property{has\_message\_to\_ref} \mbox{ has range }\concept{participant}$\vspace{0.1cm} \\
\vspace{0.4cm} \hrule \vspace{0.2cm} \noindent \textbf{Class}: $\concept{object}$\vspace{0.2cm} \hrule \vspace{0.2cm} 
\noindent \textbf{Label}: Object\vspace{0.1cm} \\
\noindent \textbf{Description}: Object, which is used in the definition of attributes 
	for all graphical elements.\vspace{0.1cm} \\
\noindent \stepcounter{idaxiom} $AX\_\arabic{idaxiom}$ $\concept{object} \sqsubseteq  (=1) \property{has\_object\_id}$\vspace{0.1cm} \\
\par \vspace{-0.1cm} \noindent \textbf{Property}: $\property{has\_object\_id}$\vspace{0.1cm} \\
\noindent \textbf{Label}: Id\vspace{0.1cm} \\
\noindent \textbf{Description}: The Id attribute provides a unique identifier 
	for all objects on a diagram. That is, each object MUST have a different
	value for the ObjectId attribute.\vspace{0.1cm} \\
\noindent \stepcounter{idaxiom} $AX\_\arabic{idaxiom}$ $\property{has\_object\_id} \mbox{ has range }\datatype{xsd:string}$\vspace{0.1cm} \\
\noindent \stepcounter{idaxiom} $AX\_\arabic{idaxiom}$ $\property{has\_object\_id} \mbox{ has domain }\concept{object}$\vspace{0.1cm} \\
\vspace{0.4cm} \hrule \vspace{0.2cm} \noindent \textbf{Class}: $\concept{output\_set}$\vspace{0.2cm} \hrule \vspace{0.2cm} 
\noindent \textbf{Label}: Output Set\vspace{0.1cm} \\
\noindent \textbf{Description}: OutputSet, which is used in the definition of 
	common attributes for Activities and for attributes of a Process, and 
	which extends the set of common BPMN Element attributes 
	(see Table B.2).\vspace{0.1cm} \\
\noindent \stepcounter{idaxiom} $AX\_\arabic{idaxiom}$ $\concept{output\_set} \sqsubseteq (\exists\property{has\_output\_set\_artifact\_output}.\concept{artifact\_output}) \sqcup \\ (\exists\property{has\_output\_set\_property\_output}.\concept{property})$\vspace{0.1cm} \\
\par \vspace{-0.1cm} \noindent \textbf{Property}: $\property{has\_output\_set\_artifact\_output}$\vspace{0.1cm} \\
\noindent \textbf{Label}: ArtifactOutput\vspace{0.1cm} \\
\noindent \textbf{Description}: Zero or more ArtifactOutputs 
	MAY be defined for each InputSet. For the combination of ArtifactOutputs
	and PropertyOutputs, there MUST be at least one item defined for the 
	OutputSet. An ArtifactOutput is an Artifact, usually a Data Object. Note
	that the Artifacts MAY also be displayed on the diagram and MAY be 
	connected to the activity through an Association--however, it is not
	required for them to be displayed. Further details about the definition 
	of an ArtifactOutput can be found in Section B.11.2 on page 268.\vspace{0.1cm} \\
\noindent \stepcounter{idaxiom} $AX\_\arabic{idaxiom}$ $\property{has\_output\_set\_artifact\_output} \mbox{ has domain }\concept{output\_set}$\vspace{0.1cm} \\
\noindent \stepcounter{idaxiom} $AX\_\arabic{idaxiom}$ $\property{has\_output\_set\_artifact\_output} \mbox{ has range }\concept{artifact\_output}$\vspace{0.1cm} \\
\par \vspace{-0.1cm} \noindent \textbf{Property}: $\property{has\_output\_set\_property\_output}$\vspace{0.1cm} \\
\noindent \textbf{Label}: PropertyOutput\vspace{0.1cm} \\
\noindent \textbf{Description}: Zero or more PropertyOutputs 
	MAY be defined for each InputSet. For the combination of ArtifactOutputs
	and PropertyOutputs, there MUST be at least one item defined for the 
	OutputSet.\vspace{0.1cm} \\
\noindent \stepcounter{idaxiom} $AX\_\arabic{idaxiom}$ $\property{has\_output\_set\_property\_output} \mbox{ has domain }\concept{output\_set}$\vspace{0.1cm} \\
\noindent \stepcounter{idaxiom} $AX\_\arabic{idaxiom}$ $\property{has\_output\_set\_property\_output} \mbox{ has range }\concept{property}$\vspace{0.1cm} \\
\vspace{0.4cm} \hrule \vspace{0.2cm} \noindent \textbf{Class}: $\concept{participant}$\vspace{0.2cm} \hrule \vspace{0.2cm} 
\noindent \textbf{Label}: Participant\vspace{0.1cm} \\
\noindent \textbf{Description}: Participant, which is used in the definition of 
	attributes for a Pool, Message, and Web service, and which extends the 
	set of common BPMN Element attributes (see Table B.2).\vspace{0.1cm} \\
\noindent \stepcounter{idaxiom} $AX\_\arabic{idaxiom}$ $\concept{participant} \sqsubseteq  (=1) \property{has\_participant\_participant\_type}$\vspace{0.1cm} \\
\par \vspace{-0.1cm} \noindent \textbf{Property}: $\property{has\_participant\_participant\_type}$\vspace{0.1cm} \\
\noindent \textbf{Label}: ParticipantType\vspace{0.1cm} \\
\noindent \textbf{Description}: \vspace{0.1cm} \\
\noindent \stepcounter{idaxiom} $AX\_\arabic{idaxiom}$ $\property{has\_participant\_participant\_type} \mbox{ has range }\datatype{xsd:string}\{"\datainstance{Role}","\datainstance{Entity}"\}$\vspace{0.1cm} \\
\noindent \stepcounter{idaxiom} $AX\_\arabic{idaxiom}$ $\property{has\_participant\_participant\_type} \mbox{ has domain }\concept{participant}$\vspace{0.1cm} \\
\noindent \stepcounter{idaxiom} $AX\_\arabic{idaxiom}$ $\concept{participant} \sqsubseteq (\exists\property{has\_participant\_participant\_type}.\{"\datainstance{Role}"\} \sqcap  (=1) \property{has\_participant\_role\_ref}) \sqcup \\ (\exists\property{has\_participant\_participant\_type}.\{"\datainstance{Entity}"\} \sqcap  (=1) \property{has\_participant\_entity\_ref})$\vspace{0.1cm} \\
\par \vspace{-0.1cm} \noindent \textbf{Property}: $\property{has\_participant\_role\_ref}$\vspace{0.1cm} \\
\noindent \textbf{Label}: RoleRef\vspace{0.1cm} \\
\noindent \textbf{Description}: If the ParticipantType = Role, then 
	a Role MUST be identified. The attributes for a Role can be found in 
	"Role on page 276."\vspace{0.1cm} \\
\noindent \stepcounter{idaxiom} $AX\_\arabic{idaxiom}$ $\property{has\_participant\_role\_ref} \mbox{ has domain }\concept{participant}$\vspace{0.1cm} \\
\noindent \stepcounter{idaxiom} $AX\_\arabic{idaxiom}$ $\property{has\_participant\_role\_ref} \mbox{ has range }\concept{role}$\vspace{0.1cm} \\
\par \vspace{-0.1cm} \noindent \textbf{Property}: $\property{has\_participant\_entity\_ref}$\vspace{0.1cm} \\
\noindent \textbf{Label}: EntityRef\vspace{0.1cm} \\
\noindent \textbf{Description}: If the ParticipantType = Entity, 
	then an Entity MUST be identified. The attributes for an Entity can be 
	found in "Condition on page 269."\vspace{0.1cm} \\
\noindent \stepcounter{idaxiom} $AX\_\arabic{idaxiom}$ $\property{has\_participant\_entity\_ref} \mbox{ has domain }\concept{participant}$\vspace{0.1cm} \\
\noindent \stepcounter{idaxiom} $AX\_\arabic{idaxiom}$ $\property{has\_participant\_entity\_ref} \mbox{ has range }\concept{entity}$\vspace{0.1cm} \\
\vspace{0.4cm} \hrule \vspace{0.2cm} \noindent \textbf{Class}: $\concept{property}$\vspace{0.2cm} \hrule \vspace{0.2cm} 
\noindent \textbf{Label}: Property\vspace{0.1cm} \\
\noindent \textbf{Description}: Property, which is used in the definition of 
	attributes for a Process and common activity attributes, and which
	extends the set of common BPMN Element attributes (see Table B.2).\vspace{0.1cm} \\
\noindent \stepcounter{idaxiom} $AX\_\arabic{idaxiom}$ $\concept{property} \sqsubseteq  (=1) \property{has\_property\_name}$\vspace{0.1cm} \\
\par \vspace{-0.1cm} \noindent \textbf{Property}: $\property{has\_property\_name}$\vspace{0.1cm} \\
\noindent \textbf{Label}: Name\vspace{0.1cm} \\
\noindent \textbf{Description}: Each Property has a Name (e.g., 
	name="Customer Name").\vspace{0.1cm} \\
\noindent \stepcounter{idaxiom} $AX\_\arabic{idaxiom}$ $\property{has\_property\_name} \mbox{ has domain }\concept{property}$\vspace{0.1cm} \\
\noindent \stepcounter{idaxiom} $AX\_\arabic{idaxiom}$ $\property{has\_property\_name} \mbox{ has range }\datatype{xsd:string}$\vspace{0.1cm} \\
\noindent \stepcounter{idaxiom} $AX\_\arabic{idaxiom}$ $\concept{property} \sqsubseteq  (=1) \property{has\_property\_type}$\vspace{0.1cm} \\
\par \vspace{-0.1cm} \noindent \textbf{Property}: $\property{has\_property\_type}$\vspace{0.1cm} \\
\noindent \textbf{Label}: Type\vspace{0.1cm} \\
\noindent \textbf{Description}: Each Property has a Type (e.g., 
	type="String"). Properties may be defined hierarchically.\vspace{0.1cm} \\
\noindent \stepcounter{idaxiom} $AX\_\arabic{idaxiom}$ $\property{has\_property\_type} \mbox{ has domain }\concept{property}$\vspace{0.1cm} \\
\noindent \stepcounter{idaxiom} $AX\_\arabic{idaxiom}$ $\property{has\_property\_type} \mbox{ has range }\datatype{xsd:string}$\vspace{0.1cm} \\
\noindent \stepcounter{idaxiom} $AX\_\arabic{idaxiom}$ $\concept{property} \sqsubseteq (\geq1) \property{has\_property\_value}$\vspace{0.1cm} \\
\par \vspace{-0.1cm} \noindent \textbf{Property}: $\property{has\_property\_value}$\vspace{0.1cm} \\
\noindent \textbf{Label}: Value\vspace{0.1cm} \\
\noindent \textbf{Description}: Each Property MAY have a Value specified.\vspace{0.1cm} \\
\noindent \stepcounter{idaxiom} $AX\_\arabic{idaxiom}$ $\property{has\_property\_value} \mbox{ has domain }\concept{property}$\vspace{0.1cm} \\
\noindent \stepcounter{idaxiom} $AX\_\arabic{idaxiom}$ $\property{has\_property\_value} \mbox{ has range }\concept{expression}$\vspace{0.1cm} \\
\noindent \stepcounter{idaxiom} $AX\_\arabic{idaxiom}$ $\concept{property} \sqsubseteq (\geq1) \property{has\_property\_correlation}$\vspace{0.1cm} \\
\par \vspace{-0.1cm} \noindent \textbf{Property}: $\property{has\_property\_correlation}$\vspace{0.1cm} \\
\noindent \textbf{Label}: Correlation\vspace{0.1cm} \\
\noindent \textbf{Description}: If the Correlation attribute is set 
	to True, then the Property is marked to be used for correlation 
	(e.g., for incoming Messages).\vspace{0.1cm} \\
\noindent \stepcounter{idaxiom} $AX\_\arabic{idaxiom}$ $\property{has\_property\_correlation} \mbox{ has domain }\concept{property}$\vspace{0.1cm} \\
\noindent \stepcounter{idaxiom} $AX\_\arabic{idaxiom}$ $\property{has\_property\_correlation} \mbox{ has range }\datatype{xsd:boolean}$\vspace{0.1cm} \\
\vspace{0.4cm} \hrule \vspace{0.2cm} \noindent \textbf{Class}: $\concept{role}$\vspace{0.2cm} \hrule \vspace{0.2cm} 
\noindent \textbf{Label}: Role\vspace{0.1cm} \\
\noindent \textbf{Description}: Role, which is used in the definition of attributes for 
	a Participant, and which extends the set of common BPMN Element 
	attributes (see Table B.2).\vspace{0.1cm} \\
\noindent \stepcounter{idaxiom} $AX\_\arabic{idaxiom}$ $\concept{role} \sqsubseteq  (=1) \property{has\_role\_name}$\vspace{0.1cm} \\
\par \vspace{-0.1cm} \noindent \textbf{Property}: $\property{has\_role\_name}$\vspace{0.1cm} \\
\noindent \textbf{Label}: Name\vspace{0.1cm} \\
\noindent \textbf{Description}: Name is an attribute that is text description of
	the Role.\vspace{0.1cm} \\
\noindent \stepcounter{idaxiom} $AX\_\arabic{idaxiom}$ $\property{has\_role\_name} \mbox{ has domain }\concept{role}$\vspace{0.1cm} \\
\noindent \stepcounter{idaxiom} $AX\_\arabic{idaxiom}$ $\property{has\_role\_name} \mbox{ has range }\datatype{xsd:string}$\vspace{0.1cm} \\
\vspace{0.4cm} \hrule \vspace{0.2cm} \noindent \textbf{Class}: $\concept{signal}$\vspace{0.2cm} \hrule \vspace{0.2cm} 
\noindent \textbf{Label}: signal\vspace{0.1cm} \\
\noindent \textbf{Description}: Signal, which is used in the definition of attributes 
	for a Start Event, End Event, Intermediate Event, and which extends the 
	set of common BPMN Element attributes (see Table B.2).\vspace{0.1cm} \\
\noindent \stepcounter{idaxiom} $AX\_\arabic{idaxiom}$ $\concept{signal} \sqsubseteq  (=1) \property{has\_signal\_name}$\vspace{0.1cm} \\
\par \vspace{-0.1cm} \noindent \textbf{Property}: $\property{has\_signal\_name}$\vspace{0.1cm} \\
\noindent \textbf{Label}: Name\vspace{0.1cm} \\
\noindent \textbf{Description}: Name is an attribute that is text description 
	of the Signal.\vspace{0.1cm} \\
\noindent \stepcounter{idaxiom} $AX\_\arabic{idaxiom}$ $\property{has\_signal\_name} \mbox{ has domain }\concept{signal}$\vspace{0.1cm} \\
\noindent \stepcounter{idaxiom} $AX\_\arabic{idaxiom}$ $\property{has\_signal\_name} \mbox{ has range }\datatype{xsd:string}$\vspace{0.1cm} \\
\par \vspace{-0.1cm} \noindent \textbf{Property}: $\property{has\_signal\_property}$\vspace{0.1cm} \\
\noindent \textbf{Label}: Property\vspace{0.1cm} \\
\noindent \textbf{Description}: Multiple Properties MAY entered for the 
	Signal. The attributes of a Property can be found in Property on page 
	276.\vspace{0.1cm} \\
\noindent \stepcounter{idaxiom} $AX\_\arabic{idaxiom}$ $\property{has\_signal\_property} \mbox{ has domain }\concept{signal}$\vspace{0.1cm} \\
\noindent \stepcounter{idaxiom} $AX\_\arabic{idaxiom}$ $\property{has\_signal\_property} \mbox{ has range }\concept{property}$\vspace{0.1cm} \\
\vspace{0.4cm} \hrule \vspace{0.2cm} \noindent \textbf{Class}: $\concept{transaction}$\vspace{0.2cm} \hrule \vspace{0.2cm} 
\noindent \textbf{Label}: Transaction\vspace{0.1cm} \\
\noindent \textbf{Description}: Transaction, which is used in the definition of 
	attributes for a Sub-Process, and which extends the set of common BPMN 
	Element attributes (see Table B.2).\vspace{0.1cm} \\
\noindent \stepcounter{idaxiom} $AX\_\arabic{idaxiom}$ $\concept{transaction} \sqsubseteq  (=1) \property{has\_transaction\_transaction\_id}$\vspace{0.1cm} \\
\par \vspace{-0.1cm} \noindent \textbf{Property}: $\property{has\_transaction\_transaction\_id}$\vspace{0.1cm} \\
\noindent \textbf{Label}: TransactionId\vspace{0.1cm} \\
\noindent \textbf{Description}: The TransactionId attribute 
	provides an identifier for the Transactions used within a diagram.\vspace{0.1cm} \\
\noindent \stepcounter{idaxiom} $AX\_\arabic{idaxiom}$ $\property{has\_transaction\_transaction\_id} \mbox{ has range }\datatype{xsd:string}$\vspace{0.1cm} \\
\noindent \stepcounter{idaxiom} $AX\_\arabic{idaxiom}$ $\property{has\_transaction\_transaction\_id} \mbox{ has domain }\concept{transaction}$\vspace{0.1cm} \\
\noindent \stepcounter{idaxiom} $AX\_\arabic{idaxiom}$ $\concept{transaction} \sqsubseteq  (=1) \property{has\_transaction\_transaction\_protocol}$\vspace{0.1cm} \\
\par \vspace{-0.1cm} \noindent \textbf{Property}: $\property{has\_transaction\_transaction\_protocol}$\vspace{0.1cm} \\
\noindent \textbf{Label}: TransactionProtocol\vspace{0.1cm} \\
\noindent \textbf{Description}: This identifies the 
	Protocol (e.g., WS-Transaction or BTP) that will be used to control the
	transactional behavior of the Sub-Process.\vspace{0.1cm} \\
\noindent \stepcounter{idaxiom} $AX\_\arabic{idaxiom}$ $\property{has\_transaction\_transaction\_protocol} \mbox{ has range }\datatype{xsd:string}$\vspace{0.1cm} \\
\noindent \stepcounter{idaxiom} $AX\_\arabic{idaxiom}$ $\property{has\_transaction\_transaction\_protocol} \mbox{ has domain }\concept{transaction}$\vspace{0.1cm} \\
\noindent \stepcounter{idaxiom} $AX\_\arabic{idaxiom}$ $\concept{transaction} \sqsubseteq  (=1) \property{has\_transaction\_transaction\_method}$\vspace{0.1cm} \\
\par \vspace{-0.1cm} \noindent \textbf{Property}: $\property{has\_transaction\_transaction\_method}$\vspace{0.1cm} \\
\noindent \textbf{Label}: TransactionMethod\vspace{0.1cm} \\
\noindent \textbf{Description}: TransactionMethod is an 
	attribute that defines the technique that will be used to undo a 
	Transaction that has been cancelled. The default is Compensate, but the
	attribute MAY be set to Store or Image.\vspace{0.1cm} \\
\noindent \stepcounter{idaxiom} $AX\_\arabic{idaxiom}$ $\property{has\_transaction\_transaction\_method} \mbox{ has range }\datatype{xsd:string}\{"\datainstance{Compensate}","\datainstance{Store}","\datainstance{Image}"\}$\vspace{0.1cm} \\
\noindent \stepcounter{idaxiom} $AX\_\arabic{idaxiom}$ $\property{has\_transaction\_transaction\_method} \mbox{ has domain }\concept{transaction}$\vspace{0.1cm} \\
\vspace{0.4cm} \hrule \vspace{0.2cm} \noindent \textbf{Class}: $\concept{web\_service}$\vspace{0.2cm} \hrule \vspace{0.2cm} 
\noindent \textbf{Label}: Web Service\vspace{0.1cm} \\
\noindent \textbf{Description}: Web Service, which is used in the definition of 
	attributes for Message Start Event, Message Intermediate Event, Message 
	End Event, Receive Task, Send Task, Service Task, and User Task, and 
	which extends the set of common BPMN Element attributes
	(see Table B.2).\vspace{0.1cm} \\
\noindent \stepcounter{idaxiom} $AX\_\arabic{idaxiom}$ $\concept{web\_service} \sqsubseteq  (=1) \property{has\_web\_service\_participant\_ref}$\vspace{0.1cm} \\
\par \vspace{-0.1cm} \noindent \textbf{Property}: $\property{has\_web\_service\_participant\_ref}$\vspace{0.1cm} \\
\noindent \textbf{Label}: ParticipantRef\vspace{0.1cm} \\
\noindent \textbf{Description}: A Participant for the Web 
	Service MUST be entered. The attributes for a Participant can be found 
	in "Participant on page 276."\vspace{0.1cm} \\
\noindent \stepcounter{idaxiom} $AX\_\arabic{idaxiom}$ $\property{has\_web\_service\_participant\_ref} \mbox{ has domain }\concept{web\_service}$\vspace{0.1cm} \\
\noindent \stepcounter{idaxiom} $AX\_\arabic{idaxiom}$ $\property{has\_web\_service\_participant\_ref} \mbox{ has range }\concept{participant}$\vspace{0.1cm} \\
\noindent \stepcounter{idaxiom} $AX\_\arabic{idaxiom}$ $\concept{web\_service} \sqsubseteq  (=1) \property{has\_web\_service\_interface}$\vspace{0.1cm} \\
\par \vspace{-0.1cm} \noindent \textbf{Property}: $\property{has\_web\_service\_interface}$\vspace{0.1cm} \\
\noindent \textbf{Label}: Interface\vspace{0.1cm} \\
\noindent \textbf{Description}: (aka portType) An Interface for the 
	Web Service MUST be entered.\vspace{0.1cm} \\
\noindent \stepcounter{idaxiom} $AX\_\arabic{idaxiom}$ $\property{has\_web\_service\_interface} \mbox{ has domain }\concept{web\_service}$\vspace{0.1cm} \\
\noindent \stepcounter{idaxiom} $AX\_\arabic{idaxiom}$ $\property{has\_web\_service\_interface} \mbox{ has range }\datatype{xsd:string}$\vspace{0.1cm} \\
\noindent \stepcounter{idaxiom} $AX\_\arabic{idaxiom}$ $\concept{web\_service} \sqsubseteq (\leq1) \property{has\_web\_service\_type}$\vspace{0.1cm} \\
\par \vspace{-0.1cm} \noindent \textbf{Property}: $\property{has\_web\_service\_operation}$\vspace{0.1cm} \\
\noindent \textbf{Label}: Operation\vspace{0.1cm} \\
\noindent \textbf{Description}: One or more Operations for the Web 
	Service MUST be entered.\vspace{0.1cm} \\
\noindent \stepcounter{idaxiom} $AX\_\arabic{idaxiom}$ $\property{has\_web\_service\_operation} \mbox{ has domain }\concept{web\_service}$\vspace{0.1cm} \\
\noindent \stepcounter{idaxiom} $AX\_\arabic{idaxiom}$ $\property{has\_web\_service\_operation} \mbox{ has range }\datatype{xsd:string}$\vspace{0.1cm} \\
\vspace{0.4cm} \hrule \vspace{0.2cm} \noindent \textbf{Class}: $\concept{process}$\vspace{0.2cm} \hrule \vspace{0.2cm} 
\noindent \textbf{Label}: Process\vspace{0.1cm} \\
\noindent \textbf{Description}: A Process is an activity performed within or across 
	companies or organizations. In BPMN a Process is depicted as a graph of 
	Flow Objects, which are a set of other activities and the controls that
	sequence them. The concept of process is intrinsically hierarchical. 
	Processes may be defined at any level from enterprise-wide processes to
	processes performed by a single person. Low-level processes may be
	grouped together to achieve a common business goal. Note that BPMN 
	defines the term Process fairly specifically and defines a Business 
	Process more generically as a set of activities that are performed 
	within an organization or across organizations. Thus a Business Process,
	as shown in a Business Process Diagram, may contain more than one 
	separate Process. Each Process may have its own Sub-Processes and would
	be contained within a Pool (Section B.8.2, on page 260). The individual
	Processes would be independent in terms of Sequence Flow, but could have
	Message Flow connecting them.\vspace{0.1cm} \\
\noindent \stepcounter{idaxiom} $AX\_\arabic{idaxiom}$ $\concept{process} \sqsubseteq  (=1) \property{has\_process\_name}$\vspace{0.1cm} \\
\par \vspace{-0.1cm} \noindent \textbf{Property}: $\property{has\_process\_name}$\vspace{0.1cm} \\
\noindent \textbf{Label}: Name\vspace{0.1cm} \\
\noindent \textbf{Description}: Name is an attribute that is a text 
	description of the object.\vspace{0.1cm} \\
\noindent \stepcounter{idaxiom} $AX\_\arabic{idaxiom}$ $\property{has\_process\_name} \mbox{ has domain }\concept{process}$\vspace{0.1cm} \\
\noindent \stepcounter{idaxiom} $AX\_\arabic{idaxiom}$ $\property{has\_process\_name} \mbox{ has range }\datatype{xsd:string}$\vspace{0.1cm} \\
\noindent \stepcounter{idaxiom} $AX\_\arabic{idaxiom}$ $\concept{process} \sqsubseteq  (=1) \property{has\_process\_process\_type}$\vspace{0.1cm} \\
\par \vspace{-0.1cm} \noindent \textbf{Property}: $\property{has\_process\_process\_type}$\vspace{0.1cm} \\
\noindent \textbf{Label}: process\_type\vspace{0.1cm} \\
\noindent \textbf{Description}: ProcessType is an attribute that
	provides information about which lower-level language the Pool will be 
	mapped. By default, the ProcessType is None (or undefined).\vspace{0.1cm} \\
\noindent \stepcounter{idaxiom} $AX\_\arabic{idaxiom}$ $\property{has\_process\_process\_type} \mbox{ has domain }\concept{process}$\vspace{0.1cm} \\
\noindent \stepcounter{idaxiom} $AX\_\arabic{idaxiom}$ $\property{has\_process\_process\_type} \mbox{ has range }\datatype{xsd:string}\{"\datainstance{None}","\datainstance{Private}","\datainstance{Abstract}","\datainstance{Collaboration}"\}$\vspace{0.1cm} \\
\noindent \stepcounter{idaxiom} $AX\_\arabic{idaxiom}$ $\concept{process} \sqsubseteq  (=1) \property{has\_process\_status}$\vspace{0.1cm} \\
\par \vspace{-0.1cm} \noindent \textbf{Property}: $\property{has\_process\_status}$\vspace{0.1cm} \\
\noindent \textbf{Label}: Status\vspace{0.1cm} \\
\noindent \textbf{Description}: The Status of a Process is determined when 
	the Process is being executed by a process engine. The Status of a 
	Process can be used within Assignment Expressions.\vspace{0.1cm} \\
\noindent \stepcounter{idaxiom} $AX\_\arabic{idaxiom}$ $\property{has\_process\_status} \mbox{ has domain }\concept{process}$\vspace{0.1cm} \\
\noindent \stepcounter{idaxiom} $AX\_\arabic{idaxiom}$ $\property{has\_process\_status} \mbox{ has range }\datatype{xsd:string}\{"\datainstance{None}","\datainstance{Ready}","\datainstance{Active}","\datainstance{Cancelled}","\datainstance{Aborting}", \\ "\datainstance{Aborted}","\datainstance{Completing}","\datainstance{Completed}"\}$\vspace{0.1cm} \\
\par \vspace{-0.1cm} \noindent \textbf{Property}: $\property{has\_process\_graphical\_elements}$\vspace{0.1cm} \\
\noindent \textbf{Label}: Graphical Elements\vspace{0.1cm} \\
\noindent \textbf{Description}: The GraphicalElements attribute
	identifies all of the objects (e.g., Events, Activities, Gateways, and 
	Artifacts) that are contained within the Process.\vspace{0.1cm} \\
\noindent \stepcounter{idaxiom} $AX\_\arabic{idaxiom}$ $\property{has\_process\_graphical\_elements} \mbox{ has domain }\concept{process}$\vspace{0.1cm} \\
\noindent \stepcounter{idaxiom} $AX\_\arabic{idaxiom}$ $\property{has\_process\_graphical\_elements} \mbox{ has range }\concept{graphical\_element}$\vspace{0.1cm} \\
\par \vspace{-0.1cm} \noindent \textbf{Property}: $\property{has\_process\_assignments}$\vspace{0.1cm} \\
\noindent \textbf{Label}: Assignments\vspace{0.1cm} \\
\noindent \textbf{Description}: One or more assignment expressions 
	MAY be made for the object. The Assignment SHALL be performed as defined
	by the AssignTime attribute (see below). The details of Assignment is 
	defined in "Assignment on page 269.".\vspace{0.1cm} \\
\noindent \stepcounter{idaxiom} $AX\_\arabic{idaxiom}$ $\property{has\_process\_assignments} \mbox{ has domain }\concept{process}$\vspace{0.1cm} \\
\noindent \stepcounter{idaxiom} $AX\_\arabic{idaxiom}$ $\property{has\_process\_assignments} \mbox{ has range }\concept{assignment}$\vspace{0.1cm} \\
\par \vspace{-0.1cm} \noindent \textbf{Property}: $\property{has\_process\_performers}$\vspace{0.1cm} \\
\noindent \textbf{Label}: Performers\vspace{0.1cm} \\
\noindent \textbf{Description}: One or more Performers MAY be entered. 
	The Performers attribute defines the resource that will be responsible 
	for the Process. The Performers entry could be in the form of a specific
        individual, a group, an organization role or position, or an
	organization.\vspace{0.1cm} \\
\noindent \stepcounter{idaxiom} $AX\_\arabic{idaxiom}$ $\property{has\_process\_performers} \mbox{ has domain }\concept{process}$\vspace{0.1cm} \\
\noindent \stepcounter{idaxiom} $AX\_\arabic{idaxiom}$ $\property{has\_process\_performers} \mbox{ has range }\datatype{xsd:string}$\vspace{0.1cm} \\
\par \vspace{-0.1cm} \noindent \textbf{Property}: $\property{has\_process\_properties}$\vspace{0.1cm} \\
\noindent \textbf{Label}: Properties\vspace{0.1cm} \\
\noindent \textbf{Description}: Modeler-defined Properties MAY be added
	to a Process. These Properties are "local" to the Process. All Tasks, 
	Sub-Process objects, and Sub-Processes that are embedded SHALL have 
	access to these Properties. The fully delineated name of these 
	properties is "process name.property name" (e.g., 
	"Add Customer.Customer Name"). If a process is embedded within another 
	Process, then the fully delineated name SHALL also be preceded by the 
	Parent Process name for as many Parents there are until the top level 
	Process. Further details about the definition of a Property can be found
	in "Property on page 276."\vspace{0.1cm} \\
\noindent \stepcounter{idaxiom} $AX\_\arabic{idaxiom}$ $\property{has\_process\_properties} \mbox{ has domain }\concept{process}$\vspace{0.1cm} \\
\noindent \stepcounter{idaxiom} $AX\_\arabic{idaxiom}$ $\property{has\_process\_properties} \mbox{ has range }\concept{property}$\vspace{0.1cm} \\
\par \vspace{-0.1cm} \noindent \textbf{Property}: $\property{has\_process\_input\_sets}$\vspace{0.1cm} \\
\noindent \textbf{Label}: Input set\vspace{0.1cm} \\
\noindent \textbf{Description}: The InputSets attribute defines the 
	data requirements for input to the Process. Zero or more InputSets MAY 
	be defined. Each Input set is sufficient to allow the Process to be 
	performed (if it has first been instantiated by the appropriate signal 
	arriving from an incoming Sequence Flow). Further details about the 
	definition of an Input-Set can be found in Section B.11.10 on page 274.\vspace{0.1cm} \\
\noindent \stepcounter{idaxiom} $AX\_\arabic{idaxiom}$ $\property{has\_process\_input\_sets} \mbox{ has domain }\concept{process}$\vspace{0.1cm} \\
\noindent \stepcounter{idaxiom} $AX\_\arabic{idaxiom}$ $\property{has\_process\_input\_sets} \mbox{ has range }\concept{input\_set}$\vspace{0.1cm} \\
\par \vspace{-0.1cm} \noindent \textbf{Property}: $\property{has\_process\_output\_sets}$\vspace{0.1cm} \\
\noindent \textbf{Label}: Output set\vspace{0.1cm} \\
\noindent \textbf{Description}: The OutputSets attribute defines the 
	data requirements for output from the Process. Zero or more OutputSets 
	MAY be defined. At the completion of the Process, only one of the 
	OutputSets may be produced--It is up to the implementation of the
	Process to determine which set will be produced. However, the IORules 
	attribute MAY indicate a relationship between an OutputSet and an 
	InputSet that started the Process. Further details about the definition
	of an OutputSet can be found in Section B.11.13 on page 275.\vspace{0.1cm} \\
\noindent \stepcounter{idaxiom} $AX\_\arabic{idaxiom}$ $\property{has\_process\_output\_sets} \mbox{ has domain }\concept{process}$\vspace{0.1cm} \\
\noindent \stepcounter{idaxiom} $AX\_\arabic{idaxiom}$ $\property{has\_process\_output\_sets} \mbox{ has range }\concept{output\_set}$\vspace{0.1cm} \\
\noindent \stepcounter{idaxiom} $AX\_\arabic{idaxiom}$ $\concept{process} \sqsubseteq  (=1) \property{has\_process\_ad\_hoc}$\vspace{0.1cm} \\
\par \vspace{-0.1cm} \noindent \textbf{Property}: $\property{has\_process\_ad\_hoc}$\vspace{0.1cm} \\
\noindent \textbf{Label}: Ad\_hoc\vspace{0.1cm} \\
\noindent \textbf{Description}: AdHoc is a boolean attribute, which has a 
	default of False. This specifies whether the Process is Ad Hoc or not. 
	The activities within an Ad Hoc Process are not controlled or sequenced 
	in a particular order, their performance is determined by the performers
	of the activities. If set to True, then the Ad Hoc marker SHALL be 
	placed at the bottom center of the Process or the Sub-Process shape for
	Ad Hoc Processes.\vspace{0.1cm} \\
\noindent \stepcounter{idaxiom} $AX\_\arabic{idaxiom}$ $\property{has\_process\_ad\_hoc} \mbox{ has domain }\concept{process}$\vspace{0.1cm} \\
\noindent \stepcounter{idaxiom} $AX\_\arabic{idaxiom}$ $\property{has\_process\_ad\_hoc} \mbox{ has range }\datatype{xsd:boolean}$\vspace{0.1cm} \\
\noindent \stepcounter{idaxiom} $AX\_\arabic{idaxiom}$ $\concept{process} \sqsubseteq (\exists\property{has\_process\_ad\_hoc}.\{"\datainstance{false}"\}) \sqcup (\exists\property{has\_process\_ad\_hoc}.\{"\datainstance{true}"\} \sqcap \\ (=1) \property{has\_process\_ad\_hoc\_ordering} \sqcap  (=1) \property{has\_process\_ad\_hoc\_completion\_condition})$\vspace{0.1cm} \\
\par \vspace{-0.1cm} \noindent \textbf{Property}: $\property{has\_process\_ad\_hoc\_ordering}$\vspace{0.1cm} \\
\noindent \textbf{Label}: AdHocOrdering\vspace{0.1cm} \\
\noindent \textbf{Description}: If the Process is Ad Hoc (the 
	AdHoc attribute is True), then the AdHocOrdering attribute MUST be 
	included. This attribute defines if the activities within the Process 
	can be performed in Parallel or must be performed sequentially. The 
	default setting is Parallel and the setting of Sequential is a 
	restriction on the performance that may be required due to shared 
	resources.\vspace{0.1cm} \\
\noindent \stepcounter{idaxiom} $AX\_\arabic{idaxiom}$ $\property{has\_process\_ad\_hoc\_ordering} \mbox{ has domain }\concept{process}$\vspace{0.1cm} \\
\noindent \stepcounter{idaxiom} $AX\_\arabic{idaxiom}$ $\property{has\_process\_ad\_hoc\_ordering} \mbox{ has range }\datatype{xsd:string}\{"\datainstance{Parallel}","\datainstance{Sequential}"\}$\vspace{0.1cm} \\
\par \vspace{-0.1cm} \noindent \textbf{Property}: $\property{has\_process\_ad\_hoc\_completion\_condition}$\vspace{0.1cm} \\
\noindent \textbf{Label}: AdHocCompletionCondition\vspace{0.1cm} \\
\noindent \textbf{Description}: If the Process is 
	Ad Hoc (the AdHoc attribute is True), then the AdHocCompletionCondition 
	attribute MUST be included. This attribute defines the conditions when 
	the Process will end.\vspace{0.1cm} \\
\noindent \stepcounter{idaxiom} $AX\_\arabic{idaxiom}$ $\property{has\_process\_ad\_hoc\_completion\_condition} \mbox{ has domain }\concept{process}$\vspace{0.1cm} \\
\noindent \stepcounter{idaxiom} $AX\_\arabic{idaxiom}$ $\property{has\_process\_ad\_hoc\_completion\_condition} \mbox{ has range }\concept{expression}$\vspace{0.1cm} \\

\vspace{0.4cm} \hrule \vspace{0.2cm} \noindent \textbf{Additional axioms described in Chapter 8 and Chapter 9 of \cite{BPMNv1.1}}\vspace{0.2cm} \hrule \vspace{0.2cm}

\noindent \stepcounter{idaxiom} $AX\_\arabic{idaxiom}$ $\concept{sequence\_flow} \sqsubseteq \forall\property{has\_connecting\_object\_source\_ref}.(\concept{intermediate\_event} \sqcup \concept{start\_event} \sqcup \concept{task} \sqcup \concept{sub\_process} \sqcup \concept{gateway})$\vspace{0.1cm} \\
\noindent \stepcounter{idaxiom} $AX\_\arabic{idaxiom}$ $\concept{sequence\_flow} \sqsubseteq \forall\property{has\_connecting\_object\_target\_ref}.(\concept{intermediate\_event} \sqcup \concept{end\_event} \sqcup \concept{task} \sqcup \concept{sub\_process} \sqcup \concept{gateway})$\vspace{0.1cm} \\
\noindent \stepcounter{idaxiom} $AX\_\arabic{idaxiom}$ $\concept{message\_flow} \sqsubseteq \forall\property{has\_connecting\_object\_source\_ref}.((\concept{intermediate\_event} \sqcap \\ \exists\property{has\_intermediate\_event\_trigger}.\concept{message\_event\_detail}) \sqcup (\concept{end\_event} \sqcap \\ \exists\property{has\_end\_event\_result}.\concept{message\_event\_detail}) \sqcup \concept{task} \sqcup \concept{sub\_process} \sqcup \concept{pool})$\vspace{0.1cm} \\
\noindent \stepcounter{idaxiom} $AX\_\arabic{idaxiom}$ $\concept{message\_flow} \sqsubseteq \forall\property{has\_connecting\_object\_target\_ref}.((\concept{intermediate\_event} \sqcap \\ \exists\property{has\_intermediate\_event\_trigger}.\concept{message\_event\_detail}) \sqcup (\concept{start\_event} \sqcap \\ \exists\property{has\_start\_event\_trigger}.\concept{message\_event\_detail}) \sqcup \concept{task} \sqcup \concept{sub\_process} \sqcup \concept{pool})$\vspace{0.1cm} \\
\noindent \stepcounter{idaxiom} $AX\_\arabic{idaxiom}$ $\concept{activity} \sqsubseteq (\forall\property{has\_flow\_object\_assignment}.(\exists\property{has\_assignment\_assign\_time}.\{"\datainstance{Start}"\} \sqcup \\ \exists\property{has\_assignment\_assign\_time}.\{"\datainstance{End}"\}))$\vspace{0.1cm} \\
\noindent \stepcounter{idaxiom} $AX\_\arabic{idaxiom}$ $\concept{start\_event} \sqsubseteq \exists\property{has\_connecting\_object\_source\_ref\_inv}.(\concept{sequence\_flow})$\vspace{0.1cm} \\
\noindent \stepcounter{idaxiom} $AX\_\arabic{idaxiom}$ $\concept{start\_event} \sqsubseteq \forall\property{has\_connecting\_object\_source\_ref\_inv}.(\concept{sequence\_flow} \sqcap \\ \exists\property{has\_sequence\_flow\_condition\_type}.\{"\datainstance{None}"\})$\vspace{0.1cm} \\
\noindent \stepcounter{idaxiom} $AX\_\arabic{idaxiom}$ $\concept{none\_intermediate\_event} \equiv \concept{intermediate\_event} \sqcap  \neg \exists\property{has\_intermediate\_event\_trigger}.\concept{event\_detail}$\vspace{0.1cm} \\
\noindent \stepcounter{idaxiom} $AX\_\arabic{idaxiom}$ $\concept{cancel\_intermediate\_event} \equiv \concept{intermediate\_event} \sqcap  (=1) \property{has\_intermediate\_event\_trigger} \sqcap \exists\property{has\_intermediate\_event\_trigger}.\concept{cancel\_event\_detail}$\vspace{0.1cm} \\
\noindent \stepcounter{idaxiom} $AX\_\arabic{idaxiom}$ $\concept{compensation\_intermediate\_event} \equiv \concept{intermediate\_event} \sqcap  (=1) \property{has\_intermediate\_event\_trigger} \sqcap \exists\property{has\_intermediate\_event\_trigger}.\concept{compensation\_event\_detail}$\vspace{0.1cm} \\
\noindent \stepcounter{idaxiom} $AX\_\arabic{idaxiom}$ $\concept{link\_intermediate\_event} \equiv \concept{intermediate\_event} \sqcap  (=1) \property{has\_intermediate\_event\_trigger} \sqcap \\ \exists\property{has\_intermediate\_event\_trigger}.\concept{link\_event\_detail}$\vspace{0.1cm} \\
\noindent \stepcounter{idaxiom} $AX\_\arabic{idaxiom}$ $\concept{error\_intermediate\_event} \equiv \concept{intermediate\_event} \sqcap  (=1) \property{has\_intermediate\_event\_trigger} \sqcap \exists\property{has\_intermediate\_event\_trigger}.\concept{error\_event\_detail}$\vspace{0.1cm} \\
\noindent \stepcounter{idaxiom} $AX\_\arabic{idaxiom}$ $\concept{conditional\_intermediate\_event} \equiv \concept{intermediate\_event} \sqcap  (=1) \property{has\_intermediate\_event\_trigger} \sqcap \exists\property{has\_intermediate\_event\_trigger}.\concept{conditional\_event\_detail}$\vspace{0.1cm} \\
\noindent \stepcounter{idaxiom} $AX\_\arabic{idaxiom}$ $\concept{message\_intermediate\_event} \equiv \concept{intermediate\_event} \sqcap  (=1) \property{has\_intermediate\_event\_trigger} \sqcap \exists\property{has\_intermediate\_event\_trigger}.\concept{message\_event\_detail}$\vspace{0.1cm} \\
\noindent \stepcounter{idaxiom} $AX\_\arabic{idaxiom}$ $\concept{timer\_intermediate\_event} \equiv \concept{intermediate\_event} \sqcap  (=1) \property{has\_intermediate\_event\_trigger} \sqcap \\ \exists\property{has\_intermediate\_event\_trigger}.\concept{timer\_event\_detail}$\vspace{0.1cm} \\
\noindent \stepcounter{idaxiom} $AX\_\arabic{idaxiom}$ $\concept{signal\_intermediate\_event} \equiv \concept{intermediate\_event} \sqcap  (=1) \property{has\_intermediate\_event\_trigger} \sqcap \exists\property{has\_intermediate\_event\_trigger}.\concept{signal\_event\_detail}$\vspace{0.1cm} \\
\noindent \stepcounter{idaxiom} $AX\_\arabic{idaxiom}$ $\concept{multiple\_intermediate\_event} \equiv \concept{intermediate\_event} \sqcap (\leq2) \property{has\_intermediate\_event\_trigger}$\vspace{0.1cm} \\
\noindent \stepcounter{idaxiom} $AX\_\arabic{idaxiom}$ $\concept{activity\_boundary\_intermediate\_event} \equiv \concept{intermediate\_event} \sqcap \\ \exists\property{has\_intermediate\_event\_target}.\concept{activity}$\vspace{0.1cm} \\
\noindent \stepcounter{idaxiom} $AX\_\arabic{idaxiom}$ $\concept{not\_activity\_boundary\_intermediate\_event} \equiv \concept{intermediate\_event} \sqcap \\ \neg \exists\property{has\_intermediate\_event\_target}.\concept{activity}$\vspace{0.1cm} \\
\noindent \stepcounter{idaxiom} $AX\_\arabic{idaxiom}$ $\concept{activity\_boundary\_intermediate\_event} \sqsubseteq (\concept{cancel\_intermediate\_event} \sqcup \\ \concept{compensation\_intermediate\_event} \sqcup \concept{error\_intermediate\_event} \sqcup \concept{conditional\_intermediate\_event} \sqcup \concept{message\_intermediate\_event} \sqcup \concept{timer\_intermediate\_event} \sqcup \concept{signal\_intermediate\_event} \sqcup \\ \concept{multiple\_intermediate\_event})$\vspace{0.1cm} \\
\noindent \stepcounter{idaxiom} $AX\_\arabic{idaxiom}$ $\concept{activity\_boundary\_intermediate\_event} \sqsubseteq (\exists\property{has\_intermediate\_event\_target}.(\concept{sub\_process} \sqcap \\ \exists\property{has\_sub\_process\_is\_a\_transaction}.\{"\datainstance{true}"\})) \sqcup (( \neg \exists\property{has\_intermediate\_event\_target}.(\concept{sub\_process} \sqcap \\ \exists\property{has\_sub\_process\_is\_a\_transaction}.\{"\datainstance{true}"\})) \sqcap ( \neg \concept{cancel\_intermediate\_event}))$\vspace{0.1cm} \\
\noindent \stepcounter{idaxiom} $AX\_\arabic{idaxiom}$ $\concept{activity\_boundary\_intermediate\_event} \sqsubseteq  \neg \exists\property{has\_connecting\_object\_target\_ref\_inv}.\concept{sequence\_flow}$\vspace{0.1cm} \\
\noindent \stepcounter{idaxiom} $AX\_\arabic{idaxiom}$ $\concept{activity\_boundary\_intermediate\_event} \sqsubseteq ( \neg \concept{compensation\_intermediate\_event} \sqcap \\ ( (=1) \property{has\_sequence\_flow\_source\_ref\_inv})) \sqcup (\concept{compensation\_intermediate\_event} \sqcap \\ \neg \exists\property{has\_sequence\_flow\_source\_ref\_inv}.\concept{sequence\_flow})$\vspace{0.1cm} \\
\noindent \stepcounter{idaxiom} $AX\_\arabic{idaxiom}$ $\concept{not\_activity\_boundary\_intermediate\_event} \sqsubseteq (\concept{none\_intermediate\_event} \sqcup \\ \concept{compensation\_intermediate\_event} \sqcup \concept{link\_intermediate\_event} \sqcup \concept{conditional\_intermediate\_event} \sqcup \concept{message\_intermediate\_event} \sqcup \concept{timer\_intermediate\_event} \sqcup \concept{signal\_intermediate\_event})$\vspace{0.1cm} \\
\noindent \stepcounter{idaxiom} $AX\_\arabic{idaxiom}$ $\concept{not\_activity\_boundary\_intermediate\_event} \sqsubseteq ( \neg (\concept{none\_intermediate\_event} \sqcup \\ \concept{compensation\_intermediate\_event}) \sqcap (\geq1) \property{has\_sequence\_flow\_target\_ref\_inv}) \sqcup ((\concept{none\_intermediate\_event} \sqcup \concept{compensation\_intermediate\_event}) \sqcap  (=1) \property{has\_sequence\_flow\_target\_ref\_inv})$\vspace{0.1cm} \\
\noindent \stepcounter{idaxiom} $AX\_\arabic{idaxiom}$ $\concept{not\_activity\_boundary\_intermediate\_event} \sqsubseteq (\concept{link\_intermediate\_event}) \sqcup \\ ( \neg \concept{link\_intermediate\_event} \sqcap  (=1) \property{has\_sequence\_flow\_source\_ref\_inv})$\vspace{0.1cm} \\
\noindent \stepcounter{idaxiom} $AX\_\arabic{idaxiom}$ $\concept{not\_activity\_boundary\_intermediate\_event} \sqsubseteq ( \neg \concept{link\_intermediate\_event}) \sqcup \\ (\concept{link\_intermediate\_event} \sqcap ( \neg (\exists\property{has\_sequence\_flow\_source\_ref\_inv}.\concept{sequence\_flow} \sqcap \\ \exists\property{has\_sequence\_flow\_target\_ref\_inv}.\concept{sequence\_flow})))$\vspace{0.1cm} \\
\noindent \stepcounter{idaxiom} $AX\_\arabic{idaxiom}$ $\concept{intermediate\_event} \sqsubseteq (( \neg \concept{message\_intermediate\_event} \sqcap  (=0) \property{has\_message\_flow\_source\_ref\_inv} \sqcap  (=0) \property{has\_message\_flow\_target\_ref\_inv}) \sqcup (\concept{message\_intermediate\_event} \sqcap (((\geq1) \property{has\_message\_flow\_source\_ref\_inv} \sqcap  (=0) \property{has\_message\_flow\_target\_ref\_inv}) \sqcup ( (=0) \property{has\_message\_flow\_source\_ref\_inv} \sqcap (\geq1) \property{has\_message\_flow\_target\_ref\_inv}))))$\vspace{0.1cm} \\
\noindent \stepcounter{idaxiom} $AX\_\arabic{idaxiom}$ $\concept{end\_event} \sqsubseteq ( \neg \exists\property{has\_end\_event\_result}.\concept{error\_event\_detail}) \sqcup (\exists\property{has\_end\_event\_result}.(\concept{error\_event\_detail} \sqcap  (=1) \property{has\_error\_detail\_error\_code}))$\vspace{0.1cm} \\
\noindent \stepcounter{idaxiom} $AX\_\arabic{idaxiom}$ $\concept{not\_activity\_boundary\_intermediate\_event} \sqsubseteq ( \neg \concept{error\_intermediate\_event}) \sqcup \\ (\exists\property{has\_intermediate\_event\_trigger}.(\concept{error\_event\_detail} \sqcap  (=1) \property{has\_error\_detail\_error\_code}))$\vspace{0.1cm} \\
\noindent \stepcounter{idaxiom} $AX\_\arabic{idaxiom}$ $\concept{receive\_task} \sqsubseteq (\exists\property{has\_receive\_task\_instantiate}.\{"\datainstance{false}"\}) \sqcup (\exists\property{has\_receive\_task\_instantiate}.\{"\datainstance{true}"\} \sqcap  \neg \exists\property{has\_activity\_loop\_type}.\concept{loop\_types})$\vspace{0.1cm} \\
\noindent \stepcounter{idaxiom} $AX\_\arabic{idaxiom}$ $\concept{receive\_task} \sqsubseteq  \neg \exists\property{has\_connecting\_object\_source\_ref\_inv}.\concept{message\_flow}$\vspace{0.1cm} \\
\noindent \stepcounter{idaxiom} $AX\_\arabic{idaxiom}$ $\concept{send\_task} \sqsubseteq  \neg \exists\property{has\_connecting\_object\_target\_ref\_inv}.\concept{message\_flow}$\vspace{0.1cm} \\
\noindent \stepcounter{idaxiom} $AX\_\arabic{idaxiom}$ $\concept{script\_task} \sqsubseteq  \neg (\exists\property{has\_connecting\_object\_target\_ref\_inv}.\concept{message\_flow} \sqcup \\ \exists\property{has\_connecting\_object\_source\_ref\_inv}.\concept{message\_flow})$\vspace{0.1cm} \\
\noindent \stepcounter{idaxiom} $AX\_\arabic{idaxiom}$ $\concept{manual\_task} \sqsubseteq  \neg (\exists\property{has\_connecting\_object\_target\_ref\_inv}.\concept{message\_flow} \sqcup \\ \exists\property{has\_connecting\_object\_source\_ref\_inv}.\concept{message\_flow})$\vspace{0.1cm} \\
\noindent \stepcounter{idaxiom} $AX\_\arabic{idaxiom}$ $\concept{gateway} \sqsubseteq (\leq2) \property{has\_sequence\_flow\_target\_ref\_inv} \sqcup ((\geq1) \property{has\_sequence\_flow\_target\_ref\_inv} \sqcap (\leq2) \property{has\_gateway\_gate})$\vspace{0.1cm} \\
\noindent \stepcounter{idaxiom} $AX\_\arabic{idaxiom}$ $\concept{event\_based\_exclusive\_gateway} \sqsubseteq (\leq2) \property{has\_gateway\_gate}$\vspace{0.1cm} \\
\noindent \stepcounter{idaxiom} $AX\_\arabic{idaxiom}$ $\property{has\_gateway\_gate\_inv} =\property{has\_gateway\_gate}^{-1}$\vspace{0.1cm} \\
\noindent \stepcounter{idaxiom} $AX\_\arabic{idaxiom}$ $\property{has\_inclusive\_gateway\_default\_gate\_inv} =\property{has\_inclusive\_gateway\_default\_gate}^{-1}$\vspace{0.1cm} \\
\noindent \stepcounter{idaxiom} $AX\_\arabic{idaxiom}$ $\property{has\_data\_based\_exclusive\_gateway\_default\_gate\_inv} =\property{has\_data\_based\_exclusive\_gateway\_default\_gate}^{-1}$\vspace{0.1cm} \\
\noindent \stepcounter{idaxiom} $AX\_\arabic{idaxiom}$ $\concept{gate} \sqsubseteq  (=1) \property{has\_gateway\_gate\_inv}$\vspace{0.1cm} \\
\noindent \stepcounter{idaxiom} $AX\_\arabic{idaxiom}$ $\concept{gate} \sqsubseteq (\exists\property{has\_gateway\_gate\_inv}.( \neg \concept{event\_based\_exclusive\_gateway})) \sqcup \\ (\exists\property{has\_gateway\_gate\_inv}.\concept{event\_based\_exclusive\_gateway} \sqcap \\ \exists\property{has\_gate\_outgoing\_sequence\_flow\_ref}.(\exists\property{has\_sequence\_flow\_condition\_type}.\{"\datainstance{None}"\}))$\vspace{0.1cm} \\
\noindent \stepcounter{idaxiom} $AX\_\arabic{idaxiom}$ $\concept{gate} \sqsubseteq (\exists\property{has\_gateway\_gate\_inv}.( \neg \concept{complex\_gateway})) \sqcup (\exists\property{has\_gateway\_gate\_inv}.\concept{complex\_gateway} \sqcap \exists\property{has\_gate\_outgoing\_sequence\_flow\_ref}.(\exists\property{has\_sequence\_flow\_condition\_type}.\{"\datainstance{None}"\}))$\vspace{0.1cm} \\
\noindent \stepcounter{idaxiom} $AX\_\arabic{idaxiom}$ $\concept{gate} \sqsubseteq (\exists\property{has\_gateway\_gate\_inv}.( \neg \concept{parallel\_gateway})) \sqcup (\exists\property{has\_gateway\_gate\_inv}.\concept{parallel\_gateway} \sqcap \exists\property{has\_gate\_outgoing\_sequence\_flow\_ref}.(\exists\property{has\_sequence\_flow\_condition\_type}.\{"\datainstance{None}"\}))$\vspace{0.1cm} \\
\noindent \stepcounter{idaxiom} $AX\_\arabic{idaxiom}$ $\concept{gate} \sqsubseteq (\exists\property{has\_gateway\_gate\_inv}.( \neg \concept{inclusive\_gateway})) \sqcup (\exists\property{has\_gateway\_gate\_inv}.\concept{inclusive\_gateway} \sqcap ( (=1) \property{has\_gateway\_gate\_inv} \sqcap \exists\property{has\_gate\_outgoing\_sequence\_flow\_ref}.(\exists\property{has\_sequence\_flow\_condition\_type}.\{"\datainstance{None}"\})) \sqcup ((\leq2) \property{has\_gateway\_gate\_inv} \sqcap \exists\property{has\_gate\_outgoing\_sequence\_flow\_ref}.(\exists\property{has\_sequence\_flow\_condition\_type}.\{"\datainstance{Expression}"\})))$\vspace{0.1cm} \\
\noindent \stepcounter{idaxiom} $AX\_\arabic{idaxiom}$ $\concept{gate} \sqsubseteq (\exists\property{has\_gateway\_gate\_inv}.( \neg \concept{data\_based\_exclusive\_gateway})) \sqcup \\ (\exists\property{has\_gateway\_gate\_inv}.\concept{data\_based\_exclusive\_gateway} \sqcap ( (=1) \property{has\_gateway\_gate\_inv} \sqcap \\ \exists\property{has\_gate\_outgoing\_sequence\_flow\_ref}.(\exists\property{has\_sequence\_flow\_condition\_type}.\{"\datainstance{None}"\})) \sqcup ((\leq2) \property{has\_gateway\_gate\_inv} \sqcap \exists\property{has\_gate\_outgoing\_sequence\_flow\_ref}.(\exists\property{has\_sequence\_flow\_condition\_type}.\{"\datainstance{Expression}"\})))$\vspace{0.1cm} \\
\noindent \stepcounter{idaxiom} $AX\_\arabic{idaxiom}$ $\concept{event\_based\_exclusive\_gateway} \sqsubseteq (\forall\property{has\_gateway\_gate}.(\exists\property{has\_gate\_outgoing\_sequence\_flow\_ref}. \\ (\exists\property{has\_connecting\_object\_target\_ref}.(\concept{receive\_task} \sqcup \concept{timer\_intermediate\_event} \sqcup \concept{signal\_intermediate\_event})))) \sqcup (\forall\property{has\_gateway\_gate}.(\exists\property{has\_gate\_outgoing\_sequence\_flow\_ref}.(\exists\property{has\_connecting\_object\_target\_ref}. \\ (\concept{message\_intermediate\_event} \sqcup \concept{timer\_intermediate\_event} \sqcup \concept{signal\_intermediate\_event}))))$\vspace{0.1cm} \\
\noindent \stepcounter{idaxiom} $AX\_\arabic{idaxiom}$ $\concept{sequence\_flow} \sqsubseteq ( \neg \exists\property{has\_sequence\_flow\_condition\_type}.\{"\datainstance{Expression}"\}) \sqcup \\ ((\exists\property{has\_sequence\_flow\_condition\_type}.\{"\datainstance{Expression}"\}) \sqcap \forall\property{has\_connecting\_object\_source\_ref}.( \neg \concept{event}))$\vspace{0.1cm} \\
\noindent \stepcounter{idaxiom} $AX\_\arabic{idaxiom}$ $\concept{sequence\_flow} \sqsubseteq ( \neg \exists\property{has\_sequence\_flow\_condition\_type}.\{"\datainstance{Expression}"\}) \sqcup \\ ((\exists\property{has\_sequence\_flow\_condition\_type}.\{"\datainstance{Expression}"\}) \sqcap \forall\property{has\_connecting\_object\_source\_ref}.( \neg \concept{parallel\_gateway}))$\vspace{0.1cm} \\
\noindent \stepcounter{idaxiom} $AX\_\arabic{idaxiom}$ $\concept{activity} \sqsubseteq ( \neg \exists\property{has\_sequence\_flow\_source\_ref\_inv}.(\concept{sequence\_flow} \\ \sqcap \exists\property{has\_sequence\_flow\_condition\_type}.\{"\datainstance{Expression}"\})) \sqcup ((\exists\property{has\_sequence\_flow\_source\_ref\_inv}.(\concept{sequence\_flow} \sqcap \exists\property{has\_sequence\_flow\_condition\_type}.\{"\datainstance{Expression}"\})) \sqcap (\leq2) \property{has\_sequence\_flow\_source\_ref\_inv})$\vspace{0.1cm} \\
\noindent \stepcounter{idaxiom} $AX\_\arabic{idaxiom}$ $\concept{sequence\_flow} \sqsubseteq (\exists\property{has\_connecting\_object\_source\_ref}.(\concept{data\_based\_exclusive\_gateway} \sqcup \\ \concept{inclusive\_gateway}) \sqcap  \neg \exists\property{has\_sequence\_flow\_condition\_type}.\{"\datainstance{None}"\}) \sqcup \\ ( \neg \exists\property{has\_connecting\_object\_source\_ref}.(\concept{data\_based\_exclusive\_gateway} \sqcup \concept{inclusive\_gateway}))$\vspace{0.1cm} \\
\noindent \stepcounter{idaxiom} $AX\_\arabic{idaxiom}$ $\concept{sequence\_flow} \sqsubseteq ( \neg \exists\property{has\_sequence\_flow\_condition\_type}.\{"\datainstance{Expression}"\}) \sqcup \\ ((\exists\property{has\_sequence\_flow\_condition\_type}.\{"\datainstance{Expression}"\}) \sqcap \forall\property{has\_connecting\_object\_source\_ref}.(\concept{task} \sqcup \concept{sub\_process} \sqcup \concept{data\_based\_exclusive\_gateway} \sqcup \concept{inclusive\_gateway}))$\vspace{0.1cm} \\
\noindent \stepcounter{idaxiom} $AX\_\arabic{idaxiom}$ $\concept{sequence\_flow} \sqsubseteq ( \neg \exists\property{has\_sequence\_flow\_condition\_type}.\{"\datainstance{Default}"\}) \sqcup \\ ((\exists\property{has\_sequence\_flow\_condition\_type}.\{"\datainstance{Default}"\}) \sqcap \forall\property{has\_connecting\_object\_source\_ref}.(\concept{activity} \sqcup \\ \concept{data\_based\_exclusive\_gateway}))$\vspace{0.1cm} \\
\noindent \stepcounter{idaxiom} $AX\_\arabic{idaxiom}$ $\concept{association} \sqsubseteq (\exists\property{has\_connecting\_object\_source\_ref}.(\concept{artifact}) \sqcap \exists\property{has\_connecting\_object\_target\_ref}. \\ (\concept{flow\_object})) \sqcup (\exists\property{has\_connecting\_object\_target\_ref}.(\concept{artifact}) \sqcap \exists\property{has\_connecting\_object\_source\_ref}.(\concept{flow\_object}))$\vspace{0.1cm} \\

\bibliographystyle{plain}
\bibliography{BPMNOntology}

\begin{thebibliography}{1}

\bibitem{BPMNv1.1}
OMG.
\newblock Business process modeling notation, v1.1.
\newblock \texttt{www.omg.org/spec/BPMN/1.1/PDF}.

\end{thebibliography}

\end{document}